\documentclass[10pt,twocolumn,letterpaper]{article}

\usepackage[pagenumbers]{cvpr} 

\usepackage{graphicx}
\usepackage{amsmath}
\usepackage{amssymb}
\usepackage{booktabs}
\usepackage[utf8]{inputenc} 
\usepackage[T1]{fontenc}    
\usepackage[pagebackref,breaklinks,colorlinks]{hyperref}
\usepackage{url}            
\usepackage{amsfonts}       
\usepackage{nicefrac}       
\usepackage{microtype}      
\usepackage{xcolor}         
\usepackage{adjustbox}
\usepackage{times}
\usepackage{epsfig}
\usepackage{amsmath}
\usepackage{amssymb}
\usepackage{color}
\usepackage{colortbl}
\usepackage{pifont}
\usepackage{multirow}
\usepackage[vlined, ruled, boxed]{algorithm2e}
\usepackage{rotating}
\usepackage{dsfont}
\usepackage{array}
\usepackage{enumitem}
\usepackage[accsupp]{axessibility}
\usepackage{lipsum}
\usepackage{float}

\definecolor{turquoise}{cmyk}{0.65,0,0.1,0.3}
\definecolor{purple}{rgb}{0.65,0,0.65}
\definecolor{dark_green}{rgb}{0, 0.5, 0}
\definecolor{orange}{rgb}{0.8, 0.6, 0.2}
\definecolor{red}{rgb}{0.9, 0.1, 0.1}
\definecolor{darkred}{rgb}{0.6, 0.1, 0.05}
\definecolor{blueish}{rgb}{0.0, 0.3, .6}
\definecolor{light_gray}{gray}{0.95}
\definecolor{pink}{rgb}{1, 0, 1}
\definecolor{greyblue}{rgb}{0.25, 0.25, 1}

\usepackage{blindtext}

\newcommand{\figref}[1]{Fig. \ref{#1}}
\newcommand{\tabref}[1]{Table \ref{#1}}

\newcommand{\secref}[1]{Sec. \ref{#1}}
\newcommand{\algref}[1]{Alg. \ref{#1}}

\renewcommand{\paragraph}[1]{\vspace{1em}\noindent\textbf{#1}.}
\newcommand\blfootnote[1]{%
  \begingroup
  \renewcommand\thefootnote{}\footnote{#1}%
  \addtocounter{footnote}{-1}%
  \endgroup
}

\usepackage[capitalize]{cleveref}
\crefname{section}{Sec.}{Secs.}
\Crefname{section}{Section}{Sections}
\Crefname{table}{Table}{Tables}
\crefname{table}{Tab.}{Tabs.}

\def\cvprPaperID{10828} 
\def\confName{CVPR}
\def\confYear{2022}

\makeatletter
 \def\hlinewd#1{%
      \noalign{\ifnum0=`}\fi\hrule \@height #1 \futurelet
      \reserved@a\@xhline}

\newcommand{\cmark}{\ding{51}}%
\newcommand{\xmark}{\ding{55}}%
\newcommand{\RomanNumeralCaps}[1]
    {\MakeUppercase{\romannumeral #1}}

\newcommand{\loss}[1]{\mathcal{L}_\text{#1}}

\definecolor{blue}{rgb}{0,0,1}

\begin{document}

\title{Pin the Memory: Learning to Generalize Semantic Segmentation}

\author{Jin Kim$^1$ \quad\quad Jiyoung Lee$^2$ \quad\quad Jungin Park$^1$ \quad\quad Dongbo Min$^3$\thanks{Corresponding authors.} \quad\quad Kwanghoon Sohn$^{1*}$ \\
$^1$Yonsei University \quad\quad $^2$NAVER AI Lab \quad\quad
$^3$Ewha Womans University\\
{\tt\small $\lbrace$kimjin928, newrun, khsohn$\rbrace$@yonsei.ac.kr} \quad\quad
\tt\small lee.j@navercorp.com \quad\quad dbmin@ewha.ac.kr}
 
\maketitle

\begin{abstract}
\vspace{-4pt}
\blfootnote{This research was supported by the National Research Foundation of Korea (NRF) grant funded by the Korea government (MSIP) (NRF2021R1A2C2006703), the Yonsei University Research Fund of 2021 (2021-22-0001), and the Mid-Career Researcher Program through the NRF of Korea (NRF-2021R1A2C2011624).}
The rise of deep neural networks has led to several breakthroughs for semantic segmentation.
In spite of this, a model trained on source domain often fails to work properly in new challenging domains, that is directly concerned with the generalization capability of the model. 
In this paper, we present a novel memory-guided domain generalization method for semantic segmentation based on meta-learning framework.
Especially, our method abstracts the conceptual knowledge of semantic classes into categorical memory which is constant beyond the domains.
Upon the meta-learning concept, we repeatedly train memory-guided networks and simulate virtual test to 1) learn how to memorize a domain-agnostic and distinct information of classes and 2) offer an externally settled memory as a class-guidance to reduce the ambiguity of representation in the test data of arbitrary unseen domain.
To this end, we also propose memory divergence and feature cohesion losses, which encourage to learn memory reading and update processes for category-aware domain generalization.
Extensive experiments for semantic segmentation demonstrate the superior generalization capability of our method over state-of-the-art works on various benchmarks.\footnote{ \url{https://github.com/Genie-Kim/PintheMemory}}
\vspace{-20pt}

\end{abstract}
\section{Introduction}
\label{sec:intro}

Semantic segmentation, assigning a semantic class label to each pixel, is a classical research topic for visual understanding in computer vision.
The recent tremendous progress in semantic segmentation has been dominated by deep neural networks trained on large amounts of densly annotated datasets.
Despite its success, models trained with a given dataset (\textit{source}) do not generalize well in a new domain (\textit{target}) that the models have not seen during training.
Overcoming the domain shift issue caused by the different data distributions of two domains is crucial to deal with unexpected and unseen data, especially for replacing human tasks such as medical diagnosing or autonomous driving.

 \begin{figure}[t]
   \centering
   {\includegraphics[width=0.9\linewidth]{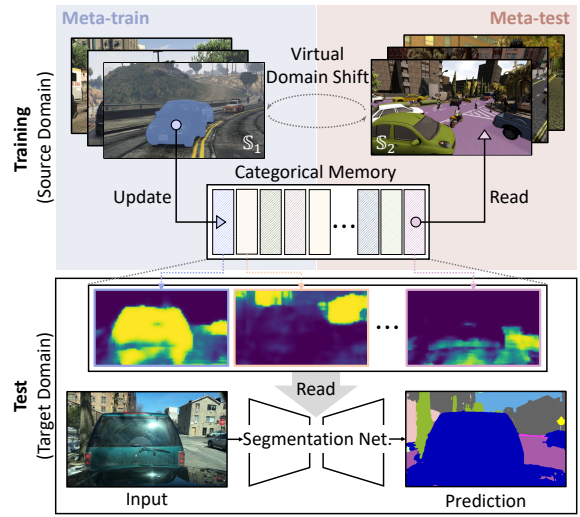}}
    \vspace{-5pt}
  \caption{The illustration of our memory-guided meta-learning algorithm for domain generalization. Our method learns how to memorize domain-agnostic categorical knowledge that can provide an external guide to the test data in unseen target domain.} \label{fig:1}
  \vspace{-10pt}
\end{figure}

In order to mitigate severe performance degradation from the domain shift~\cite{ben2007analysis,ganin2016domain}, unsupervised domain adaptation (UDA) approaches~\cite{du2019ssf, luo2019taking, wang2020classes} have been proposed to bridge the domain gap using unlabelled images of the target domain.
These methods have introduced inventive learning strategies to learn domain invariant features~\cite{huang2020contextual,zhang2021prototypical,zhang2019category,zheng2021rectifying,zou2018unsupervised,guan2021scale} or align source and target domain to unified space~\cite{he2021multi,hoffman2018cycada,saito2018maximum,zhao2019multi,zhao2021madan}.
Though they have shown impressive results against domain shift, collecting data from the target domain is often impractical.
Moreover, the scalability of the model is restricted as UDA requires network re-training or fine-tuning for the new target domain, thereby exposing limitations in terms of being able to generalize to `any' unseen domains.

To overcome those limitations, domain generalization (DG) methods have been developed to learn robust models against variants of data distribution across arbitrary unseen domains~\cite{jigsaw,epifcr,eisnet,RSC,mldg,XAI}.
It is much harder than UDA in that no target domain data is available during training.
Some methods heuristically define domain-biased information as style (\eg, texture, color) to explicitly augment it~\cite{yue2019domain,huang2021fsdr}, or erase style through instance normalization~\cite{pan2018ibnnet} and channel covariance whitening~\cite{choi2021robustnet}.
Despite their efforts, existing methods still show limited performance for use in real-world applications. 
But, it is natural that human visual system adapts stably even when facing scenes that they have never seen before.
Where does this difference in generalization ability between humans and machines come from?

We argue that there is an important missing piece in this puzzle.
The conceptual knowledge of humans~\cite{binder2011neurobiology}, also known as semantic memory, is abstracted from actual experiences in the reusable form and is generalized to support a variety of cognitive activities such as event reconstruction~\cite{irish2012considering,irish2013pivotal} and object recognition~\cite{saumier2002semantic}.
Inspired by this, we consider that human's knowledge concept can be effectively utilized in domain generalization by remembering the shared information of each class.
For example, the style of the car may vary depending on the domain, but the basic features to configure the car (\eg wheel, door, bumper, headlight) remain unchanged.
Namely, the guidance of such prior knowledge about concurrent features can help to improve the generalization capability of machines.

In this work, we propose a novel memory-guided meta-learning framework to capture and memorize co-occurrent categorical knowledge between objects of the same class across domains.
The objective of this framework is to assign shared information of each class into external memory slots and reuse the categorical concept for robust semantic segmentation in arbitrary unseen domains.
To this end, we split source domain data into meta-training and meta-testing sets to explicitly mimic domain shift in the inference, allowing the network to store and invoke memory corresponding to domain-agnostic prototypes of class patterns, as shown in \figref{fig:1}.
That is, our method enables category-aware generalization for semantic segmentation, unlike previous DG approaches~\cite{pan2018ibnnet,choi2021robustnet,yue2019domain} that only concentrate on globally inferring domain-agnostic representations.
Moreover, we introduce a memory divergence loss and a feature cohesion loss which boost discriminative power of memory and make more domain-invariant representations from the encoder, respectively.
Consequently, our method achieves superior performance gain over existing DG approaches on multiple unseen real-world benchmarks.
Without re-training or fine-tuning, our results are even on par with the  multi-source UDA methods~\cite{zhao2018mdan, zhao2021madan, zhao2019multi, he2021multi}, where the training images are given from both source and target domains. 

In summary, our key contributions are as follows:
(\romannumeral 1) We present a novel approach to domain generalization for semantic segmentation with memory module to exploit domain-agnostic categorical knowledge of classes.
(\romannumeral 2) We introduce the memory-guided meta-learning algorithm that improves the representation power of the memory-guided feature by exposing the model to mismatched data distribution. 
(\romannumeral 3) We propose two complementary losses, including memory divergence loss and feature cohesion loss, that promote power for an embedded feature to find the apposite class memory. 
(\romannumeral 4) Extensive experiments prove the significance of category-aware generalization on both single- or multi-source settings.

\vspace{-3pt}

\section{Related Work}
\label{sec:related}

\vspace{-14pt}
\paragraph{Domain adaptation and generalization}
There are wide investigations towards better generalization of deep networks to mitigate the domain distribution discrepancy between source (training) and target (testing) domains.
In particular, unsupervised domain adaptation (UDA) approaches have been proposed to rectify such domain mismatch by leveraging the unlabeled target images for training~\cite{hoffman2018cycada,zhang2021prototypical,chen2019progressive, kumar2018co, zhao2019learning, du2019ssf, luo2019taking, wang2020classes}.
Recently, multi-source UDA methods~\cite{zhao2018mdan, zhao2021madan, zhao2019multi, he2021multi} have been introduced in a more practical scenario, where the training data is collected from multiple synthetic datasets~\cite{gtav,synthia,virtualkitti}.
Despite those efforts, deep networks often suffer from unseen novel domains in the real-world.
It yields a domain generalization (DG) problem~\cite{blanchard2011generalizing}, that is more challenging than UDA in that the target domain data is not available.
Recent works on DG are roughly categorized as two-fold: learning domain-invariant features~\cite{motiian2017unified,li2018domain,muandet2013domain,XAI} and augmenting the training samples~\cite{zhou2020learning,zhou2020domain,jigsaw,eisnet,RSC}.
However, the majority of the DG methods still focuses on the task of classifying the entire image into one class, while our approach aims to generalize the networks to prevent a large performance drop of semantic segmentation in urban scene.

\begin{figure*}[!ht]
	\centering
    {\includegraphics[width = 0.97\linewidth]{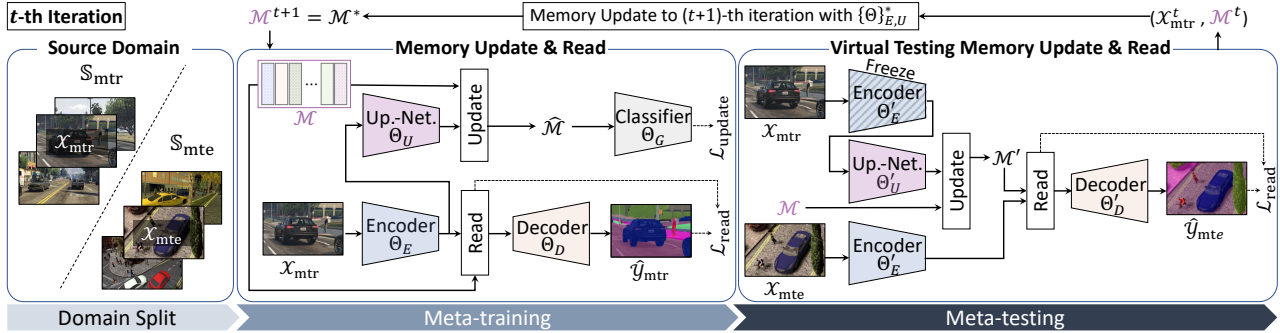}}
    \vspace{-7pt}
 	\caption{Overall training process of our method, consisting of domain split, meta-training and meta-testing steps for every iteration.
 	}\label{fig:2}\vspace{-10pt}
\end{figure*}
 
\vspace{-10pt}
\paragraph{Domain generalization for semantic segmentation}
This task has received relatively less attention compared to its importance in many real-world applications including autonomous driving in the wild.
One of the promising solutions is a domain randomization~\cite{yue2019domain,huang2021fsdr} to generate new training samples using data augmentation.
However, it requires a lot of cost for training and it is practically difficult to cover the real-world distribution with the data augmentation only.
Alternatively, based on the theoretical intuition from the normalization, some approaches have tried to normalize global feature by erasing the style-specific information of each domain~\cite{pan2018ibnnet,cnsn,choi2021robustnet}. 
In contrast to those methods that concern global representation only, we propose a categorical memory-guided framework for class-wise domain generalization.
Meanwhile, recent papers~\cite{asg, csg} pointed out a crucial role of the diversity of learned feature from the synthetic data to prevent overfitting to the source domain in segmentation task.
Inspired by this, we employ a meta-learning to our framework for virtually testing the stored memory under different data distribution, promoting the only common knowledge of class to be saved for generalization.

\vspace{-10pt}
\paragraph{Meta-learning}
The model-agnostic meta-learning~\cite{maml, metauniversal} is one of the most popular methods of meta-learning (a.k.a \textit{learning-to-learn}), where an episodic training scheme has been designed for making multi-order of gradient descent for few-shot learning.
The key idea of the episodic training, separating the learning steps into meta-train and meta-test to mimic the training and evaluation steps, has inspired other studies~\cite{mldg,metareg, masf, li2019feature, li2020sequential,epifcr,sbdg} to develop meta-learning based methods for domain generalization.
Most related to ours, Zhen \etal~\cite{zhen2020learning} recently proposed a long-term memory with meta-learning that stores semantic information for few-shot learning, where the gradient from the updating memory does not feedback to the networks. 
Zhao \etal~\cite{mldgreid} claimed that the asynchronous gradient update among the sub-networks destabilizes meta-optimization, and simply treated the memory as a non-parametric module to solve the problem.
Our method is orthogonal to these works in that we aims to learn the network to generalize categorical memory update and reading process through meta-learning.

\vspace{-10pt}
\paragraph{Memory networks}
The recent advances of memory networks~\cite{santoro2016meta,ramalho2018adaptive,bartunov2019meta} enhance the capability of neural networks by recording information stably.
Although \cite{wu2021learning, zhen2020learning} proposed the long-term memory modules with meta-learning like our method, they improved reading performance only without consideration for memory writing.
Compared to the previous works, our memory module stores long-term memory in whole training steps with meta-learning, which helps to robustly read and write memory to domain shift.
The memory in~\cite{bartunov2019meta} approximated to neural networks requiring several computations to read memory, but our method is more efficient than~\cite{bartunov2019meta} with a once estimation.
Significantly, the memory networks have been effective in several segmentation-related tasks~\cite{alonso2021semi, wang2021exploring, hu2021region, jin2021mining,xie2021efficient, hu2021learning, oh2019video, wu2021learning}. 
For instance, Jin \etal~\cite{jin2021mining} stored dataset-level surrounding contexts of various classes to augment pixel-level representations.
On the contrary, we store domain-agnostic information into the memory to contain common features of semantic categories. 


\section{Proposed Method}\vspace{-3pt}

\subsection{Problem Statement and Overview} \vspace{-3pt}

Given an image from an unseen target domain, domain generalization aims to protect the performance of the segmentation network trained with a set of observable source, $\mathbb{S}$, where the networks basically consist of encoder and decoder (pixel-wise classifier).
An intuitive approach to DG is to learn the segmentation networks by simply combining all source domains into one training dataset and training with standard segmentation loss such as cross-entropy~\cite{fcn}.
However, this naive aggregation method is overly suited to the source domain and thus shows enormous performance deterioration when domain shift occurs in the inference.

To solve this problem, we propose a memory-guided meta-learning framework to prevent performance degradation of semantic segmentation in the unseen domains at test time, as shown in \figref{fig:2}.
By configuring an artificial domain shift with data augmentation or domain splitting, we allow the network to update and read memory on the specified domains in the meta-learning framework so that the network learns how to remember conceptual knowledge in the presence of the domain shift.
In the following section, we first describe the memory read and update procedure (\secref{sec:32}) and then memory-guided meta-learning framework with loss functions (\secref{sec:33}).\vspace{-3pt}

\subsection{Memory Module}\label{sec:32} \vspace{-3pt}
The memory module is incorporated with the segmentation backbone network to memorize the common features of each class into memory matrix $\mathcal{M} \in \mathbb{R}^{N \times C}$, where $N$ is the number of classes and $C$ is a channel dimension of the encoder feature.
We next explain initialization, update and reading processes of our memory module in details.

\begin{figure}[t]
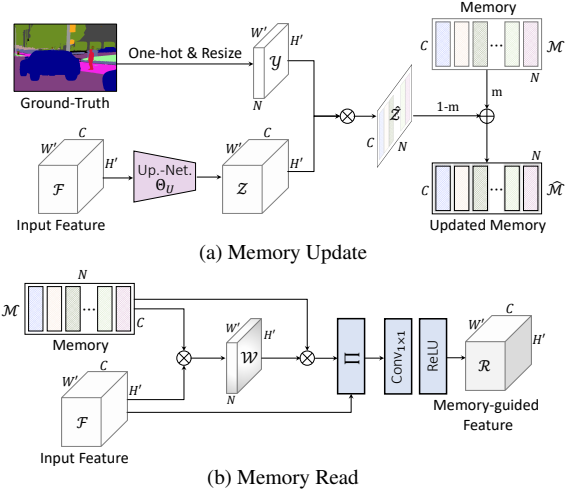

    \centering
      \begin{subfigure}{0.9\linewidth}
          \includegraphics[width=1.0\linewidth]{figure/update.pdf}
          \caption{Memory Update}
          \label{fig:3a}
      \end{subfigure}
      \hfill
      \\
      \centering
      \begin{subfigure}{0.9\linewidth}
          \includegraphics[width=1.0\linewidth]{figure/read.pdf}
          \vspace{-15pt}
          \caption{Memory Read}
          \label{fig:3b}
      \end{subfigure}
      \hfill
      \\
      \vspace{-7pt}
      \caption{Illustration of memory update and reading operations.
      }
      \label{fig:3} 
      \vspace{-10pt}
\end{figure}

\vspace{-10pt}
\paragraph{Initialization}
As the preliminary step, $\ell_2$-normalized feature maps are extracted from all training images in the source domains through an encoder $E$ with parameters $\Theta_E$, pre-trained on ImageNet~\cite{2015imagenet}.
To initialize the memory matrix with these feature maps, we calculate a mean feature vector for each class by masking the regions with ground-truth segmentation maps.
Since the initial memory matrix composing of class-wise mean vectors is in a very noisy state, our method learns to update this by storing more discriminative and domain-agnostic class-wise features in the memory.

\vspace{-10pt}
\paragraph{Update}
We adopt a memory updating network $U$ consists of a $1 \times1$ convolution layer with the residual connection.
As shown in \figref{fig:3a}, the memory updating network with parameters $\Theta_U$ transforms $\ell_2$-normalized feature map $\mathcal{F} \in \mathbb{R}^{C \times H' \times W'}$ of an input image $\mathcal{X} \in \mathbb{R}^{3 \times H \times W}$ into $\mathcal{Z} = U(\mathcal{F})$, where $H\times W$ is an original size of the image, and $H'\times W'$ is a reduced size by a pooling operation in the backbone networks\footnote{$H',W'$ varies depending on the output stride of backbone networks such as FCN~\cite{fcn}, DeepLabV2~\cite{deeplabv2}, DeepLabV3+~\cite{deeplabv3p}, etc.}.
In order to update the $n$-th item $\mathcal{M}[n]$ in the class-wise memory, we perform an average pooling over the masked region by referring to the segmentation mask of the $n$-th class as follows:
\begin{equation}\label{eq:pooling}
    \hat{\mathcal{Z}}[n] = (\mathcal{Y}[n]\mathcal{Z}^\top)/K_n,
\end{equation}
where $K_n$ is the number of pixels belonging to $n$-th class in the ground-truth, $\hat{\mathcal{Z}} \in \mathbb{R}^{N\times C}$ is a masked feature map and $\mathcal{Y}$ is a one-hot segmentation ground-truth which has a size of $N \times H'W'$.
Note that $\mathcal{Z}$ is reshaped as $C \times H'W'$.
Then the $n$-th channel of masked feature vector $\hat{\mathcal{Z}}[n]$ is used to update a memory item using moving average.
\begin{equation}\label{eq:momentumupdate}
    \hat{\mathcal{M}}[n] = m\cdot \mathcal{M}[n] + (1-m)\cdot \hat{\mathcal{Z}}[n],
\end{equation}
where $\mathcal{\hat{M}}[n]$ is an updated memory and the momentum $m$ is set as 0.8 empirically.
This is repeated for all classes, which is expressed as below:
\begin{equation}\label{eq:memupdate}
        \hat{\mathcal{M}} = \textbf{update}(\mathcal{M},\mathcal{X};\lbrace\Theta\rbrace_{E,U}),
\end{equation}
where a set of parameters $\Theta_E$ and $\Theta_U$ is denoted as $\lbrace\Theta\rbrace_{E,U}$.

\vspace{-8pt}
\paragraph{Read}
As depicted in \figref{fig:3b}, we read the stored memory items with the encoded feature map $\mathcal{F}$ to represent a memory-guided feature map $\mathcal{R} \in \mathbb{R}^{C \times H' \times W'}$ which is used in the decoder.
To aggregate a corresponding memory item along each feature location, we compute an memory weight matrix $\mathcal{W} \in \mathbb{R}^{N \times H' \times W'}$ via cosine similarity and normalize it with softmax function as:
\begin{equation}\label{eq:readweight}
    \mathcal{W}[n] = \cfrac{\exp(\mathcal{M}[n]\mathcal{F} )}{\sum_{n'=1}^{N}{\exp(\mathcal{M}[n'] \mathcal{F})}},
\end{equation}
where $\mathcal{F}$ and $\mathcal{W}$ are permuted as $C \times H'W'$ and $N \times H'W'$ respectively.
The memory-guided feature map $\mathcal{R}$ is obtained by fusing the original feature map $\mathcal{F}$ and weighted memory feature $\mathcal{M}^\top \mathcal{W}$ as follows:
\begin{equation}\label{eq:reading}
    \mathcal{R} = \texttt{ReLU}( \texttt{Conv}_{1\times1}(\Pi(\mathcal{F}, \mathcal{M}^\top\mathcal{W})) ),
\end{equation}
where $\Pi(\cdot)$ denotes a concatenation operation.
Note that $\mathcal{M}^\top\mathcal{W}$ is re-permuted to have a size of $C \times H' \times W'$. 
We add $1 \times 1$ convolution layer to make the channel size of $\mathcal{R}$ to $C$.
Finally, a predicted segmentation probability map $\hat{\mathcal{Y}}$ is estimated by passing $\mathcal{R}$ into the decoder.
From now on, we denote the $1 \times 1$ convolution layer with decoder as $D$ with parameters $\Theta_D$.

\subsection{Learning to Generalize Update and Read}\label{sec:33}
\vspace{-3pt}

Compared to the previous DG methods based on meta-learning~\cite{mldg,li2020sequential,metareg} that do not use external prior knowledge, our method leverages meta-learning to achieve two goals.
First, the domain invariant categorical knowledge of each class is saved in a form of external memory that can offer a class-wise guidance for robustly segmenting an image from unseen domains.
Second, we reinforce our network to robustly classify each unseen image pixel to a category label against intra-class and cross-domain variations.
Specifically, we randomly split the available source domains $\mathbb{S}$ into meta-train domains $\mathbb{S}_{\text{mtr}}$ and meta-test domains $\mathbb{S}_{\text{mte}}$ at every iteration step. Then, we repeatably memorize class-wise features from $\mathbb{S}_{\text{mtr}}$ and test whether the network properly works with the memory on the held-out $\mathbb{S}_{\text{mte}}$. The overall training procedure is summarized in \figref{fig:2} and \algref{alg:1}.

\vspace{-8pt}
\paragraph{Meta-training}
Given an input image $\mathcal{X}_{\text{mtr}} \in \mathbb{S}_{\text{mtr}}$, the encoder computes a feature map $\mathcal{F}_{\text{mtr}}$ and augments it by using the memory $\mathcal{M}$ through the reading operation.
We calculate a per-pixel cross-entropy loss~\cite{fcn}, \ie segmentation loss $\loss{\text{seg}}$, with ground-truth map $\mathcal{Y}_{\text{mtr}}$ and the estimated output $ \hat{\mathcal{Y}}_{\text{mtr}}$ from the decoder.
However, $\loss{\text{seg}}$ does not necessarily guarantee that the encoder features in the same class lie close in the feature embedding space.
Therefore, we further propose a feature cohesion loss $\loss{\text{coh}}$ to encourage semantic features to be locally assembled based on each memory item:
    \begin{equation}\label{eq:cohloss}
        \loss{\text{coh}} = \cfrac{1}{H'W'}\sum\nolimits_{j=1}^{H'W'} -\mathcal{Y}^{\top}_\text{mtr}[j]\log(\mathcal{W}_{\text{mtr}}[j]),
    \end{equation}
where $\mathcal{W}_{\text{mtr}}$ is computed as \eqref{eq:readweight}.

In addition, the class-wise features in the memory should be far enough apart from each other to be discriminative.
To ensure this, we propose a memory divergence loss $\loss{\text{div}}$ that increases the distance between memory items, as well as maximizes the decision margin:
\vspace{-4pt}
\begin{equation}\label{eq:divloss}
\begin{aligned}
    \loss{\text{div}} = 
    & \sum\nolimits_{n=1}^{N}(  -\mathcal{I}[n]\log(G(\hat{\mathcal{M}}[n]^\top)) \\ 
    & + 2\cdot\sum\nolimits_{n'\neq n}^{N}{ \frac{\text{max}(\hat{\mathcal{M}}[n]\hat{\mathcal{M}}[n']^\top, 0)}{N(N-1)}}),
\end{aligned}
    \vspace{-4pt}
\end{equation}
where $\mathcal{I}$ is the identity matrix of size $N\times N$, and a memory classifier $G$ includes a FC layer with parameters $\Theta_G$ and has an output size of $N$ after the softmax.
In \eqref{eq:divloss}, the first term is for the memory classification, and the second term is similar to cosine embedding loss~\cite{wang2018cosface} with a margin set to 0, empirically scaled double.
While the divergence loss improves inter-class dispersion, the feature cohesion loss increases intra-class compactness of the encoder features among distinct memory items.
We carefully note that $\loss{\text{div}}$ is calculated for the newly estimated memory $\hat{\mathcal{M}}$, while the reading process use the $\mathcal{M}$ updated in the last iteration step.
It is because the reading process aims to guide the feature map well with previously saved memory, and the update process focuses on saving even better patterns into memory and widening the gap between memory items with $\loss{\text{div}}$.

To clarify, we define $\loss{\text{read}}$ with the segmentation and feature cohesion losses computed in the memory reading operation, and $\loss{\text{update}}$ with the memory divergence loss computed when updating the memory item, respectively:
\vspace{-4pt}
\begin{equation}\label{eq:readupdateloss}
    \begin{aligned}
        &  \loss{\text{read}} (\mathcal{M},\mathcal{X}_{\text{mtr}}\:;\lbrace \Theta \rbrace_{ E,D }) =  \loss{\text{seg}} + \lambda_{1} \loss{\text{coh}}, \\
        &  \loss{\text{update}}(\mathcal{M},\mathcal{X}_{\text{mtr}}\:;\lbrace \Theta \rbrace_{ E,U,G })  =\lambda_{2} \loss{\text{div}},
    \end{aligned}
    \vspace{-4pt}
\end{equation}
where $\lambda_{1}$ and $\lambda_{2}$ are hyper-parameters.
Consequently, the updated network parameters are obtained as follows:
\vspace{-4pt}
\begin{equation}\label{eq:netupdate1}
    \begin{aligned}
       \lbrace\Theta\rbrace'_{E,U,D}, \Theta_G^* 
       & \leftarrow  \lbrace\Theta\rbrace_{E,U,D,G} \\
       &  - \alpha\nabla_{\Theta}
       \loss{\text{read}}(  \mathcal{M},\mathcal{X}_{\text{mtr}}\:;\lbrace \Theta \rbrace_{ E,D } ) \\ 
       & -\alpha\nabla_{\Theta}
       \loss{\text{update}}(\mathcal{M},\mathcal{X}_{\text{mtr}}\:;\lbrace \Theta \rbrace_{ E,U,G } ),
    \end{aligned}
    \vspace{-4pt}
\end{equation}
where $\alpha$ is a learning rate of the meta-training step. 
Since the memory classifier $G$ is not used in the meta-testing step, $\Theta^*_G$ is the final updated parameter of $G$ in this training iteration.

\vspace{-10pt}
\paragraph{Meta-testing}
The goal of meta-testing in our method is to not only virtually simulate \textit{testing the networks} on new data statistics but also characterize \textit{learning to update categorical memory} to work well across the domains.
Moreover, the effectiveness of the memory divergence loss for the updating network $U$ should be tested within the meta-testing process.

\begin{algorithm}[t]
\small
\caption{Overall Training Procedure}\label{alg:1}

 Initialize $\lbrace\Theta\rbrace_{E,U,D,G}$ and $\mathcal{M}$ at $t=0$
 
 \While{$t<T$}{
    {Randomly split $\mathbb{S}$ into $\mathbb{S}_\text{mtr}$ and $\mathbb{S}_\text{mte}$}
    
    \SetKwProg{Fn}{}{:}{}
    \Fn{\textbf{Meta-training}}{
    
    Sample batch $\mathcal{X}_\text{mtr}^t$ = $\lbrace \mathcal{X}_\text{mtr}^b\rbrace^{B}_{b=1}$ from $\mathbb{S}_\text{mtr}$\\
    Compute $\loss{\text{read}}$ with ($\mathcal{X}_\text{mtr}^t$, $\mathcal{M}$, $\lbrace \Theta \rbrace_{ E,D }$)\\
    $\hat{\mathcal{M}} \leftarrow \textbf{update}(\mathcal{M},\mathcal{X}_\text{mtr}^t;\lbrace\Theta\rbrace_{E,U})$\\
    Compute $\loss{\text{update}}$ with ($\hat{\mathcal{M}}$, $\Theta_{G}$)\\
    Update $\lbrace\Theta\rbrace'_{E,U,D}$,$\Theta_G^*$ from $\lbrace\Theta\rbrace_{E,U,D,G}$ in \eqref{eq:netupdate1}\\
    }
    
    \SetKwProg{Fn}{}{:}{}
    \Fn{\textbf{Meta-testing}}{
    $\mathcal{M}' \leftarrow
    \textbf{update}(\mathcal{M},\mathcal{X}_\text{mtr}^t;\text{copy}(\Theta'_E),\Theta'_U)$\\
    Sample batch  $\mathcal{X}_\text{mte}^t$ = $\lbrace \mathcal{X}_\text{mte}^b\rbrace^{B}_{b=1}$ from $\mathbb{S}_\text{mte}$\\
    Compute $\loss{\text{read}}$ with ($\mathcal{X}_\text{mte}^t, \; \mathcal{M}', \: \lbrace \Theta \rbrace'_{ E,D }$ )\\
    Update $\lbrace \Theta \rbrace_{ E,U,D }^{*}$ from $\lbrace \Theta \rbrace_{ E,U,D }'$ in \eqref{eq:netupdate2}\\
    $\mathcal{M}^{*} \leftarrow \textbf{update}(\mathcal{M},\mathcal{X}^{t}_{\text{mtr}} \:;
    \text{copy}(\lbrace \Theta \rbrace_{ E,U }^{*}))$\\
    }
    
    $\lbrace\Theta\rbrace_{E,U,D,G}\leftarrow \lbrace\Theta\rbrace_{E,U,D,G}^{*}$\\
    $\mathcal{M}\leftarrow\mathcal{M}^{*}$\\
    $t\leftarrow t+1$\\
 }
\end{algorithm}
\setlength{\textfloatsep}{0pt}

With these reasons, we carefully design meta-testing process that re-updates the memory using the meta-updated networks' parameters $\lbrace\Theta\rbrace'_{E,U}$ and the meta-train image $\mathcal{X}_{\text{mtr}}$:
\begin{equation}\label{eq:memsubupdate}
    \mathcal{M}' = \textbf{update}(\mathcal{M},\mathcal{X}_{\text{mtr}} \:;
    \text{copy}(\Theta'_E),\Theta'_U),
\end{equation}
where $\text{copy}(\Theta'_E)$ indicates $\Theta'_E$ is frozen.
We obtain the memory once again with meta-train data, not meta-test data, because we will reuse the learned memory without update process in inference. Since this memory $\mathcal{M}'$ is used to segment meta-test data $\mathcal{X}_{\text{mte}}$,
this novel step also allows the memory updating network's parameters $\Theta_U$ to receive the second-order gradient feedback on whether the updated memory $\mathcal{M}'$ is applicable on different domains.
By freezing the encoder's parameter $\Theta'_E$, we can avoid unstable meta-learning caused by the asynchronous gradient update between the encoder and the other networks~\cite{mldgreid}.
Guided by $\mathcal{M}'$, the network parameters are updated with the reading loss $\loss{\text{read}}$ for the image $\mathcal{X}_{\text{mte}}$ from meta-test domain $\mathbb{S}_{\text{mte}}$ as follows:
\begin{equation}\label{eq:netupdate2}
\begin{aligned}
    \lbrace \Theta \rbrace_{ E,U,D }^{*} \leftarrow
    & \lbrace \Theta \rbrace_{ E,U,D }\\
    & - \beta \nabla_{\Theta}\loss{\text{read}}(\mathcal{M}',\mathcal{X}_{\text{mte}}\:;\lbrace \Theta \rbrace_{ E,U,D }'),
\end{aligned}
\end{equation}
where $\beta$ is a learning rate of the meta-testing step.
Note that the second-order gradient is generated from the last term of \eqref{eq:netupdate2} by differentiating ${\lbrace \Theta \rbrace}'$ obtained from \eqref{eq:netupdate1} with the original parameters ${\lbrace \Theta \rbrace}$.
Using the updated network parameters, we initialize the memory $\mathcal{M}^{*}$ that will be used in the next training iteration step:
\begin{equation}\label{eq:memlastupdate}
    \mathcal{M}^{*}= \textbf{update}(\mathcal{M},\mathcal{X}_{\text{mtr}} \:;
    \text{copy}(\lbrace \Theta \rbrace_{ E,U }^{*})).
\end{equation}
The optimization in meta-testing step allows (1) writing the domain-agnostic features to the current memory $\mathcal{M}$ from the meta-train image in \eqref{eq:memlastupdate} and (2) ensuring the generalization capability of the memory-guided feature of meta-test image.

\vspace{-3pt}
\begin{table*}[!ht]
  \centering
    \resizebox{\linewidth}{!}{
    \begin{tabular}{
    >{\raggedright}m{0.015\linewidth}>{\raggedright}m{0.12\linewidth}|
    >{\centering}p{0.025\linewidth}>{\centering}p{0.025\linewidth}
    >{\centering}p{0.025\linewidth}>{\centering}p{0.025\linewidth}
    >{\centering}p{0.025\linewidth}>{\centering}p{0.025\linewidth}
    >{\centering}p{0.025\linewidth}>{\centering}p{0.025\linewidth}
    >{\centering}p{0.025\linewidth}>{\centering}p{0.025\linewidth}
    >{\centering}p{0.025\linewidth}>{\centering}p{0.025\linewidth}
    >{\centering}p{0.025\linewidth}>{\centering}p{0.025\linewidth}
    >{\centering}p{0.025\linewidth}>{\centering}p{0.025\linewidth}
    >{\centering}p{0.025\linewidth}>{\centering}p{0.025\linewidth}
    >{\centering}p{0.025\linewidth}|>{\raggedright}m{0.099\linewidth}}
    \hlinewd{0.8pt}
     & Methods 
     & \begin{sideways}road\end{sideways} & \begin{sideways}sidewalk\end{sideways} 
     & \begin{sideways}building\end{sideways} & \begin{sideways}wall\end{sideways} 
     & \begin{sideways}fence\end{sideways} & \begin{sideways}pole\end{sideways} 
     & \begin{sideways}t-light\end{sideways} & \begin{sideways}t-sign\end{sideways} 
     & \begin{sideways}vegetation\end{sideways} & \begin{sideways}terrain\end{sideways} 
     & \begin{sideways}sky\end{sideways} & \begin{sideways}person\end{sideways} 
     & \begin{sideways}rider\end{sideways} & \begin{sideways}car\end{sideways} 
     & \begin{sideways}truck\end{sideways} & \begin{sideways}bus\end{sideways} 
     & \begin{sideways}train\end{sideways} & \begin{sideways}m-bike\end{sideways} 
     & \begin{sideways}bicycle\end{sideways} & mIoU(\%) \tabularnewline
\hline
\hline
    \multirow{6}[4]{*}{\begin{sideways}Cityscapes\end{sideways}} 
    & Baseline$^\dagger$ & 72.7  & 36.4  & 64.9  & 11.9  & 2.8   & 31.0  & 37.7  & 20.0  & 84.9  & 14.0  & 71.9  & 65.3  & 9.9   & 84.7  & 11.6  & 25.4  & 0.0   & 10.6  & 18.1  
    & 35.46  \tabularnewline
    & IBN-Net$^\dagger$~\cite{pan2018ibnnet} & 68.3  & 29.5  & 69.7  & 17.4  & 1.8   & 30.7  & 36.2  & 20.2  & 85.4  & 18.2  & 81.8  & 64.7  & 12.9  & 82.7  & 13.0  & 16.2  & 0.0   & 8.2   & 22.2  
    & 35.55 ({\textcolor{red}{$0.1$}}) \tabularnewline
    & RobustNet$^\dagger$~\cite{choi2021robustnet} & 82.6  & 40.1  & 73.4  & 17.4  & 1.4   & 34.2  & 38.6  & 18.5  & 84.9  & 16.9  & 81.9  & 65.2  & 11.4  & 84.7  & 7.2   & 23.6  & 0.0   & 10.4  & \textbf{23.9} 
    & 37.69 ({\textcolor{red}{$2.2$}}) \tabularnewline
    \cline{2-22}          
    & \cellcolor{light_gray} Baseline & \cellcolor{light_gray}49.1  & \cellcolor{light_gray}28.0  & \cellcolor{light_gray}69.8  & \cellcolor{light_gray}21.1  & \cellcolor{light_gray}12.2  & \cellcolor{light_gray}21.5  & \cellcolor{light_gray}\textbf{39.3} &\cellcolor{light_gray} 13.0  & \cellcolor{light_gray}81.8  &\cellcolor{light_gray}\textbf{33.7} & \cellcolor{light_gray}68.7  & \cellcolor{light_gray}66.0  & \cellcolor{light_gray}18.2  & \cellcolor{light_gray}38.1  & \cellcolor{light_gray}20.7  & \cellcolor{light_gray}15.6  & \cellcolor{light_gray}3.6   & \cellcolor{light_gray}16.4  & \cellcolor{light_gray}18.4  & \cellcolor{light_gray}33.42 \tabularnewline
    & MLDG$^{\ddagger}$~\cite{mldg} & 75.8  & 37.4  & 78.1  & \textbf{27.6} & 8.5   & \textbf{37.4} & 31.6  & 18.7  & 84.0  & 16.2  & 70.2  & \textbf{66.3} & 16.7  & 74.0  & 20.4  & \textbf{38.4} & 0.0   & 20.4  & 16.1  & 38.84 ({\textcolor{red}{{5.4}}}) \tabularnewline
    
    & \cellcolor{light_gray} \textbf{Ours}  & \cellcolor{light_gray}\textbf{85.3} & \cellcolor{light_gray}\textbf{45.3} & \cellcolor{light_gray}\textbf{82.5} & \cellcolor{light_gray}26.3  & \cellcolor{light_gray}\textbf{19.9} & \cellcolor{light_gray}34.9  & \cellcolor{light_gray}39.0  & \cellcolor{light_gray}\textbf{24.0} & \cellcolor{light_gray}\textbf{85.8} & \cellcolor{light_gray}24.0  & \cellcolor{light_gray}\textbf{82.8} & \cellcolor{light_gray}64.7  & \cellcolor{light_gray}\textbf{21.3} & \cellcolor{light_gray}\textbf{85.7} & \cellcolor{light_gray}\textbf{32.0} & \cellcolor{light_gray}38.2  & \cellcolor{light_gray}\textbf{6.7} & \cellcolor{light_gray}\textbf{26.0} & \cellcolor{light_gray}21.5  & \cellcolor{light_gray}\textbf{44.51 ({\textcolor{red}{\textbf{11.1}}}) } \tabularnewline
    \hline
    \hline
    \multirow{6}[4]{*}{\begin{sideways}BDD100K\end{sideways}}
    & Baseline$^\dagger$ & 44.6  & 26.1  & 34.7  & 1.8   & 6.9   & 29.5  & 39.1  & 20.5  & 64.9  & 10.8  & 51.6  & 50.6  & 10.2  & 63.9  & 1.1   & 4.8   & 0.0   & 5.5   & 10.1  & 25.09 \tabularnewline
    & IBN-Net$^\dagger$~\cite{pan2018ibnnet} & 53.8  & 25.0  & 55.4  & 2.8   & 14.8  & 32.9  & 39.7  & 26.3  & \textbf{71.7} & 16.4  & 85.9  & \textbf{57.4} & \textbf{17.5} & 56.9  & 5.3   & 6.0   & 0.0   & 18.5  & \textbf{25.4} & 32.18 (\textcolor{red}{7.1}) \tabularnewline
  & RobustNet$^\dagger$~\cite{choi2021robustnet} & 69.5  & 35.0  & 60.9  & 4.1   & 13.1  & \textbf{36.6} & \textbf{40.5} & \textbf{27.3} & 71.6  & 14.0  & 83.6  & 56.0  & 17.3  & 61.9  & 4.4   & 8.8   & 0.0   & 24.3  & 18.9  & 34.09 (\textcolor{red}{9.0}) \tabularnewline
\cline{2-22}          & \cellcolor{light_gray} Baseline & \cellcolor{light_gray}54.5  & \cellcolor{light_gray}26.0  & \cellcolor{light_gray}44.0  & \cellcolor{light_gray}3.4   & \cellcolor{light_gray}20.9  & \cellcolor{light_gray}30.1  & \cellcolor{light_gray}37.4  & \cellcolor{light_gray}15.9  & \cellcolor{light_gray}65.7  & \cellcolor{light_gray}22.7  & \cellcolor{light_gray}42.3  & \cellcolor{light_gray}50.9  & \cellcolor{light_gray}14.7  & \cellcolor{light_gray}58.0  & \cellcolor{light_gray}17.5  & \cellcolor{light_gray}14.1  & \cellcolor{light_gray}0.0   & \cellcolor{light_gray}\textbf{25.0} & \cellcolor{light_gray}9.4   & \cellcolor{light_gray}29.07 \tabularnewline
  & MLDG$^{\ddagger}$~\cite{mldg} & 54.0  & 33.4  & 61.0  & \textbf{6.4} & 25.3  & 35.5  & 35.5  & 19.0  & 71.5  & 20.0  & 75.8  & 53.7  & 13.4  & 46.2  & 7.3   & \textbf{34.4} & 0.0   & 9.5   & 5.3   & 31.95 (\textcolor{red}{2.9}) \tabularnewline
  &\cellcolor{light_gray} \textbf{Ours}  & \cellcolor{light_gray}\textbf{79.3} & \cellcolor{light_gray}\textbf{39.1} & \cellcolor{light_gray}\textbf{69.0} & \cellcolor{light_gray}6.2   & \cellcolor{light_gray}\textbf{32.8} & \cellcolor{light_gray}32.1  & \cellcolor{light_gray}36.7  & \cellcolor{light_gray}26.9  & \cellcolor{light_gray}71.3  & \cellcolor{light_gray}\textbf{25.9} & \cellcolor{light_gray}\textbf{86.3} & \cellcolor{light_gray}49.4  & \cellcolor{light_gray}12.5  & \cellcolor{light_gray}\textbf{75.2} & \cellcolor{light_gray}\textbf{20.6} & \cellcolor{light_gray}31.6  & \cellcolor{light_gray}0.0   & \cellcolor{light_gray}17.9  & \cellcolor{light_gray}10.7  & \cellcolor{light_gray}\textbf{38.07 (\textcolor{red}{9.0})} \tabularnewline
    \hline
    \hline
    \multirow{6}[4]{*}{\begin{sideways}Mapillary\end{sideways}} & Baseline$^\dagger$ & 62.0  & 36.3  & 32.5  & 9.5   & 7.7   & 29.9  & 40.5  & 22.5  & 78.6  & \textbf{40.9} & 61.0  & 59.4  & 6.4   & 78.3  & 5.1   & 5.1   & 0.1   & 9.0   & 21.8  & 31.94 \tabularnewline
  & IBN-Net$^\dagger$~\cite{pan2018ibnnet} & 67.4  & 38.8  & 51.3  & 10.2  & 7.6   & 36.0  & 40.1  & 40.8  & \textbf{80.3} & 39.9  & \textbf{92.1} & 61.8  & 14.0  & 74.4  & 10.7  & 9.4   & 3.5   & 15.3  & \textbf{25.4} & 38.09 (\textcolor{red}{6.2}) \tabularnewline
  & RobustNet$^\dagger$~\cite{choi2021robustnet} & 78.0  & \textbf{41.0} & 56.6  & 13.1  & 6.2   & \textbf{39.4} & \textbf{41.3} & 36.1  & 79.5  & 34.7  & 90.0  & 61.0  & 12.0  & 76.1  & 10.7  & 13.1  & 0.8   & 16.9  & 24.8  & 38.49 (\textcolor{red}{6.6}) \tabularnewline
\cline{2-22}          & \cellcolor{light_gray} Baseline & \cellcolor{light_gray}53.4  & \cellcolor{light_gray}25.9  & \cellcolor{light_gray}44.7  & \cellcolor{light_gray}11.1  & \cellcolor{light_gray}19.0  & \cellcolor{light_gray}28.4  & \cellcolor{light_gray}36.2  & \cellcolor{light_gray}15.8  & \cellcolor{light_gray}71.3  & \cellcolor{light_gray}27.1  & \cellcolor{light_gray}66.1  & \cellcolor{light_gray}58.6  & \cellcolor{light_gray}11.7  & \cellcolor{light_gray}64.2  & \cellcolor{light_gray}20.1  & \cellcolor{light_gray}1.1   & \cellcolor{light_gray}\textbf{11.4} & \cellcolor{light_gray}\textbf{23.1} & \cellcolor{light_gray}22.3  & \cellcolor{light_gray}32.19 \tabularnewline
  & MLDG$^{\ddagger}$~\cite{mldg} & 69.4  & 36.0  & 58.6  & \textbf{19.4} & 16.8  & 37.6  & 31.3  & 28.8  & 76.7  & 36.9  & 81.6  & 43.4  & 15.5  & 59.1  & 21.4  & 8.1   & 1.3   & 16.8  & 17.9  & 35.60 (\textcolor{red}{3.7}) \tabularnewline
  &\cellcolor{light_gray} \textbf{Ours}  & \cellcolor{light_gray}\textbf{78.0} & \cellcolor{light_gray}40.8  & \cellcolor{light_gray}\textbf{71.1} & \cellcolor{light_gray}14.6  & \cellcolor{light_gray}\textbf{27.0} & \cellcolor{light_gray}34.2  & \cellcolor{light_gray}40.7  & \cellcolor{light_gray}\textbf{50.3} & \cellcolor{light_gray}77.1  & \cellcolor{light_gray}26.2  & \cellcolor{light_gray}90.0  & \cellcolor{light_gray}\textbf{63.1} & \cellcolor{light_gray}\textbf{24.0} & \cellcolor{light_gray}\textbf{81.6} & \cellcolor{light_gray}\textbf{30.5} & \cellcolor{light_gray}\textbf{15.5} & \cellcolor{light_gray}5.3   & \cellcolor{light_gray}18.7  & \cellcolor{light_gray}22.7  & \cellcolor{light_gray}\textbf{42.70 (\textcolor{red}{10.5})} \tabularnewline
\hlinewd{0.8pt}
\end{tabular}
}
    \vspace{-8pt}
    \caption{\textbf{Source (G+S)$\rightarrow$Target (C, B, M):} Mean IoU(\%) and per-class IoU(\%) comparison of other SOTA DG methods for semantic segmentation. We report the mIoU improvement as red text. The networks are DeepLabV3+ with ResNet50 and results with $^\dagger$ are from~\cite{choi2021robustnet}.}
    \label{tab:gsdg}
    \vspace{-10pt}
\end{table*}%

\section{Experiments} \vspace{-3pt}
\label{sec:experiments}

\subsection{Experimental Setup}
    \vspace{-10pt}  
    \paragraph{Datasets}
    We conduct the experiments on six different datasets to prove the generalization ability of our method.\vspace{-7pt}
    \begin{itemize}[leftmargin=*]
        \item \textbf{Real datasets:} \textbf{C}ityscapes~\cite{cordts2016cityscapes} includes 3,450 finely-annotated images collected from 50 different cities, primarily Germany. We use only a finely-annotated set for training and validation.
        \textbf{B}DD100K~\cite{yu2020bdd100k} contains 8K diverse urban driving scene images collected from various locations in the US.
        \textbf{M}apillary~\cite{neuhold2017mapillary} is a real street-view dataset including 25K images collected from all around the world.
        \textbf{I}DD~\cite{varma2019idd} contains 10,004 images captured from Indian roads.
        The road scenes in the IDD, which contain animals and muddy, are significantly different from the existing datasets mainly collected in Europe or US.
        \vspace{-7pt}
        
        \item \textbf{Synthetic datasets:} 
        \textbf{G}TA\RomanNumeralCaps{5}~\cite{gtav} includes 24,966 driving-scene images generated from a game engine. It has 19 object categories compatible with the real-world datasets.
        \textbf{S}ynthia~\cite{synthia} is another synthetic dataset simulating different seasons, weather, and illumination conditions from multiple viewpoints. The Synthia dataset contains 9,400 photo-realistic synthetic images annotated into 16 categories compatible with the GTA\RomanNumeralCaps{5}.
    \end{itemize}
    
    \vspace{-13pt}
    \paragraph{Metrics} Following the standard-setting~\cite{choi2021robustnet, he2021multi}, we report mean Intersection over Union (mIoU) averaged over all classes to measure the segmentation performance.

    \vspace{-10pt}
    \paragraph{Implementation details}
    We conducted experiments by adopting DeepLabV3+~\cite{deeplabv3p} and DeepLabV2~\cite{deeplabv2} with ResNet50 and ResNet101~\cite{resnet} as a semantic segmentation architecture, respectively,
    where the output stride is 16 for DeepLabV3+.
    All backbones were initialized with ImageNet~\cite{2015imagenet} pre-trained model.
    We set the maximum iterations to 120K but early stop at 30K iterations, except for ResNet-101 models trained for 70K.
    The hyper-parameters, $\lambda_{1}$ and $\lambda_{2}$, were empirically set to 0.02 and 0.2.
    Further details for optimization and training are explained in the supplementary material.
    In all experiments, we denote the networks trained with aggregated source domains as a \textit{baseline}.
    To conduct experiments, we re-implemented several DG methods and marked them with $^\ddagger$.

\vspace{-3pt}
\subsection{Results}
\vspace{-3pt}


\begin{figure*}[!t]
    \centering
      \begin{subfigure}{0.1425\linewidth}
          \includegraphics[width=1.0\linewidth]{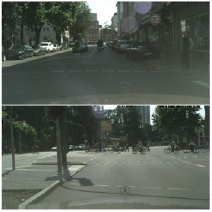}\hfill
          \caption{Images}
      \end{subfigure}
      \hspace{-5pt}
      \begin{subfigure}{0.1425\linewidth}
          \includegraphics[width=1.0\linewidth]{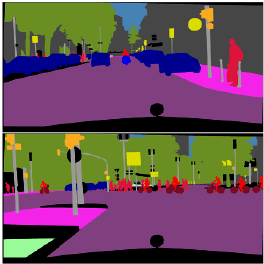}\hfill
          \caption{Ground Truth}
      \end{subfigure}
      \hspace{-5pt}
      \begin{subfigure}{0.1425\linewidth}
          \includegraphics[width=1.0\linewidth]{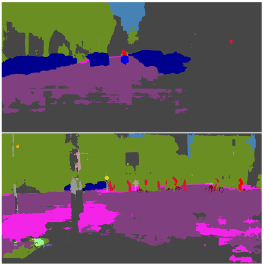}\hfill
          \caption{Baseline}
      \end{subfigure}
      \hspace{-5pt}
      \begin{subfigure}{0.1425\linewidth}
          \includegraphics[width=1.0\linewidth]{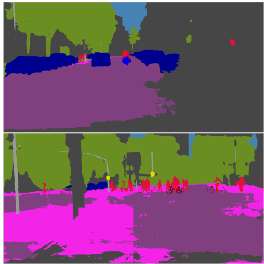}\hfill
          \caption{IBN-Net~\cite{pan2018ibnnet}}
      \end{subfigure}
      \hspace{-5pt}
      \begin{subfigure}{0.1425\linewidth}
          \includegraphics[width=1.0\linewidth]{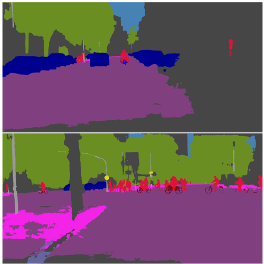}\hfill
          \caption{RobustNet~\cite{choi2021robustnet}}
      \end{subfigure}
      \hspace{-5pt}
      \begin{subfigure}{0.1410\linewidth}
          \includegraphics[width=1.0\linewidth]{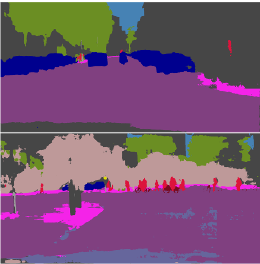}\hfill
          \caption{MLDG~\cite{mldg}}
      \end{subfigure}
      \hspace{-5pt}
      \begin{subfigure}{0.1425\linewidth}
          \includegraphics[width=1.0\linewidth]{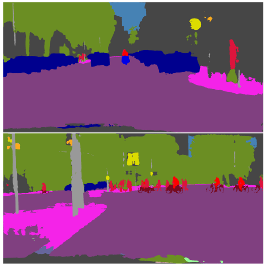}\hfill
          \caption{Ours}
      \end{subfigure}
      \\
       \vspace{-7pt}
 	\caption{\textbf{Source (G+S)$\rightarrow$Target (C):} Qualitative comparison on the Cityscapes dataset. All methods adopt DeepLabV3+ with ResNet50. (Best viewed in color.)}\label{fig:qualcity}\vspace{-16pt}
    \label{fig:city_qual}
\end{figure*}

\begin{table}[t]
  \centering
    \resizebox{1\linewidth}{!}{
    \begin{tabular}{l|c|c|c|c}
    \hlinewd{0.8pt}
    Methods & Cityscapes     & \multicolumn{1}{c|}{BDD100K}  &  \multicolumn{1}{c|}{Mapillary}     & Avg. \\
    \hline
    \hline
    Baseline & 52.51  & 47.47  & 54.70  & 51.56  \\
    IBN-Net$^\ddagger$~\cite{pan2018ibnnet}  & 54.39   & 48.91   & 56.06   & 53.12 \\
    RobustNet$^\ddagger$~\cite{choi2021robustnet}  & 54.70  & 49.00  & 56.90  & 53.53  \\
    MLDG$^\ddagger$~\cite{mldg}  & 54.76  & 48.52  & 55.94  & 53.07 \\
    TSMLDG$^\ddagger$~\cite{tsmldg} & 53.02  & 46.43  & 52.76  & 50.70 \\
    \rowcolor{light_gray}
    \textbf{Ours}   & \textbf{56.57} & \textbf{50.18} & \textbf{58.31} & \textbf{55.02} \\
    \hlinewd{0.8pt}
    \end{tabular}}
    \vspace{-8pt}
    \caption{\textbf{Source (G+S+I)$\rightarrow$Target (C, B, M):} 
    Mean IoU(\%) comparison of other state-of-the-art DG methods, where all networks are trained with two synthetic (GTA\RomanNumeralCaps{5}, Synthia) and one real (IDD) datasets.  All methods adopt DeepLabV3+ with ResNet50.}
    \vspace{4pt}
  \label{tab:gsidg}
\end{table}%

\vspace{-10pt}
\paragraph{Comparison with state-of-the-art}
\tabref{tab:gsdg} summarizes the test results on the most popular real-world dataset benchmarks, where the models were trained on multi-source domains (GTA\RomanNumeralCaps{5} and Synthia).
We conduct comparisons with the re-implemented vanilla meta-learning method without the memory module (MLDG) and the normalization-based methods (IBN-Net and RobustNet) based on the results reported in the paper~\cite{choi2021robustnet}.
While the existing normalization-based methods are slightly better than the baseline performance, our approach consistently outperforms the state-of-the-arts (SOTA) by a large margin on all real-world datasets.
It demonstrates that the generalization methods by erasing the visual style of domains makes it hard to leverage multi-source domain information well.
Especially, our approach shows significantly improved gain over baseline as $11.1\%$ on the Cityscapes, $9.0\%$ on the BDD100K and $10.5\%$ on the Mapillary.
Furthermore, compared to MLDG~\cite{mldg} that uses meta-learning framework like ours, our method proves the effectiveness of the categorical memory to improve the generalization capability.
\figref{fig:qualcity} shows qualitative results and more results are provided in supplementary materials.
To further verify the performance variation when more source data is used, we add one more real dataset (IDD) to the source domain following~\cite{tsmldg}.
Since the IDD dataset significantly differs from the existing real datasets, this scenario assumes a new generalization scenario where the available real dataset looks very different from the target domain's culture.
In \tabref{tab:gsidg}, we can see that our method also outperforms all previous approaches in this setting.

\begin{table}[!t]
  \centering
    \resizebox{1\linewidth}{!}{
        \begin{tabular}{cl|c|cl|c}
    \hlinewd{0.8pt}
    & Methods & Cityscapes &  & Methods & Cityscapes \\
    \hline
    \hline
    \multirow{6}{*}{\begin{sideways}FCN-8s\end{sideways}} & Baseline & 32.5  & \multirow{6}{*}{\begin{sideways}DeepLabV3+\end{sideways}} & Baseline$^\dagger$ & 29.0 \\
          & DRPC~\cite{yue2019domain} & 37.4  &       & IBN-Net$^\dagger$~\cite{pan2018ibnnet} & 33.9 \\
\cline{2-3}          & Baseline & 21.4  &       & RobustNet$^\dagger$~\cite{choi2021robustnet} & 36.6 \\
\cline{5-6}          & CNSN~\cite{cnsn} & 36.5  &       &  \cellcolor{light_gray} Baseline & \cellcolor{light_gray} 31.6 \\
\cline{2-3}          & Baseline & 23.3  &       & MLDG~\cite{mldg} & 36.7 \\
          & ASG~\cite{asg} & 31.9  &       &  \cellcolor{light_gray} \textbf{Ours}  & \cellcolor{light_gray} \textbf{41.0} \\
    \hlinewd{0.8pt}
    \end{tabular}
    }
    \vspace{-7pt}
    \caption{\textbf{Source (G)$\rightarrow$Target (C):} Mean IoU(\%) comparison of other SOTA methods with various segmentation models with ResNet50.
    Results with $^\dagger$ are from~\cite{choi2021robustnet}.
    The results on other datasets are reported in supplementary materials.}
    \vspace{-10pt}
  \label{tab:gdgc}
\end{table}

\begin{table}[t]
  \centering
    \resizebox{\linewidth}{!}{
    \begin{tabular}{
    >{\raggedright}m{0.3\linewidth}|>{\centering}m{0.17\linewidth}|
    >{\centering}m{0.19\linewidth}|>{\centering}m{0.19\linewidth}}
    \hlinewd{0.8pt}
    Methods & w/Target & Cityscapes     & BDD100K \tabularnewline
    \hline
    \hline
    Baseline & \xmark & 40.0  & 37.4  \tabularnewline
    CyCADA~\cite{hoffman2018cycada}$^\dagger$ & \cmark & 39.3 & 37.2  \tabularnewline
    MDAN~\cite{zhao2018mdan}$^\dagger$ & \cmark & 36.0 & 29.4  \tabularnewline
    MADAN~\cite{zhao2019multi}$^\dagger$ & \cmark & 45.4 & 40.4  \tabularnewline
    MADAN+~\cite{zhao2021madan} & \cmark & 48.5 & 42.7 \tabularnewline
    CLSS~\cite{he2021multi} & \cmark & \textbf{54.0} & N/A \tabularnewline
    \rowcolor{light_gray}
    \textbf{Ours}  & \xmark &  49.4 & \textbf{45.5}  \tabularnewline
    \hlinewd{0.8pt}
    \end{tabular}
    }
    \vspace{-7pt}
    \caption{\textbf{Source (G+S)$\rightarrow$Target (C, B):} Mean IoU(\%) comparison of other multi-source UDA methods. The segmentation models are all DeepLabV2 with ResNet101. Results with $^\dagger$ are from~\cite{zhao2021madan}.}
    \vspace{4pt}
  \label{tab:gsuda} 
\end{table}%

\tabref{tab:gdgc} shows the results evaluated on the Cityscapes dataset  with various segmentation models regarding to single-source domain generalization setting.
Like~\cite{choi2021robustnet}, we generate virtual domain shift by photometric transformations such as Gaussian blur or color jitter in this setting.
Even though the networks are trained on the GTA\RomanNumeralCaps{5} dataset only, our method obtains the best generalization performance with a relatively high-performance gain.
It thus points out that category-aware generalization, like our method, should be encouraged importantly to further research in this area.

\vspace{-10pt}
\paragraph{Comparison with UDA}
We also compared our result with state-of-the-art UDA methods~\cite{zhao2018mdan,zhao2021madan,zhao2019multi,he2021multi} trained on multiple synthetic datasets.
For a fair comparison, we reported mIoU score over 16 object classes in \tabref{tab:gsuda}.
All other methods excluding baseline and ours used training images from target domain to learn their models.
It is interesting that even though UDA is a much easier setting than domain generalization, our DG method achieved the highest performance on the BDD100K.
Except for CLSS~\cite{he2021multi}, our method also showed competitive results on the Cityscapes.
We argue that our method, guided by domain-agnostic categorical memory learned from multi-source domains, is more effective to deal with diverse real-world cases than UDA methods.

\vspace{-3pt}
\subsection{Ablation Study and Discussion} \vspace{-3pt}
All experiments in this subsection uses the model trained on GTA\RomanNumeralCaps{5} and Synthia datasets adopting DeepLabV3+~\cite{deeplabv3p} with ResNet50~\cite{resnet}.

\begin{table}[!t]
  \centering
  \resizebox{0.85\linewidth}{!}{
    \begin{tabular}{cc|c|c|c|c}
    \hlinewd{0.8pt}
    $\loss{\text{coh}}$ & $\loss{\text{div}}$  & Cityscapes     & BDD100K     & Mapillary     & Avg. \\
    \hline
    \hline
    \rowcolor{light_gray}
    \cmark &        &  43.85 & 38.01 & 41.66 & 41.17 \\
           & \cmark &  42.08 & 36.80 & 40.83 & 39.90 \\
    \rowcolor{light_gray}
    \rowcolor{light_gray}
    \cmark & \cmark & \textbf{44.51} & \textbf{38.07} & \textbf{42.70} & \textbf{41.76} \\
    \hlinewd{0.8pt}
    \end{tabular}}
    \vspace{-7pt}
  \caption{Ablation study on the variants of loss.}
  \label{tab:loss}%
  \vspace{-10pt}
\end{table}%
\begin{table}[!t]
    \centering
    \resizebox{0.95\linewidth}{!}{
        \begin{tabular}{c|cc|c|c|c|c}
        \hlinewd{0.8pt}
        \multirow{2}{*}{Memory}     & \multicolumn{2}{|c|}{Freeze} \vspace{1pt}& \multirow{2}{*}{Cityscapes} & \multirow{2}{*}{BDD100K}   & \multirow{2}{*}{Mapillary}   & \multirow{2}{*}{Avg.} \tabularnewline
        \cline{2-3}
            &  $E$ &  $U$ &      &      &      &  \tabularnewline
        \hline
        \hline
        \rowcolor{light_gray}
        $\hat{\mathcal{M}}$ & -     & -     &   40.64 &  31.06   &  31.59  & 34.43 \\
        $\mathcal{M}'$      & \xmark & \xmark & 41.65 & 36.56 & 38.80 & 39.00 \\
        \rowcolor{light_gray}
        $\mathcal{M}'$      & \cmark & \xmark & \textbf{44.51} & \textbf{38.07} & \textbf{42.70} & \textbf{41.76} \\
        $\mathcal{M}'$      & \cmark & \cmark & 41.67 & 32.04 & 33.90 & 35.87 \\
        \hlinewd{0.8pt}
        \end{tabular}}
        \vspace{-7pt}
    \caption{Ablation study on the variants of the memory update strategy in meta-testing step.}
    \vspace{-10pt}
    \label{tab:updatememory}
\end{table}
\begin{table}[!t]
  \centering
  \resizebox{1\linewidth}{!}{
    \begin{tabular}{
    >{\centering}m{0.12\linewidth}>{\centering}m{0.13\linewidth}
    >{\centering}m{0.07\linewidth}|>{\centering}m{0.17\linewidth}
    |>{\centering}m{0.17\linewidth}|>{\centering}m{0.16\linewidth}
    |>{\centering}m{0.09\linewidth}
    }
    \hlinewd{0.8pt}
    Training & Memory & M.L.  & Cityscapes     & BDD100K     & Mapillary     & Avg. \tabularnewline
    \hline
    \hline
    \rowcolor{light_gray}
    Agg.    & \xmark & \xmark & 33.42  & 29.07  & 31.90  & 31.46  \tabularnewline
    Agg.    & \cmark & \xmark & 38.28  & 31.46  & 32.25  & 34.00  \tabularnewline
    \rowcolor{light_gray}
    Episodic & \xmark & \cmark & 38.84  & 31.95  & 35.60  & 35.46  \tabularnewline
    Episodic & \cmark & \xmark & 41.50  & 38.00  & 40.22  & 39.91  \tabularnewline
    \rowcolor{light_gray}
    Episodic & \cmark & \cmark & \textbf{44.51} & \textbf{38.07} & \textbf{42.70} & \textbf{41.76} \tabularnewline
    \hlinewd{0.8pt}
    \end{tabular}}
    \vspace{-7pt}
  \caption{Ablation study on the variants of the memory learning method. We denote the baseline as `Agg.', episodic training as `Episodic' and the second-order gradient as `M.L.'.}
  \vspace{4pt}
  \label{tab:mlframework}
\end{table}

\noindent\textbf{Loss.}
To verify the effectiveness of the proposed losses, we study different loss combinations of $\mathcal{L}_\text{coh}$ and $\mathcal{L}_\text{div}$.
From \tabref{tab:loss}, we observe that both loss terms make substantial contributions to the performance gain for all datasets by operating complementary each other.

\noindent\textbf{Memory update strategy.}
In \eqref{eq:memsubupdate} of the meta-testing step, we freeze the encoder parameters and re-update the memory.
To show the effectiveness of this update scheme, we conduct an ablation study with the combination for the encoder $E$ and memory updating network $U$ in \tabref{tab:updatememory}.
Without re-updating the memory in the meta-testing step, a severe generalization performance drop occurs since the updating network is not updated.
Additionally, a similar degradation occurs when both encoder and the updating networks are frozen, because the generalization objective for memory updating is not evenly set.
This means that our novel step is operated for memorizing domain-agnostic features while stabilizing meta-optimization.

\begin{table}[!t]
    \centering
    \resizebox{0.85\linewidth}{!}{
        \begin{tabular}{lccc}
        \hlinewd{0.8pt}
        Methods     & \# of Params     & GFLOPs     & Time (ms) \\
        \hline
        \hline
        \rowcolor{light_gray}
        Baseline                            & 45.08M & 277.16 & 13.51 \\
        IBN-Net$^\ddagger$~\cite{pan2018ibnnet}        & 45.08M & 277.24 & 14.31 \\
        \rowcolor{light_gray}
        RobustNet$^\ddagger$~\cite{choi2021robustnet}  & 45.08M & 277.20 & 14.53 \\
        MLDG$^\ddagger$~\cite{mldg}                    & 45.08M & 277.16 & 13.85 \\
        \rowcolor{light_gray}
        \textbf{Ours} & 45.22M & 277.69 & 13.56 \\
        \hlinewd{0.8pt}
        \end{tabular}}
        \vspace{-7pt}
  \caption{Comparison of computational cost. Tested with the image size of 2048$\times$1024 on NVIDIA TITAN RTX GPU. We averaged the inference time over 500 trials.}
  \label{tab:cost}
  \vspace{-10pt}
\end{table}

\begin{figure}[t]
\begin{center}
    \centering
    \begin{subfigure}{1.0\linewidth}
        \raisebox{12pt}{\rotatebox[origin=c]{90}{\footnotesize{Label}}\hspace{3pt}}
        \includegraphics[width=0.9\linewidth]{figure/tsne_label.pdf}
    \end{subfigure}
    \hfill
    \\
    \vspace{-1pt}
    \centering
    \begin{subfigure}{1.0\linewidth}
        \raisebox{52pt}{\rotatebox[origin=c]{90}{\footnotesize{RobustNet~\cite{choi2021robustnet}}}\hspace{1pt}}
        \includegraphics[width=0.9\linewidth]{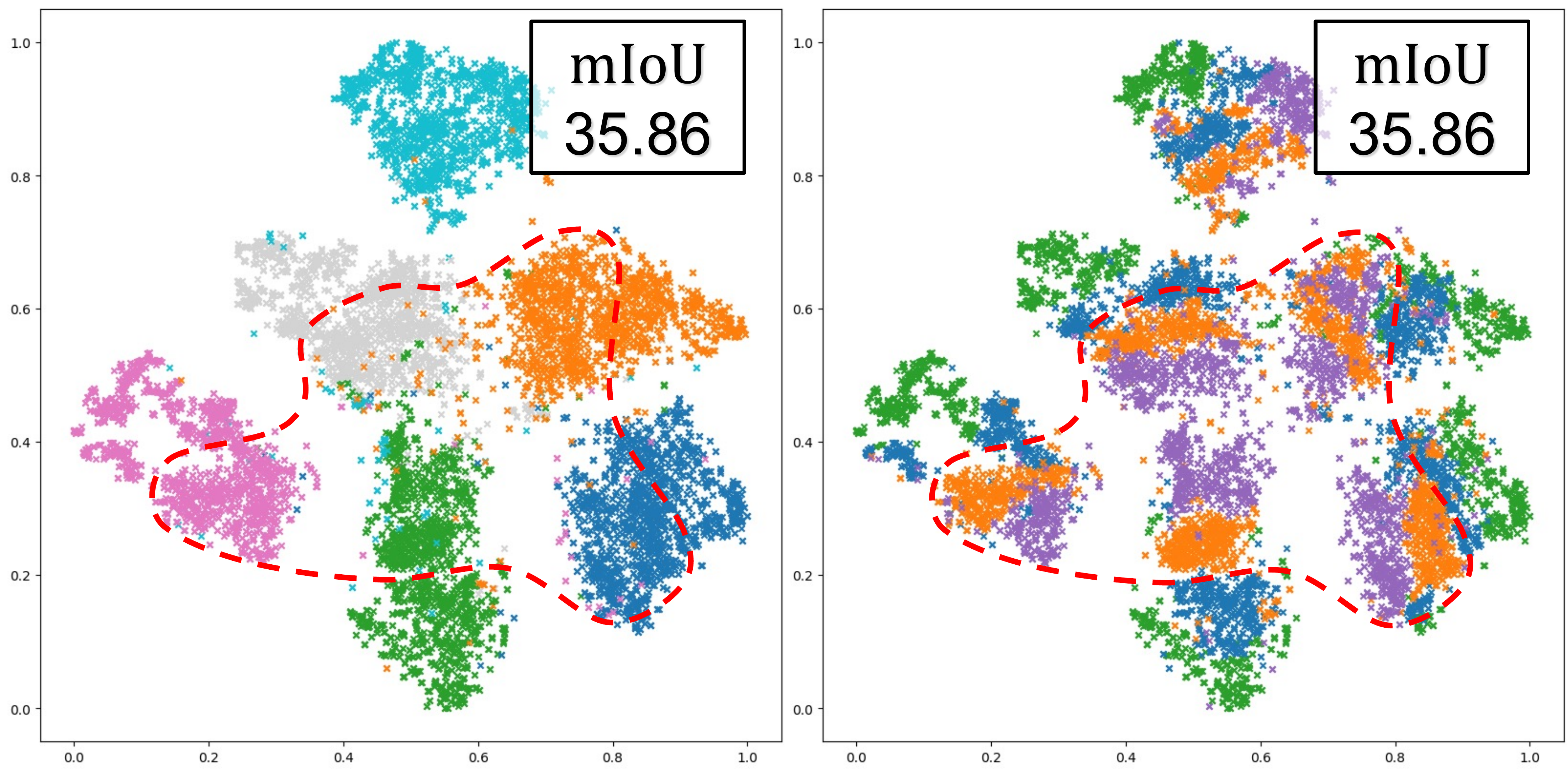}
    \end{subfigure}
    \hfill
    \\
    \vspace{-1pt}
    \centering
    \begin{subfigure}{1.0\linewidth}
        \raisebox{52pt}{\rotatebox[origin=c]{90}{\footnotesize{Ours}}\hspace{1.6pt}}
        \includegraphics[width=0.9\linewidth]{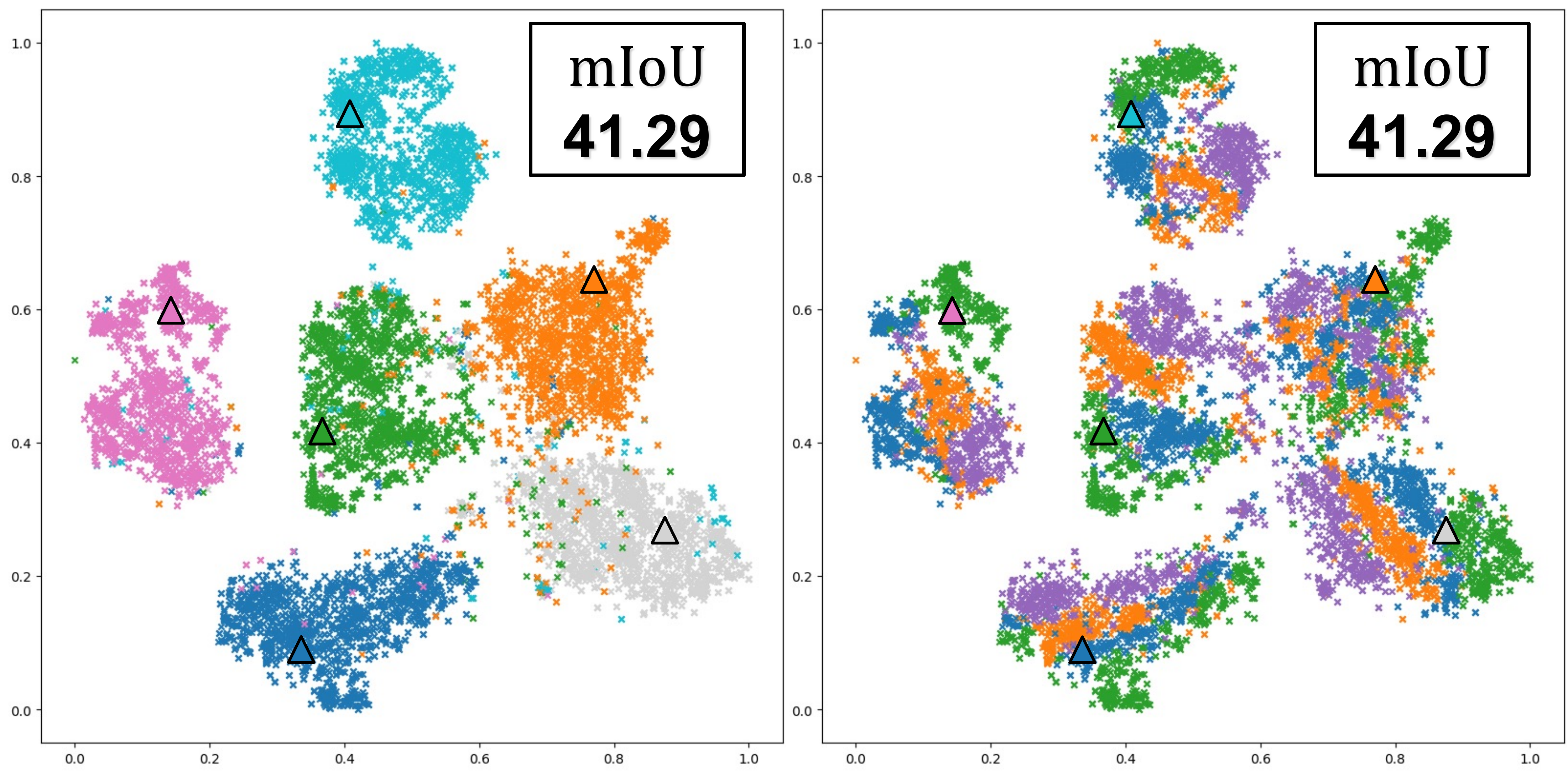}
    \end{subfigure}
    \hfill
    \\
    \vspace{-8pt}
    \caption{\textbf{Source (G+S)$\rightarrow$Target (C, B):} t-SNE visualization of extracted features. Colors indicate different categories in the first column and different domains in the second column. Memory is pointed as triangle. The mIoU scores are the average scores of the target domains.}
    \label{fig:tsne} \vspace{-3pt}
\end{center}
\end{figure}

\noindent\textbf{Memory learning framework.}
We analyze our method by dividing it into three key contributing factors: memory, second-order gradient (`M.L.'~\cite{maml} in \tabref{tab:mlframework}) and episodic learning without `M.L.' that creates an artificial domain shift environment.
In \tabref{tab:mlframework}, for the memory with episodic learning, we perform memory update on $\mathbb{S}_\text{mtr}$ and memory reading on $\mathbb{S}_\text{mte}$.
In this process, we do not calculate the second-order gradient as no meta-testing step.
To sum up, our method enhances the generalization capability most effectively with the combination of all these variants.

\noindent\textbf{Visualization of t-SNE.}
In \figref{fig:tsne}, we visualize the pixel representations with the models learned with our method and normalization based DG method~\cite{choi2021robustnet}.
In the second column, we can see that the unseen domain features of RobustNet  tend to agglomerate with each other. 
In contrast, our method significantly reduces the tendency that features belonging to the same domain but of different classes aggregate with each other, especially in pole class.
At the same time, our method shows superior generalization performance than RobustNet. 
It demonstrates that our method effectively integrates source domain information by generalized categorical knowledge.

\noindent\textbf{Running time and complexity.}
In \tabref{tab:cost} we compare with exisiting DG methods in respect of the computational cost.
Since our method requires memory, the few amount of parameters increases. However, the inference time is competitive to other methods.
Therefore, it can be said that our memory module is cost-effective with a high generalization score gain compared to the cost it occupies. 

\begin{figure}[t]
      \centering
      \begin{subfigure}{0.3269\linewidth}
      \centering
          \includegraphics[width=1.0\linewidth]{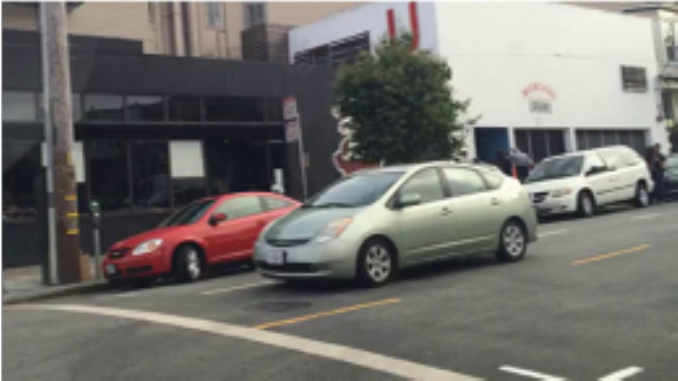}\label{fig:memacta}\hfill
          \caption{Input}
      \end{subfigure}
      \hspace{-4pt}
      \begin{subfigure}{0.3269\linewidth}
      \centering
          \includegraphics[width=1.0\linewidth]{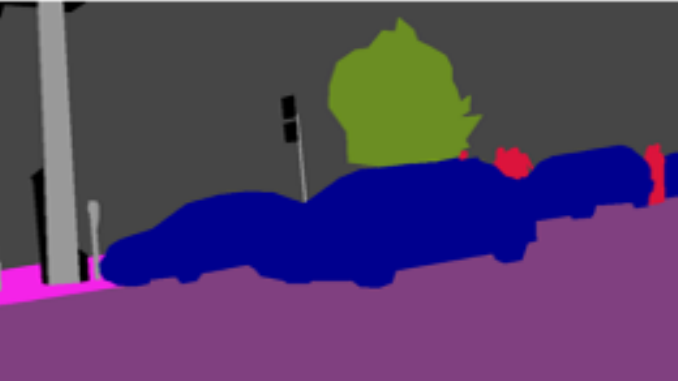}\label{fig:memactb}\hfill
          \caption{Ground Truth}
      \end{subfigure}
      \hspace{-4pt}
      \begin{subfigure}{0.3269\linewidth}
      \centering
          \includegraphics[width=1.0\linewidth]{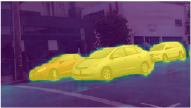}\label{fig:memactc}\hfill
          \caption{Car}
      \end{subfigure}
      \vspace{-1pt}
      \\
      \centering
      \begin{subfigure}{0.329\linewidth}
      \centering
          \includegraphics[width=1.0\linewidth]{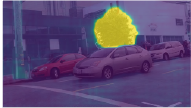}\label{fig:memactd}\hfill
          \caption{Vegetation}
      \end{subfigure}
      \hspace{-4.5pt}
      \begin{subfigure}{0.329\linewidth}
      \centering
          \includegraphics[width=1.0\linewidth]{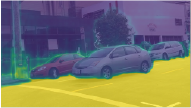}\label{fig:memacte}\hfill
          \caption{Road}
      \end{subfigure}
      \hspace{-5.3pt}
      \begin{subfigure}{0.332\linewidth}
      \centering
          \includegraphics[width=1.0\linewidth]{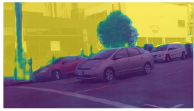}\label{fig:memactf}\hfill
          \caption{Building}
      \end{subfigure}
      \hspace{1pt}
      \\
      \vspace{-6pt}
      \caption{\textbf{Source (G+S)$\rightarrow$Target (B):} Visualization of the channels of memory weight matrix from \eqref{eq:readweight} with the BDD100K dataset.
      } \label{fig:memact} \vspace{7pt}
\end{figure}

\noindent\textbf{Visualization of memory activation.}
As illustrated in \figref{fig:memact}, we visualize the memory weight for the input image from the unseen domain.
We can see that regions are activated by a memory slot corresponding to each class.

\vspace{-3pt}

\section{Conclusion and Future Work}\label{sec:conclusion}
\vspace{-5pt}
We have presented the memory-guided meta-learning framework for robust semantic segmentation regardless of domain shift, with the novel memory divergence and feature cohesion losses.
The ablation studies clearly demonstrate the effectiveness of each component and loss in our method.
Finally, we have demonstrate that our method significantly outperforms other methods in domain generalization settings and also show competitive performance with domain adaptation methods.
However, current DG methods (including this work) for semantic segmentation have limited to the assign the pixel corresponding to the underlying closed-set classes.
To expand this work to a more practical scenario, we should consider the open-set segmentation which can be an appealing topic for DG in semantic segmentation.
We remain a plethora of avenues for this as future work.

{\small
\bibliographystyle{ieee_fullname}
\bibliography{egbib}
}

\clearpage
\clearpage
\newpage

\section*{Appendix}

\appendix

In this document, we describe second-order gradient flow of our method and details of experiments, and provide additional ablation study for analysis of memory update.
Moreover, we complement qualitative and quantitative comparisons to state-of-the-art methods.

\section{Second-Order Gradient Flow}
\begin{figure}[!t]
	\centering
    {\includegraphics[width=1\linewidth]{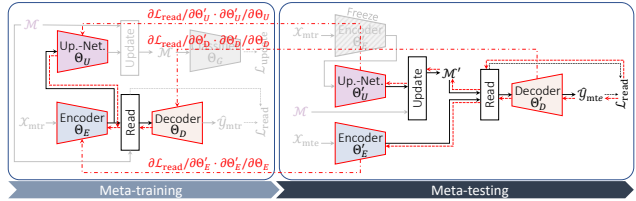}}
    \vspace{-15pt}
 	\caption{Illustration of the gradient flow (red dotted lines) in the optimization of meta-testing step.
 	}\label{fig:gradientflow}
 	\vspace{3pt}
 \end{figure}

In \figref{fig:gradientflow}, we depict the gradient flow of the optimization in the meta-testing step.
In this process, we compute the gradient of the original parameters $\lbrace\Theta\rbrace_{E,U,D}$ for the meta-testing loss and generate the second-order gradients by differentiating the parameters  $\lbrace\Theta\rbrace'_{E,U,D}$ used in the meta-testing step with the original parameters.
These second-order gradients make the original parameters learn to (1) write the domain-independent features to the current memory $\mathcal{M}$ from the meta-train image and (2) ensure the generalization ability of the memory-guided feature for the meta-test image.
\section{Implementation Details}
\subsection{Data Split and Augmentation}
The batch size per domain was 4 for multi-source domain training and 8 for single-source domain training.
Following the setting from RobustNet~\cite{choi2021robustnet}, standard augmentations such as color jittering (brightness of 0.4, contrast of
0.4, saturation of 0.4, and hue of 0.1), Gaussian blur, random cropping, random horizontal flipping, and random scaling with the range of [0.5, 2.0] were conducted to prevent the model from overfitting.
To create an artificial domain shift even in a single source domain generalization setting, we applied higher intensity random color jittering (brightness of 0.8, contrast of
0.8, saturation of 0.8, and hue of 0.3) and Gaussian blur only to the images used in the meta-testing step.

\subsection{Training and Optimization}
We implemented our approach with PyTorch and conducted experiments by adopting DeepLabV3+~\cite{deeplabv3p} with ResNet-50~\cite{resnet} backbone network.
The output stride of DeepLabV3+ was set to 16 and adopted the auxiliary per-pixel cross-entropy loss proposed in PSPNet~\cite{zhao2017pyramid} with a coefficient of 0.4 to make a fair comparison with the normalization based DG method~\cite{choi2021robustnet}.
We performed memory operation using the feature map of 256 channel dimensions after the ASPP~\cite{deeplabv3p} module to leverage the multiple receptive fields and reduce GPU memory usage.
We also adopted DeepLabV2~\cite{deeplabv2} with ResNet-101 for a fair comparison with multi-source unsupervised domain adaptation methods.
For all the experiment, we initialized backbones with ImageNet~\cite{2015imagenet} pre-trained model.
The optimizer was SGD with momentum of 0.9. The learning rate of the meta-testing step $\beta$ was 1e-2 initially and decreased with exponential learning rate policy with the gamma of 9.
The learning rate of the meta-training step $\alpha$ was set to 1/4 of the outer learning rate $\beta$ to stabilize the gradient-based meta optimization~\cite{maml,antoniou2018train}.
We set the maximum iterations to 120K but early stop at 30K iterations, except for ResNet-101 models trained for 70K.
The coefficients of memory divergence loss and feature cohesion loss, $\lambda_{1}$ and $\lambda_{2}$, was set to 0.02 and 0.2, respectively.

\begin{figure*}[!ht]
    \centering
    \hspace{4pt}
      \includegraphics[width=0.9\linewidth]{figure/classlabel.pdf}
      \includegraphics[width=0.03\linewidth]{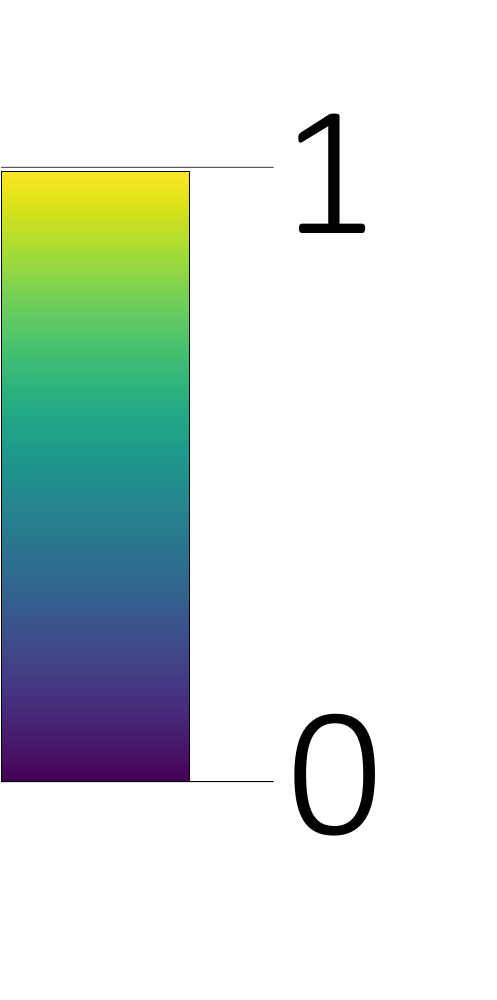}
    \\
    \centering
    \begin{subfigure}{0.161\linewidth}
      \includegraphics[width=1.0\linewidth]{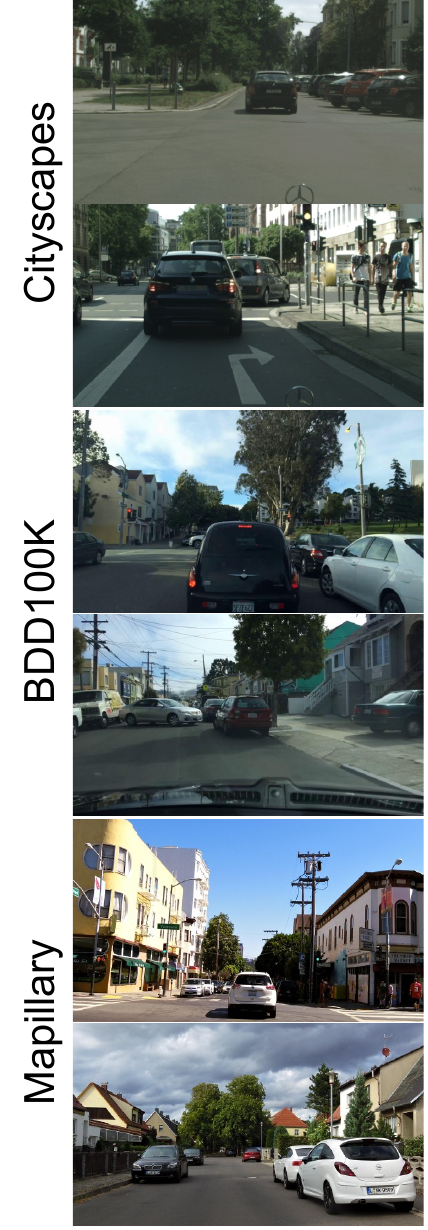}
      \caption{Images}
    \end{subfigure}
    \hspace{-5pt}
    \begin{subfigure}{0.1388\linewidth}
      \includegraphics[width=1.0\linewidth]{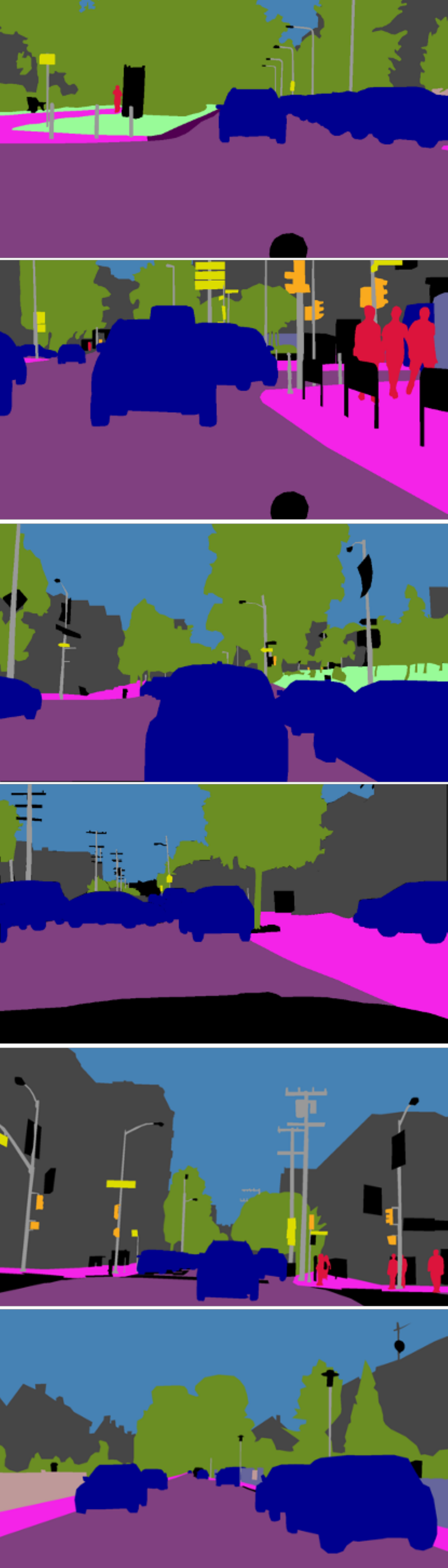}
      \caption{Ground Truth}
    \end{subfigure}
    \hspace{-4pt}
    \begin{subfigure}{0.1388\linewidth}
      \includegraphics[width=1.0\linewidth]{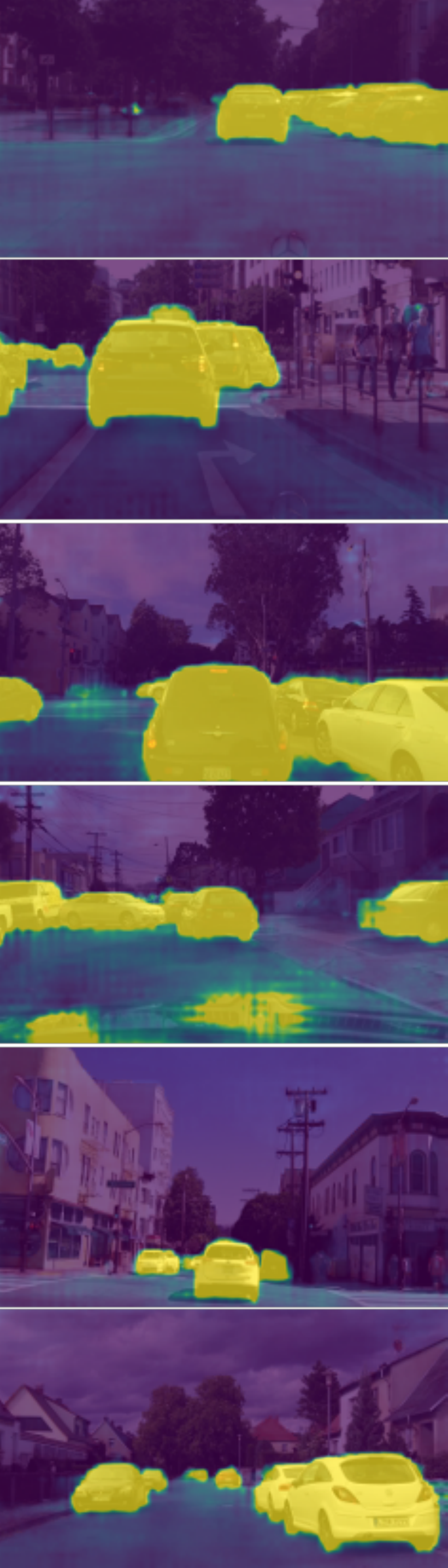}
      \caption{Car}
    \end{subfigure}
    \hspace{-4pt}
    \begin{subfigure}{0.1388\linewidth}
      \includegraphics[width=1.0\linewidth]{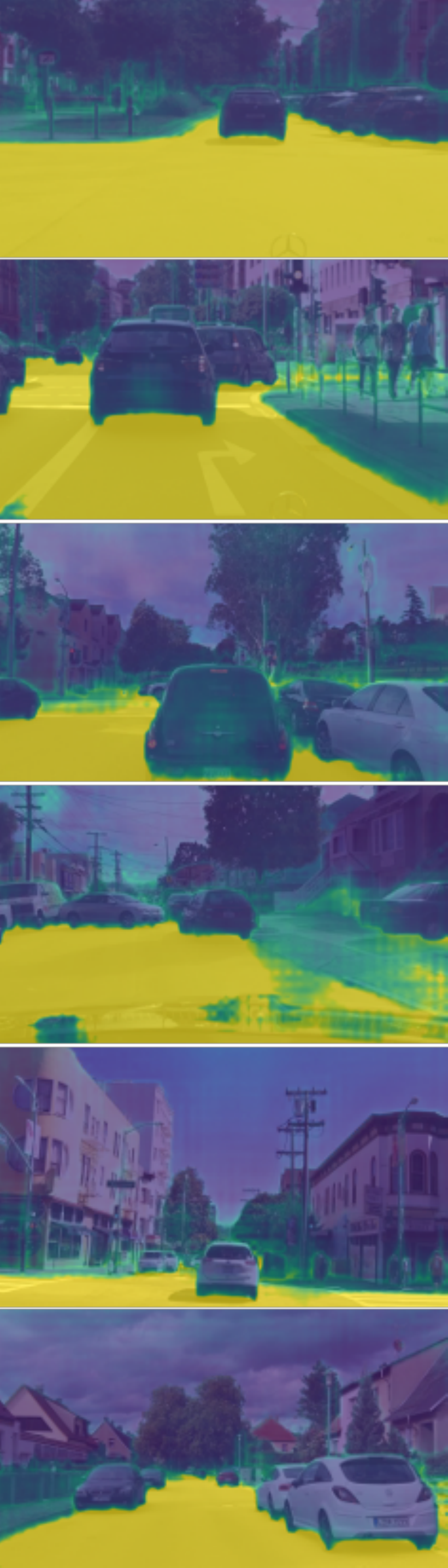}
      \caption{Road}
    \end{subfigure}
    \hspace{-4pt}
    \begin{subfigure}{0.1388\linewidth}
      \includegraphics[width=1.0\linewidth]{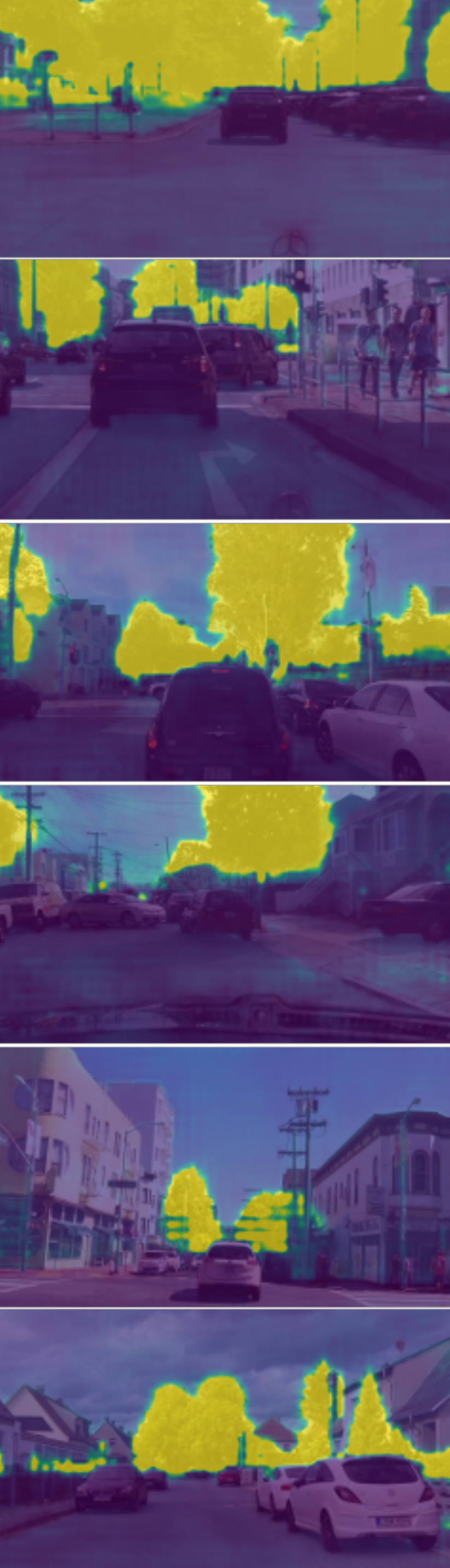}
      \caption{Vegetation}
    \end{subfigure}
    \hspace{-4pt}
    \begin{subfigure}{0.1388\linewidth}
      \includegraphics[width=1.0\linewidth]{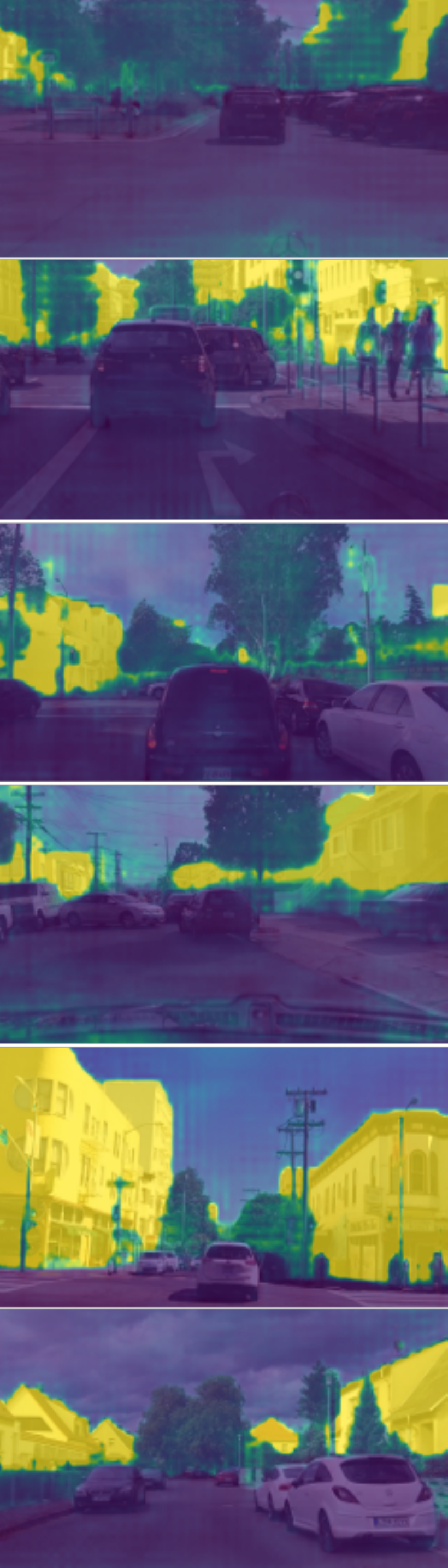}
      \caption{Building}
    \end{subfigure}
    \hspace{-4pt}
    \begin{subfigure}{0.1388\linewidth}
      \includegraphics[width=1.0\linewidth]{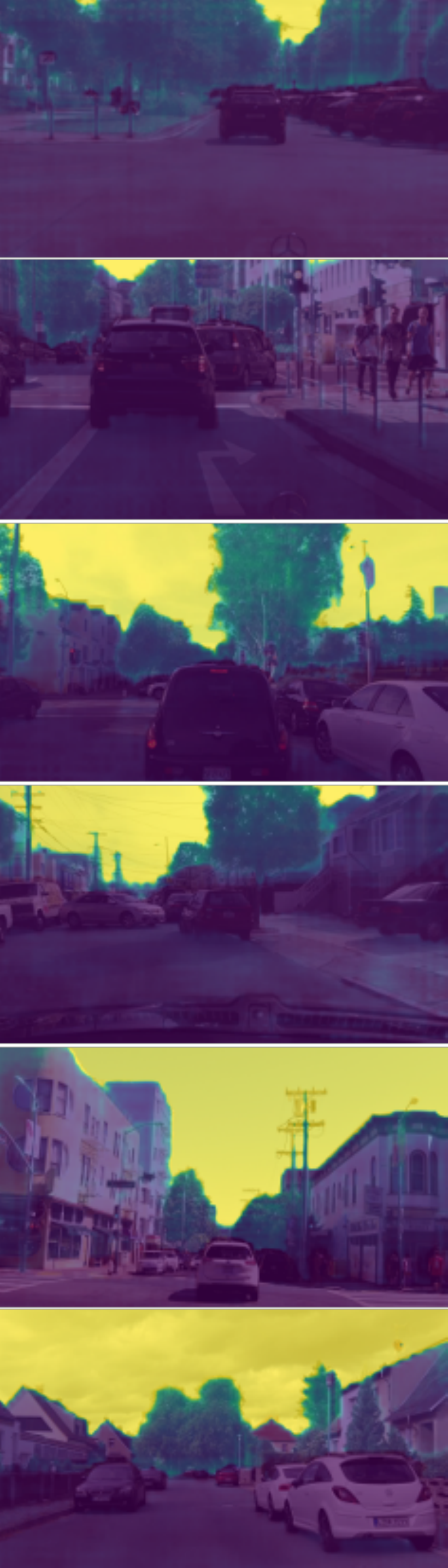}
      \caption{Sky}
    \end{subfigure}
    \\
    \vspace{-8pt}
 	\caption{\textbf{Source (G+S)$\rightarrow$Target (C, B, M):} Visualization of the memory weights for each class on the Cityscapes, BDD100K and Mapillary dataset. We adopt DeepLabV3+ with ResNet50.}
    \label{fig:memactsup}
    \vspace{-4pt}
\end{figure*}

\subsection{Re-implemented Methods}

While IBN-Net~\cite{pan2018ibnnet} improved generalization ability by mixing instance normalization and batch normalization in the backbone, RobustNet~\cite{choi2021robustnet} previously have shown SOTA performance by selectively removing the channel covariance of the backbone.
We re-implemented these two methods by setting the hyper-parameters according to the public code by RobustNet~\cite{choi2021robustnet}\footnote{https://github.com/shachoi/RobustNet}.
To verify the effectiveness of our memory-guided meta-learning method, we re-implemented the MLDG~\cite{mldg} which is meta-learning based DG method.
The augmentations and learning rates of MLDG were same with our method.
Recently, TSMLDG~\cite{tsmldg} purely uses meta-learning for DG and proposes a method for target-domain batch normalization on test-time.
We re-implemented TSMLDG by setting the test-batch size to 4 and updating batch statistics of the MLDG model in testing time on the unseen target domain according to the code of TSMLDG\footnote{https://github.com/koncle/TSMLDG}.

\section{Additional Results}
\subsection{Ablation Study}
\vspace{-8pt}
\paragraph{Analysis of memory updating network}
To verify the effectiveness of the memory updating network, we conduct an ablation study about memory updating network.
In \tabref{tab:updatingnet}, we can observe that the memory updating network has notable
contribution to the performance gain for all datasets by storing generalizable features into the memory.

\begin{table}[!t]
  \centering
     \resizebox{1\linewidth}{!}{
    \begin{tabular}
    {c|c|c|c|c}
    \hlinewd{0.8pt}
    Memory Update Net. & Cityscapes     & BDD100K     & Mapillary     & Avg. \\
    \hline
    \hline
    \xmark &  41.28   &  37.25 &   40.64 & 39.72 \\
   \rowcolor{light_gray} \cmark & \textbf{44.51} & \textbf{38.07} & \textbf{42.70} & \textbf{41.76} \\
    \hlinewd{0.8pt}
    \end{tabular}}
    \vspace{-5pt}
  \caption{\textbf{Source (G+S)$\rightarrow$Target (C, B, M):} Performance with or without memory updating network.}
  \label{tab:updatingnet}%
  \vspace{-8pt}
\end{table}%

\paragraph{More visualization of memory activation}
To complement the Fig. 6 of the main paper, we additionally visualize the memory weight for the input image from all the unseen datasets in \figref{fig:memactsup}.
Regardless of the environment, the memory corresponding to each object category is well activated, so that the feature of the pixel can receive a guide of the appropriate memory feature.
In addition, the results demonstrate that our memory item contains the generic features of the categories, even though the memory has been trained on synthetic datasets.

\begin{table}[!t]
  \centering
  \resizebox{1\linewidth}{!}{
    \begin{tabular}{
    >{\centering}m{0.23\linewidth}|>{\centering}m{0.05\linewidth}
    >{\centering}m{0.05\linewidth}>{\centering}m{0.05\linewidth}
    |>{\centering}m{0.15\linewidth}>{\centering}m{0.15\linewidth}
    >{\centering}m{0.15\linewidth}|>{\centering}m{0.07\linewidth}
    }
    \hlinewd{0.8pt}
    Methods & $\loss{\text{seg}}$ & $\loss{\text{coh}}$ & $\loss{\text{div}}$  & Cityscapes     & BDD100K     & Mapillary     & Avg. \tabularnewline
    \hline
    \hline
    IBN-Net~\cite{pan2018ibnnet} & \cmark & \xmark & \xmark & 35.55     & 32.18     & 38.09     & 35.27 \tabularnewline
    \hline
    MLDG~\cite{mldg}  & \cmark & \xmark & \xmark & 38.84     & 31.95     & 35.60     & 35.46 \tabularnewline
    \hline
    \multirow{2}{*}{Ours} & \cmark & \xmark & \xmark &  38.22 & 33.12 & 37.10 & 36.15 \tabularnewline
    & \cmark & \cmark & \cmark & \textbf{44.51} & \textbf{38.07} & \textbf{42.70} & \textbf{41.76} \tabularnewline
    \hlinewd{0.8pt}
    \end{tabular}}
    \vspace{-6pt}
    \caption{\textbf{Source (G+S)$\rightarrow$Target (C, B, M):} Mean IoU(\%) comparison between the DG methods with only standard segmentation loss, $\loss{\text{seg}}$. All networks are DeepLabV3+ with ResNet50.}
    \vspace{4pt}
  \label{tab:rebutloss}%
\end{table}%

\vspace{-2pt}
\paragraph{Loss comparison with previous works}
To convincingly compare our proposed losses with previous works, we re-implemented our model using only standard loss (cross entropy) in \tabref{tab:rebutloss}. 
Without the proposed losses, our method still shows competitive performance against IBN-Net~\cite{pan2018ibnnet} and MLDG~\cite{mldg} due to the help of memory items. 
Moreover, $\loss{\text{coh}}$ and $\loss{\text{div}}$ lead to substantial performance gain by facilitating the effective memory read/update procedures in training.

 \begin{figure}[t]
   \centering
   {\includegraphics[width=1\linewidth]{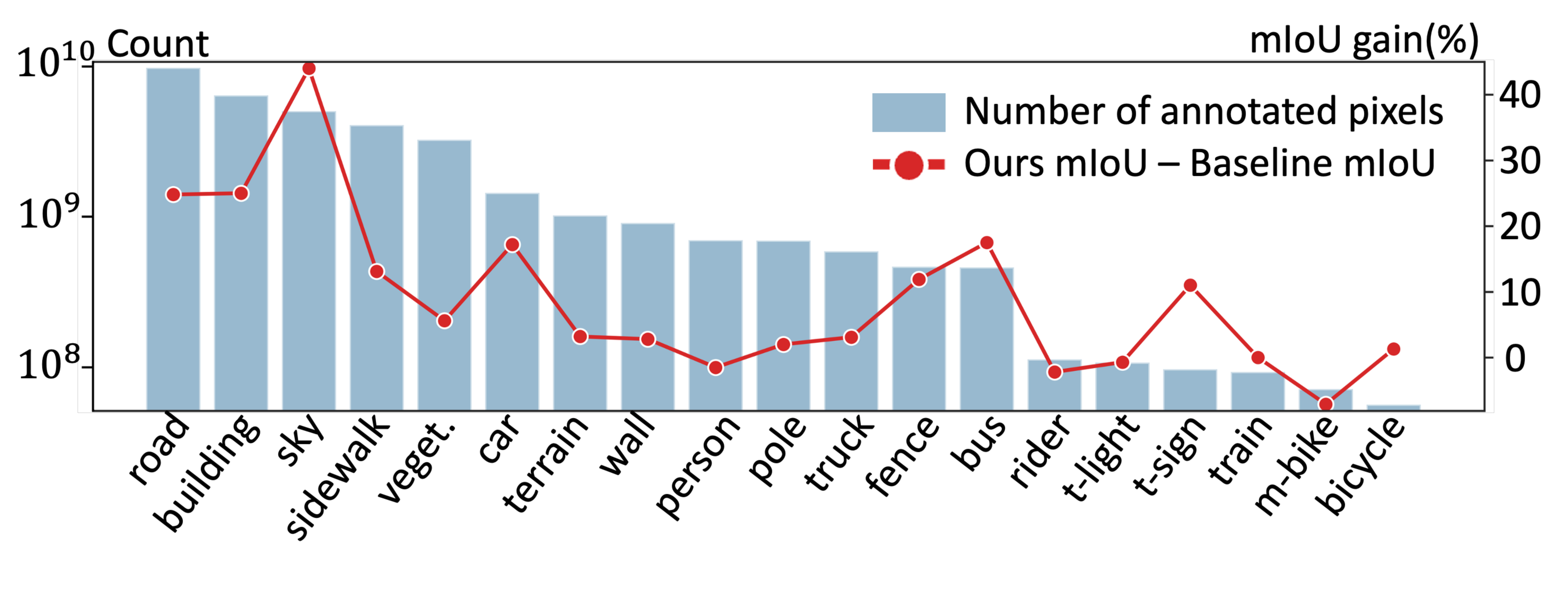}}
   \vspace{-18pt}
  \caption{The correlation between the number of pixels per class in source datasets (G, S) and performance gain on BDD100K dataset.}
  \label{fig:histograph}
  \vspace{4pt}
\end{figure}

\vspace{-3pt}
\paragraph{Correlation between performance gain and class distribution}
The generalization capability usually benefits from the diversity and amount of the training samples.
However, the data imbalance between classes in current benchmarks is significant since the different occurrence frequency and variants of shape among classes.
In \figref{fig:histograph}, we analyze the correlation between the performance gain over the baseline in Table 1 of the main paper and the number of training samples.
While the high mIoU gain is attained for the class (e.g. road, building, sky) with sufficient training samples, it becomes lower for some minor classes.
We remain this problem due to the limitation of current benchmarks as future work.

\begin{table}[!t]
  \centering
  \resizebox{0.75\linewidth}{!}{
    \begin{tabular}{c|c|c|c}
    \hlinewd{0.8pt}
    Methods & BDD100K & Mapillary & Avg. \\
    \hline
    \hline
    MCIBI~\cite{jin2021mining} & 41.65 & 50.18 & 45.92 \\
    \rowcolor{light_gray} Ours &  \textbf{46.78} & \textbf{55.10} & \textbf{50.94} \\
    \hlinewd{0.8pt}
    \end{tabular}}
  \caption{\textbf{Source (C)$\rightarrow$Target (B, M):} Mean IoU(\%) comparison with MCIBI~\cite{jin2021mining}. All networks are DeepLabV3 with ResNet50.}
  \label{tab:mcibi}%
\end{table}%

\paragraph{Comparison with MCIBI}
We conduct comparison with MCIBI~\cite{jin2021mining} which is a memory network designed for conducting semantic segmentation on seen domain dataset.
To compare generalization performance, we used the author-provided MCIBI model pre-trained on Cityscapes and evaluated on the other real datasets regarding single-source setting.
In \tabref{tab:mcibi}, we can see that our memory module outperforms MCIBI on unseen domain datasets.
It thus points out that using our non-parametric memory loss and leveraging meta-learning to store shared information among the same class play important roles in improving generalization capability of the segmentation network.

\begin{table}[!t]
  \centering
    \resizebox{1\linewidth}{!}{
    \begin{tabular}{c|l|c|c|c|c}
    \hlinewd{0.8pt}
    Backbone & Methods & Seg. model & Cityscapes   & BDD100K     & Mapillary \tabularnewline
    \hline
    \hline
    \multirow{8}{*}{Resnet50} &
    Baseline & \multirow{2}{*}{FCN-8s} &  32.50 & 26.70  & 25.70 \tabularnewline
     & DRPC~\cite{yue2019domain} &       &  37.40 & 32.10 & 34.10  \tabularnewline
    \cline{2-6} 
     & Baseline$^\dagger$ &  &  29.00 & 25.10  & 28.20 \tabularnewline
     & IBN-Net$^\dagger$~\cite{pan2018ibnnet} &         &  33.90 & 32.30 & 37.80 \tabularnewline
     & RobustNet$^\dagger$~\cite{choi2021robustnet} &        &   36.60 & 35.20  & 40.30  \tabularnewline
      & \cellcolor{light_gray} Baseline & \cellcolor{light_gray}  &  \cellcolor{light_gray}31.60 & \cellcolor{light_gray}26.70  & \cellcolor{light_gray}29.00 \tabularnewline
     & MLDG$^\ddagger$~\cite{mldg} &     &  36.70  & 32.10  &  32.20  \tabularnewline
      & \cellcolor{light_gray} \textbf{Ours}  & \cellcolor{light_gray}\multirow{-6}{*}{DeepLabV3+} &   \cellcolor{light_gray}41.00  & \cellcolor{light_gray}34.60  & \cellcolor{light_gray}37.40  \tabularnewline
    \hline
    \multirow{2}{*}{Resnet101}  & FSDR~\cite{huang2021fsdr} &  &  44.75 & 39.66 & 40.87  \tabularnewline
     & \cellcolor{light_gray} \textbf{Ours}  & \cellcolor{light_gray}\multirow{-2}{*}{DeepLabV2} &  \cellcolor{light_gray}\textbf{44.90} & \cellcolor{light_gray}\textbf{39.71} & \cellcolor{light_gray}\textbf{41.31}  \tabularnewline
    \hlinewd{0.8pt}
    \end{tabular}}
    \caption{\textbf{Source (G)$\rightarrow$Target (C, B, M):} Mean IoU(\%) comparison of other SOTA methods using various segmentation models and backbones. MLDG~\cite{mldg} is re-implemented. Results with $^\dagger$ are from~\cite{choi2021robustnet}.}
    \vspace{6pt}
  \label{tab:gdgcbm}
\end{table}%

\begin{table*}[!ht]
  \centering
    \resizebox{\linewidth}{!}{
    \begin{tabular}{
    >{\raggedright}m{0.015\linewidth}>{\raggedright}m{0.131\linewidth}|
    >{\centering}p{0.027\linewidth}>{\centering}p{0.027\linewidth}
    >{\centering}p{0.027\linewidth}>{\centering}p{0.027\linewidth}
    >{\centering}p{0.027\linewidth}>{\centering}p{0.027\linewidth}
    >{\centering}p{0.027\linewidth}>{\centering}p{0.027\linewidth}
    >{\centering}p{0.027\linewidth}>{\centering}p{0.027\linewidth}
    >{\centering}p{0.027\linewidth}>{\centering}p{0.027\linewidth}
    >{\centering}p{0.027\linewidth}>{\centering}p{0.027\linewidth}
    >{\centering}p{0.027\linewidth}>{\centering}p{0.027\linewidth}
    >{\centering}p{0.027\linewidth}>{\centering}p{0.027\linewidth}
    >{\centering}p{0.027\linewidth}|>{\centering}m{0.08\linewidth}}
    \hlinewd{0.8pt}
    & Methods
    & \begin{sideways}road\end{sideways} & \begin{sideways}sidewalk\end{sideways}
    & \begin{sideways}building\end{sideways} & \begin{sideways}wall\end{sideways}
    & \begin{sideways}fence\end{sideways} & \begin{sideways}pole\end{sideways}
    & \begin{sideways}t-light\end{sideways} & \begin{sideways}t-sign\end{sideways}
    & \begin{sideways}vegettion\end{sideways} & \begin{sideways}terrain\end{sideways}
    & \begin{sideways}sky\end{sideways} & \begin{sideways}person\end{sideways}
    & \begin{sideways}rider\end{sideways} & \begin{sideways}car\end{sideways}
    & \begin{sideways}truck\end{sideways} & \begin{sideways}bus\end{sideways}
    & \begin{sideways}train\end{sideways} & \begin{sideways}m-bike\end{sideways}
    & \begin{sideways}bicycle\end{sideways} & mIoU(\%) \tabularnewline
    \hline
    \hline
  \multirow{6}[2]{*}{\begin{sideways}Cityscapes\end{sideways}} & Baseline & 88.6  & 45.9  & 85.5  & 38.2  & 29.7  & 46.0  & \textbf{45.0} & 41.6  & 88.6  & \textbf{43.3} & \textbf{93.2} & 73.5  & 44.0  & 81.4  & 46.3  & 29.3  & 0.3   & 30.0  & 47.3  & 52.5  \tabularnewline
          & IBN-Net$^\ddagger$~\cite{pan2018ibnnet} & 90.2  & 52.0  & 86.9  & 38.4  & 31.8  & 47.8  & 43.6  & 43.8  & \textbf{89.3} & 42.3  & 91.9  & 72.8  & 42.8  & 82.3  & 50.5  & 48.6  & 0.2   & 28.8  & 49.3  & 54.4  \tabularnewline
          & RobustNet$^\ddagger$~\cite{choi2021robustnet} & 90.4  & 48.1  & 86.8  & 36.1  & 34.6  & 47.3  & 39.3  & \textbf{43.9} & 89.2  & 40.7  & 92.1  & 73.2  & 44.6  & 87.8  & 51.7  & 50.8  & 0.0   & 32.2  & 50.6  & 54.7  \tabularnewline
          & MLDG$^\ddagger$~\cite{mldg} & 91.2  & 50.8  & 87.4  & 39.5  & 30.4  & \textbf{49.0} & 39.4  & 42.5  & 89.1  & 39.2  & 93.0  & \textbf{74.1} & \textbf{46.0} & 86.4  & 50.3  & 49.6  & 0.6   & 31.4  & 50.5  & 54.8  \tabularnewline
          & TSMLDG$^\ddagger$~\cite{tsmldg} & \textbf{92.1} & \textbf{52.7} & 87.4  & 37.1  & 31.3  & 48.5  & 40.5  & 42.7  & 89.1  & 39.2  & 92.6  & 72.1  & 41.8  & 89.0  & 49.3  & 47.2  & 0.6   & 18.5  & 35.8  & 53.0  \tabularnewline
          &\cellcolor{light_gray} \textbf{Ours}  & \cellcolor{light_gray}91.0  & \cellcolor{light_gray}51.6  & \cellcolor{light_gray}\textbf{87.9} & \cellcolor{light_gray}\textbf{43.1} & \cellcolor{light_gray}\textbf{36.6} & \cellcolor{light_gray}47.6  & \cellcolor{light_gray}38.7  & \cellcolor{light_gray}43.1  & \cellcolor{light_gray}\textbf{89.3} & \cellcolor{light_gray}41.8  & \cellcolor{light_gray}93.0  & \cellcolor{light_gray}73.9  & \cellcolor{light_gray}41.9  & \cellcolor{light_gray}\textbf{89.1} & \cellcolor{light_gray}\textbf{58.9} & \cellcolor{light_gray}\textbf{55.8} & \cellcolor{light_gray}\textbf{2.0} & \cellcolor{light_gray}\textbf{37.2} & \cellcolor{light_gray}\textbf{52.5} & \cellcolor{light_gray}\textbf{56.6} \tabularnewline
    \hline
    \hline
    \multirow{6}[2]{*}{\begin{sideways}BDD100k\end{sideways}} & Baseline & 89.8  & 42.7  & 76.8  & 14.1  & 41.9  & 43.6  & 34.7  & 31.7  & 81.0  & 40.6  & 90.3  & 62.2  & 26.4  & 82.2  & 26.7  & 40.2  & 0.0   & 38.1  & 38.8  & 47.5  \tabularnewline
          & IBN-Net$^\ddagger$~\cite{pan2018ibnnet} & 88.5  & 46.7  & \textbf{78.7} & \textbf{20.6} & 40.8  & \textbf{45.4} & \textbf{39.4} & 32.8  & \textbf{82.8} & 42.1  & \textbf{91.6} & 61.3  & 21.7  & 80.7  & 33.7  & \textbf{59.8} & 0.0   & 23.4  & 39.4  & 48.9  \tabularnewline
          & RobustNet$^\ddagger$~\cite{choi2021robustnet} & 90.3  & 42.6  & 77.7  & 20.4  & 39.9  & 44.6  & 36.6  & 33.3  & \textbf{82.8} & \textbf{43.8} & 90.8  & 61.6  & 21.7  & \textbf{84.2} & 32.3  & 57.7  & 0.0   & 24.8  & \textbf{46.2} & 49.0  \tabularnewline
          & MLDG$^\ddagger$~\cite{mldg} & 90.0  & 45.7  & 75.8  & 15.1  & \textbf{43.6} & 43.1  & 36.4  & 32.0  & 82.3  & 41.2  & 89.8  & 61.1  & 19.5  & 80.9  & 33.4  & 52.1  & 0.0   & 39.5  & 40.4  & 48.5  \tabularnewline
          & TSMLDG$^\ddagger$~\cite{tsmldg} & \textbf{90.8} & 45.4  & 78.0  & 16.4  & 34.9  & 44.5  & 38.2  & \textbf{34.7} & 81.7  & 37.3  & 91.4  & 57.6  & 12.9  & 84.1  & \textbf{34.3} & 53.8  & 0.0   & 9.0   & 36.9  & 46.4  \tabularnewline
          &\cellcolor{light_gray}\textbf{Ours}  & \cellcolor{light_gray}90.4  & \cellcolor{light_gray}\textbf{52.5} & \cellcolor{light_gray}75.2  & \cellcolor{light_gray}18.2  & \cellcolor{light_gray}41.8  & \cellcolor{light_gray}43.9  & \cellcolor{light_gray}38.6  & \cellcolor{light_gray}34.4  & \cellcolor{light_gray}82.5  & \cellcolor{light_gray}40.0  & \cellcolor{light_gray}89.7  & \cellcolor{light_gray}\textbf{62.5} & \cellcolor{light_gray}\textbf{26.5} & \cellcolor{light_gray}83.3  & \cellcolor{light_gray}31.0  & \cellcolor{light_gray}56.2  & \cellcolor{light_gray}0.0   & \cellcolor{light_gray}\textbf{46.2} & \cellcolor{light_gray}40.5  & \cellcolor{light_gray}\textbf{50.2} \tabularnewline
    \hline
    \hline
    \multirow{6}[2]{*}{\begin{sideways}Mapillary\end{sideways}} & Baseline & 87.8  & 40.3  & 81.2  & 29.2  & 37.9  & 51.5  & 42.6  & 63.7  & 87.2  & 48.4  & 97.2  & 71.4  & 44.9  & 85.9  & 50.7  & 30.9  & 0.5   & 47.5  & 40.5  & 54.7  \tabularnewline
          & IBN-Net$^\ddagger$~\cite{pan2018ibnnet} & 88.5  & 44.9  & \textbf{83.6} & 35.3  & 38.3  & \textbf{53.1} & 43.7  & 63.4  & 87.5  & 47.8  & \textbf{97.4} & 71.6  & 48.3  & 86.1  & 47.8  & 41.0  & 3.9   & 45.8  & 37.1  & 56.1  \tabularnewline
          & RobustNet$^\ddagger$~\cite{choi2021robustnet} & 88.2  & 43.5  & 83.1  & 34.2  & 39.4  & 52.5  & 40.2  & 62.6  & 87.3  & 48.4  & 97.3  & 72.3  & \textbf{51.8} & 87.7  & 48.7  & 51.7  & \textbf{7.3} & 45.4  & 39.8  & 56.9  \tabularnewline
          & MLDG$^\ddagger$~\cite{mldg} & 88.0  & 39.0  & 82.9  & 36.6  & 40.3  & 51.6  & 41.7  & 64.4  & 87.6  & 45.7  & 96.9  & \textbf{73.0} & 51.6  & 87.3  & 39.0  & 44.3  & 3.5   & 48.5  & 41.0  & 55.9  \tabularnewline
          & TSMLDG$^\ddagger$~\cite{tsmldg} & 86.1  & 45.7  & 79.2  & 31.4  & 39.9  & 52.2  & \textbf{44.4} & 61.8  & 84.2  & 38.5  & 88.1  & 68.8  & 49.2  & 86.6  & 31.0  & 31.8  & 5.3   & 42.7  & 35.3  & 52.7  \tabularnewline
          &\cellcolor{light_gray} \textbf{Ours}  & \cellcolor{light_gray}\textbf{89.2} & \cellcolor{light_gray}\textbf{48.1} & \cellcolor{light_gray}83.2  & \cellcolor{light_gray}\textbf{36.9} & \cellcolor{light_gray}\textbf{40.6} & \cellcolor{light_gray}52.4  & \cellcolor{light_gray}42.3  & \cellcolor{light_gray}\textbf{64.8} & \cellcolor{light_gray}\textbf{87.7} & \cellcolor{light_gray}\textbf{49.6} & \cellcolor{light_gray}97.3  & \cellcolor{light_gray}72.2  & \cellcolor{light_gray}47.3  & \cellcolor{light_gray}\textbf{89.2} & \cellcolor{light_gray}\textbf{53.6} & \cellcolor{light_gray}\textbf{55.9} & \cellcolor{light_gray}3.9   & \cellcolor{light_gray}\textbf{49.4} & \cellcolor{light_gray}\textbf{44.2} & \cellcolor{light_gray}\textbf{58.3} \tabularnewline
    \hlinewd{0.8pt}
    \end{tabular}
    }
    \vspace{-5pt}
    \caption{\textbf{Source (G+S+I)$\rightarrow$Target (C, B, M):} Mean IoU(\%) and per-class IoU(\%) comparison of other state-of-the-art DG methods for semantic segmentation. We re-implemented all methods using DeepLabV3+ with ResNet50 backbone. We re-implement other methods and mark them with $^\ddagger$.}
    \label{tab:gsidg_class}
    \vspace{-4pt}
\end{table*}%

\begin{table*}[!ht]
  \centering
    \resizebox{\linewidth}{!}{
    \begin{tabular}{
    >{\raggedright}m{0.015\linewidth}>{\raggedright}m{0.12\linewidth}|>{\centering}m{0.07\linewidth}|
    >{\centering}p{0.025\linewidth}>{\centering}p{0.025\linewidth}
    >{\centering}p{0.025\linewidth}>{\centering}p{0.025\linewidth}
    >{\centering}p{0.025\linewidth}>{\centering}p{0.025\linewidth}
    >{\centering}p{0.025\linewidth}>{\centering}p{0.025\linewidth}
    >{\centering}p{0.025\linewidth}>{\centering}p{0.025\linewidth}
    >{\centering}p{0.025\linewidth}>{\centering}p{0.025\linewidth}
    >{\centering}p{0.025\linewidth}>{\centering}p{0.025\linewidth}
    >{\centering}p{0.025\linewidth}>{\centering}p{0.025\linewidth}
    |>{\centering}m{0.08\linewidth}}
    \hlinewd{0.8pt}
     & Methods & w/Target
     & \begin{sideways}road\end{sideways} & \begin{sideways}sidewalk\end{sideways}
     & \begin{sideways}building\end{sideways} & \begin{sideways}wall\end{sideways}
     & \begin{sideways}fence\end{sideways} & \begin{sideways}pole\end{sideways}
     & \begin{sideways}t-light\end{sideways} & \begin{sideways}t-sign\end{sideways}
     & \begin{sideways}vegettion\end{sideways} & \begin{sideways}sky\end{sideways}
     & \begin{sideways}person\end{sideways} & \begin{sideways}rider\end{sideways}
     & \begin{sideways}car\end{sideways} & \begin{sideways}bus\end{sideways}
     & \begin{sideways}m-bike\end{sideways} & \begin{sideways}bicycle\end{sideways} & mIoU(\%) \tabularnewline
    \hline
    \hline
    \multirow{7}[2]{*}{\begin{sideways}Cityscapes\end{sideways}} & Baseline & \xmark & 77.1  & 32.4  & 75.3  & 13.8  & 11.5  & 29.0  & 13.7  & 10.3  & 81.5  & 79.1  & 53.1  & 10.2  & 80.2  & 39.0  & 21.9  & 11.5  & 40.0 \tabularnewline
          & CyCADA$^\dagger$~\cite{hoffman2018cycada} & \cmark & 86.8  & 41.4  & 74.7  & 15.5  & 3.4   & 27.3  & 3.8   & 0.2   & 73.2  & 72.4  & 51.9  & 12.7  & 82.7  & 41.8  & 18.5  & 23.3  & 39.3 \tabularnewline
          & MDAN$^\dagger$~\cite{zhao2018mdan} & \cmark & 80.6  & 34.4  & 73.9  & 15.9  & 1.9   & 22.9  & 0.1   & 0.0   & 73.6  & 58.9  & 48.4  & 12.2  & 78.8  & 36.8  & 14.2  & 23.7  & 36.0 \tabularnewline
          & MADAN$^\dagger$~\cite{zhao2019multi} & \cmark & 88.1  & 46.1  & 79.9  & 26.4  & 7.4   & 30.6  & 19.0  & 19.9  & 80.4  & 75.9  & 55.6  & 15.6  & 84.1  & \textbf{47.0} & 23.3  & 26.3  & 45.4 \tabularnewline
          & MADAN+$^\dagger$~\cite{zhao2021madan} & \cmark & \textbf{90.9} & \textbf{49.7} & 64.9  & 24.6  & 13.0  & \textbf{39.2} & 40.0  & 21.4  & 80.2  & \textbf{86.1} & 57.3  & \textbf{25.0} & 84.7  & 35.7  & 25.2  & \textbf{38.2} & 48.5 \tabularnewline
          & CLSS~\cite{he2021multi} & \cmark & -     & -     & -     & -     & -     & -     & -     & -     & -     & -     & -     & -     & -     & -     & -     & -     & \textbf{54.0} \tabularnewline
          &\cellcolor{light_gray} \textbf{Ours}  & \cellcolor{light_gray}\xmark & \cellcolor{light_gray}87.4  & \cellcolor{light_gray}42.7  & \cellcolor{light_gray}\textbf{82.6} & \cellcolor{light_gray}\textbf{29.9} & \cellcolor{light_gray}\textbf{21.5} & \cellcolor{light_gray}\textbf{39.2} & \cellcolor{light_gray}\textbf{48.5} & \cellcolor{light_gray}\textbf{34.2} & \cellcolor{light_gray}\textbf{85.2} & \cellcolor{light_gray}71.8  & \cellcolor{light_gray}\textbf{66.6} & \cellcolor{light_gray}17.6  & \cellcolor{light_gray}\textbf{88.8} & \cellcolor{light_gray}21.5  & \cellcolor{light_gray}\textbf{26.0} & \cellcolor{light_gray}26.5  & \cellcolor{light_gray}49.4 \tabularnewline
    \hline
    \hline
    \multirow{6}[2]{*}{\begin{sideways}BDD100K\end{sideways}} & Baseline & \xmark & 55.3  & 20.9  & 73.9  & 15.9  & 18.9  & 29.9  & 11.3  & 11.9  & 79.7  & 76.2  & 54.7  & 10.3  & 79.7  & 29.3  & 17.2  & 14.1  & 37.4 \tabularnewline
          & CyCADA$^\dagger$~\cite{hoffman2018cycada} & \cmark & 64.9  & 33.6  & 73.3  & 15.8  & 15.3  & 29.2  & 15.9  & 21.4  & 79.3  & 79.0  & 52.0  & 12.7  & 49.7  & 14.0  & 17.5  & 22.5  & 37.2 \tabularnewline
          & MDAN$^\dagger$~\cite{zhao2018mdan} & \cmark & 57.6  & 31.2  & 53.5  & 6.5   & 0.6   & 20.3  & 0.0   & 0.0   & 73.0  & 61.7  & 40.9  & 9.8   & 60.4  & 29.2  & 10.3  & 15.6  & 29.4 \tabularnewline
          & MADAN$^\dagger$~\cite{zhao2019multi} & \cmark & 74.5  & 32.4  & 71.3  & 16.5  & 16.3  & 30.6  & 15.1  & 25.1  & \textbf{80.6} & 78.7  & 52.2  & 12.4  & 70.5  & 34.0  & 18.4  & 19.4  & 40.4 \tabularnewline
          & MADAN+$^\dagger$~\cite{zhao2021madan} & \cmark & \textbf{87.8} & \textbf{44.2} & \textbf{78.6} & \textbf{22.4} & 6.8   & 29.1  & 11.5  & 5.3   & 79.6  & 74.6  & 53.6  & 14.6  & \textbf{83.0} & \textbf{43.4} & 19.1  & \textbf{30.2} & 42.7 \tabularnewline
          &\cellcolor{light_gray} \textbf{Ours}  & \cellcolor{light_gray}\xmark & \cellcolor{light_gray}84.5  & \cellcolor{light_gray}39.8  & \cellcolor{light_gray}69.7  & \cellcolor{light_gray}9.0   & \cellcolor{light_gray}\textbf{26.3} & \cellcolor{light_gray}\textbf{36.1} & \cellcolor{light_gray}\textbf{43.3} & \cellcolor{light_gray}\textbf{31.3} & \cellcolor{light_gray}73.5  & \cellcolor{light_gray}\textbf{87.1} & \cellcolor{light_gray}\textbf{59.2} & \cellcolor{light_gray}\textbf{25.5} & \cellcolor{light_gray}81.9  & \cellcolor{light_gray}6.6   & \cellcolor{light_gray}\textbf{38.3} & \cellcolor{light_gray}15.2  & \cellcolor{light_gray}\textbf{45.5} \tabularnewline
    \hlinewd{0.8pt}
    \end{tabular}
}
    \vspace{-5pt}
    \caption{\textbf{Source (G+S)$\rightarrow$Target (C, B):} Mean IoU(\%) and per-class IoU(\%) comparison of other multi-source UDA methods. The segmentation models are all DeepLabV2 with ResNet101. Results with $^\dagger$ are from~\cite{zhao2021madan}.}
    \label{tab:gsuda_class}
    \vspace{-7pt}
\end{table*}%

\subsection{Full Comparison with State-Of-The-Art.}
\vspace{-3pt}
\paragraph{Quantitative results}
\tabref{tab:gdgcbm} shows the results evaluated on the real datasets with various segmentation models regarding to single-source domain generalization setting.
Even though the networks were trained on the GTA\RomanNumeralCaps{5} dataset only, our method obtained the best generalization performance on the Cityscapes dataset.
Our method also achieved a relatively high-performance gain over our baseline results on the BDD100K and Mapillary datasets.
We also compare with the performance of FSDR~\cite{huang2021fsdr} where we used the author-provided model parameters of FSDR pre-trained on GTAV. Our model performs better than FSDR on all the target domain datasets.

Furthermore, we report the per-class IoU scores for Table 2 and Table 4 of the main paper in \tabref{tab:gsidg_class} and \tabref{tab:gsuda_class}, respectively.
\tabref{tab:gsidg_class} shows the performance of Cityscapes, BDD100K, and Mapillary with DG models trained on GTA5, Synthia, and IDD datasets.
The results show that our method increased the average performance of each class without overfitting a specific category in the unseen domain.
In \tabref{tab:gsuda_class}, we compare the performance on the real-world datasets with state-of-the-art multi-source UDA methods that leverage target domain images on training time.
Although UDA is a much easier setting than domain generalization, our DG method achieved the highest performance on the BDD100K and competitive results on the Cityscapes.

\paragraph{Qualitative results}
To qualitatively describe the superiority of our method, we compare the segmentation results with other state-of-the-art DG methods.
We trained all DG methods on multi-source synthetic datasets (\ie GTA\RomanNumeralCaps{5}~\cite{gtav} and Synthia~\cite{synthia}), and tested on the \textit{unseen} real-world datasets~\cite{cordts2016cityscapes, yu2020bdd100k,neuhold2017mapillary}.

\begin{figure*}[t]
	\centering
    \includegraphics[width=0.9\linewidth]{figure/classlabel.pdf}\hfill \\
    \centering
    \begin{subfigure}{0.1426\linewidth}
      \includegraphics[width=1.0\linewidth]{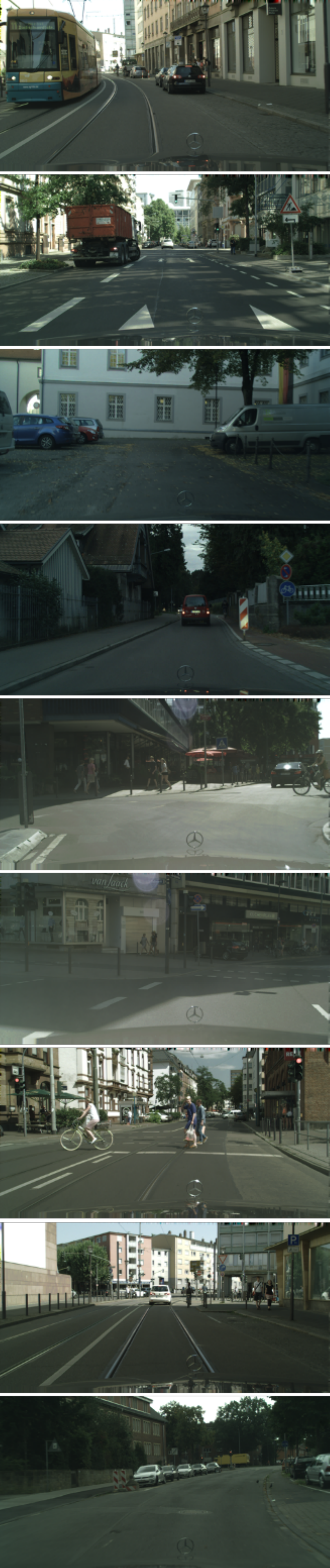}
      \caption{Images}
    \end{subfigure}
    \hspace{-4.5pt}
    \begin{subfigure}{0.1422\linewidth}
      \includegraphics[width=1.0\linewidth]{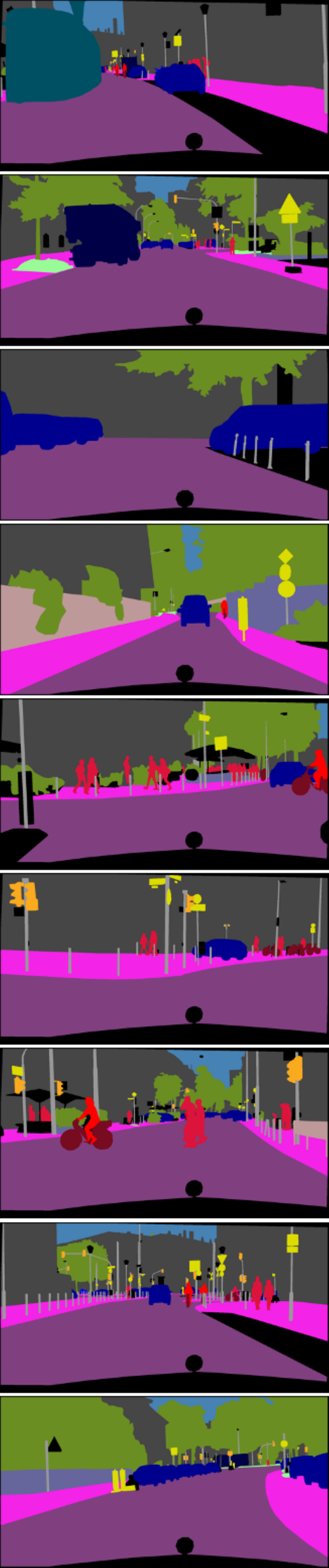}
      \caption{Ground Truth}
    \end{subfigure}
    \hspace{-4.5pt}
    \begin{subfigure}{0.1426\linewidth}
      \includegraphics[width=1.0\linewidth]{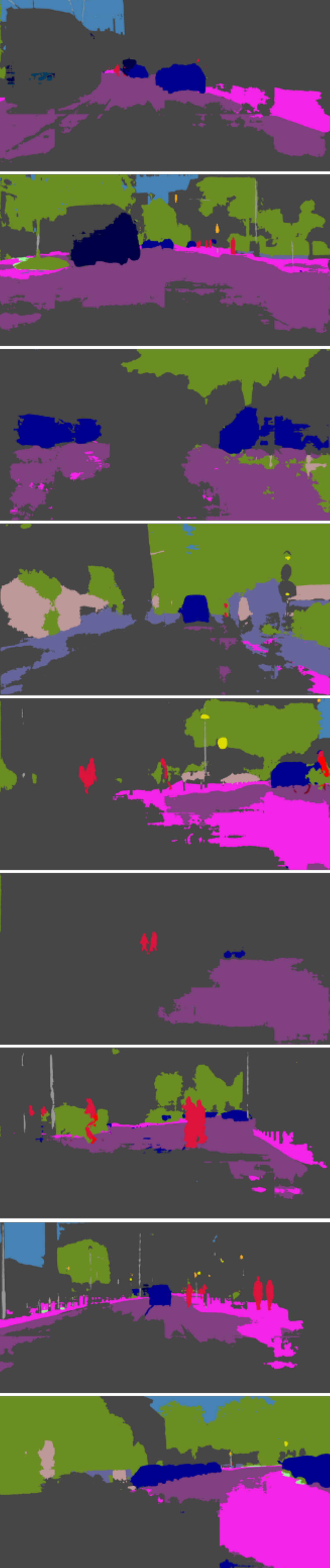}
      \caption{Baseline}
    \end{subfigure}
    \hspace{-4.5pt}
    \begin{subfigure}{0.1426\linewidth}
      \includegraphics[width=1.0\linewidth]{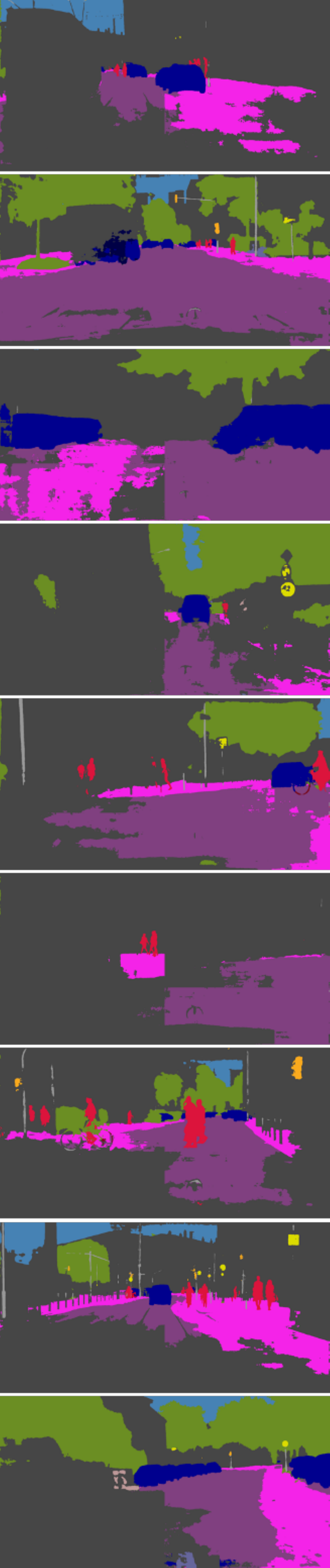}
      \caption{IBN-Net~\cite{pan2018ibnnet}}
    \end{subfigure}
    \hspace{-4.5pt}
    \begin{subfigure}{0.1426\linewidth}
      \includegraphics[width=1.0\linewidth]{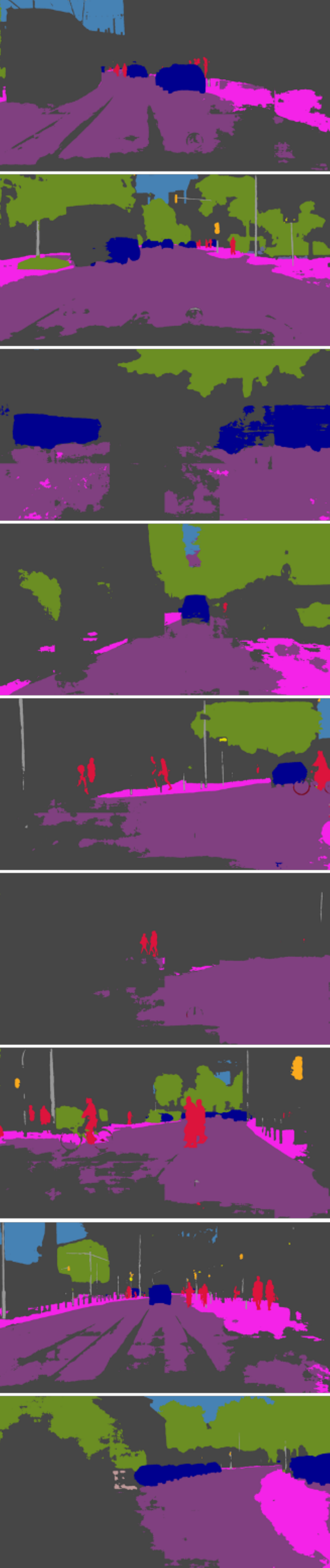}
      \caption{RobustNet~\cite{choi2021robustnet}}
    \end{subfigure}
    \hspace{-4.5pt}
    \begin{subfigure}{0.1426\linewidth}
      \includegraphics[width=1.0\linewidth]{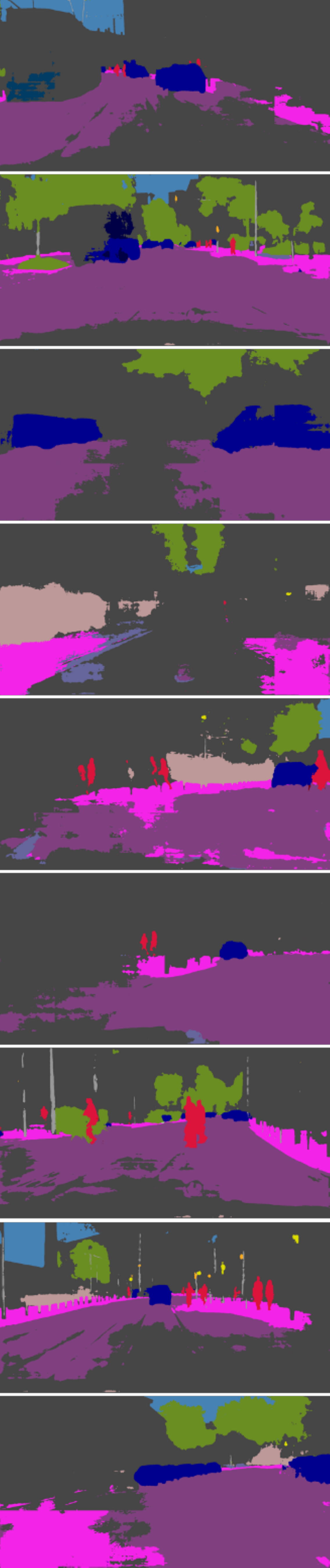}
      \caption{MLDG~\cite{mldg}}
    \end{subfigure}
    \hspace{-4.5pt}
    \begin{subfigure}{0.1426\linewidth}
      \includegraphics[width=1.0\linewidth]{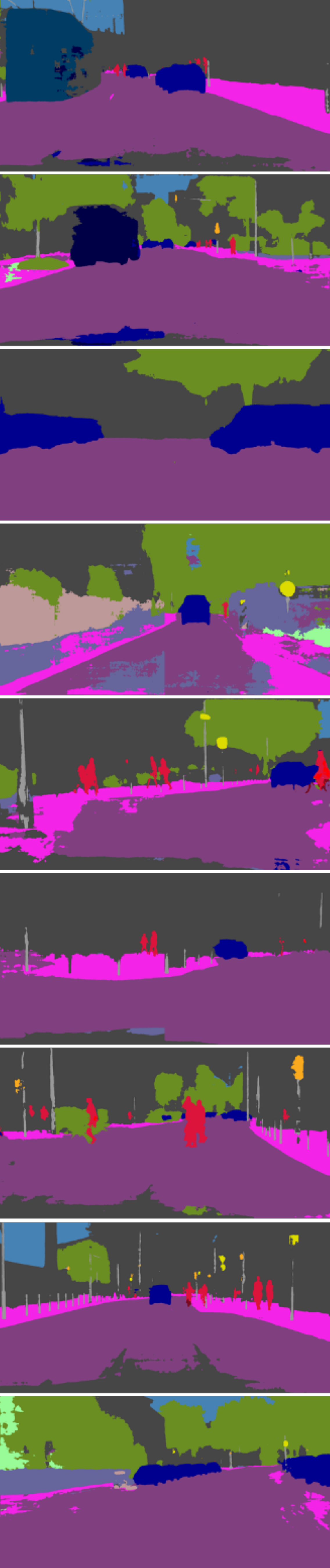}
      \caption{Ours}
    \end{subfigure}
    
    \vspace{-9pt}
 	\caption{
 	\textbf{Source (G+S)$\rightarrow$Target (C):} Qualitative comparison on the Cityscapes dataset. All methods adopt DeepLabV3+ with ResNet50. (Best viewed in color.)
 	}\label{fig:qualcitysup}\vspace{-8pt}
\end{figure*}

In \figref{fig:qualcitysup}, we firstly conduct an additional comparison of the segmentation results on the Cityscapes~\cite{cordts2016cityscapes} dataset.
The baseline model showed weakness to changes in brightness due to shadows or changes in places such as side streets and parking lots in the real world.
In addition, results from all the other methods were damaged to predict objects such as trains or trucks in the real world.
In contrast, our method predicted road, train, truck, and vegetation relatively well, showing robustness to domain change.

\begin{figure*}[t]
	\centering
    \includegraphics[width=0.9\linewidth]{figure/classlabel.pdf}\hfill
    \\
    {\raisebox{15pt}{\rotatebox[origin=c]{90}{\footnotesize Diverse Weathers}\hspace{1.6pt}}}
    \begin{subfigure}{0.1385\linewidth}
      \includegraphics[width=1.0\linewidth]{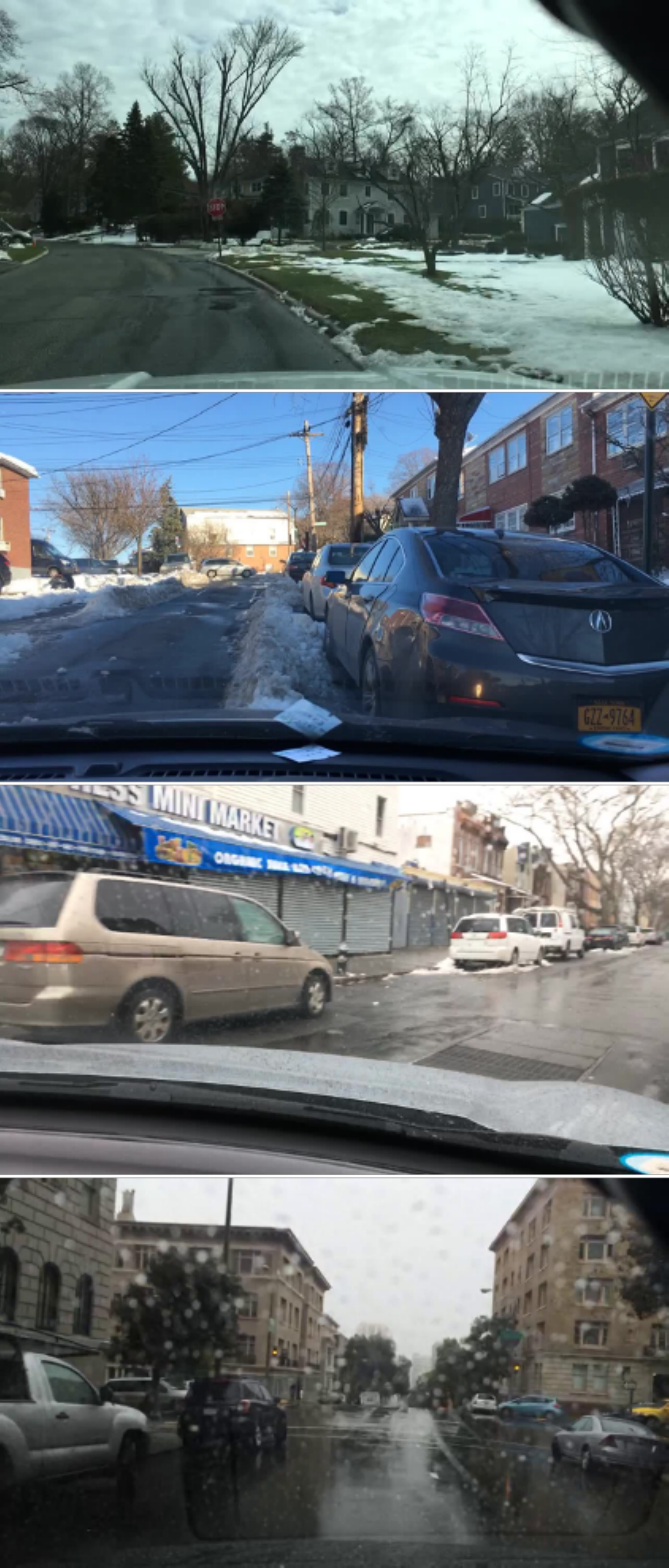}
      \caption{Images}
    \end{subfigure}
    \hspace{-4.5pt}
    \begin{subfigure}{0.1385\linewidth}
      \includegraphics[width=1.0\linewidth]{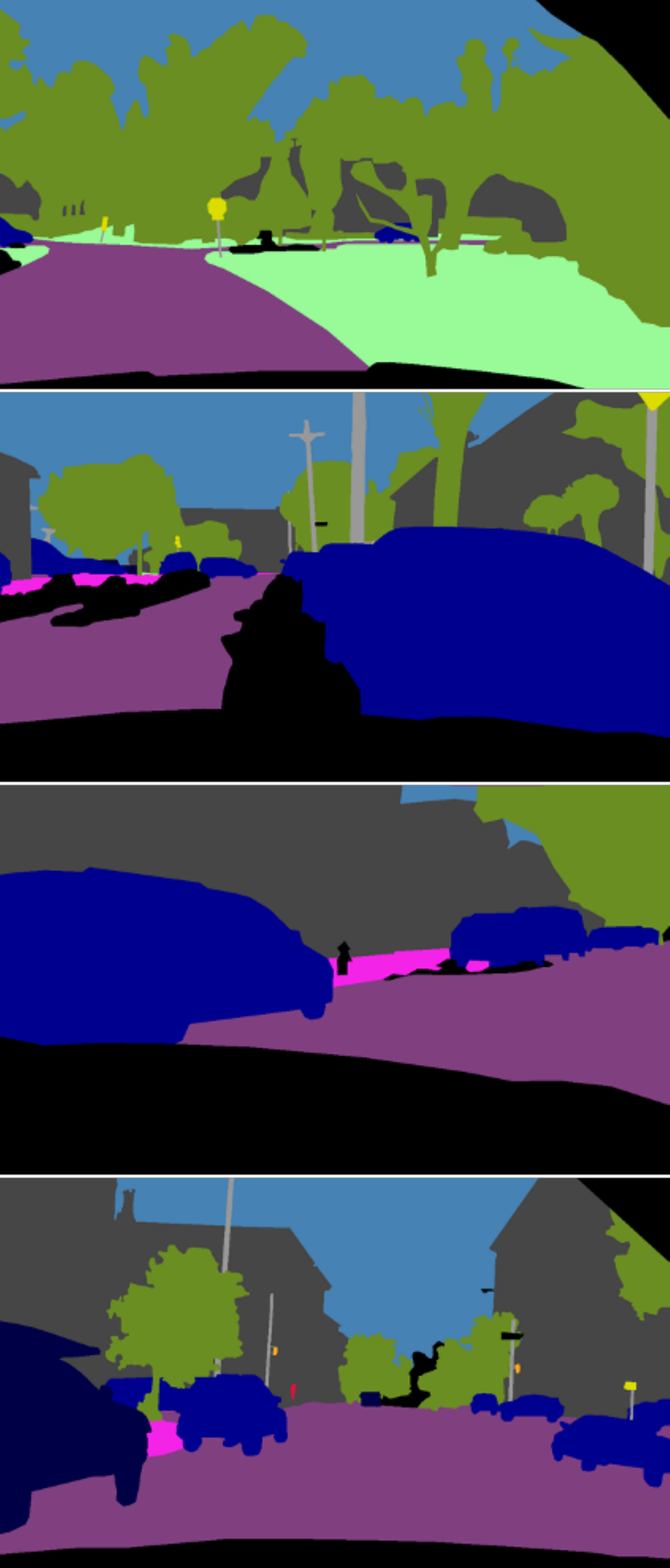}
      \caption{Ground Truth}
    \end{subfigure}
    \hspace{-4.5pt}
    \begin{subfigure}{0.1385\linewidth}
      \includegraphics[width=1.0\linewidth]{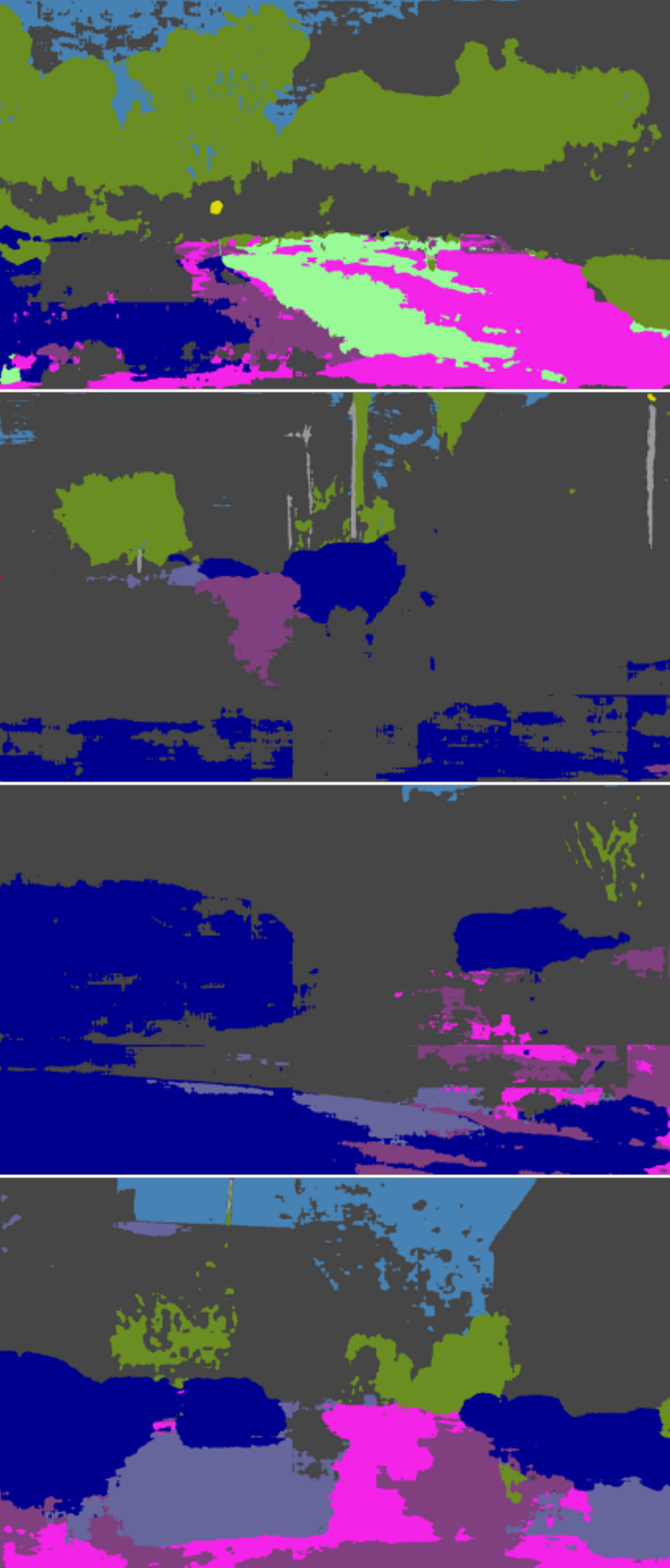}
      \caption{Baseline}
    \end{subfigure}
    \hspace{-4.5pt}
    \begin{subfigure}{0.1382\linewidth}
      \includegraphics[width=1.0\linewidth]{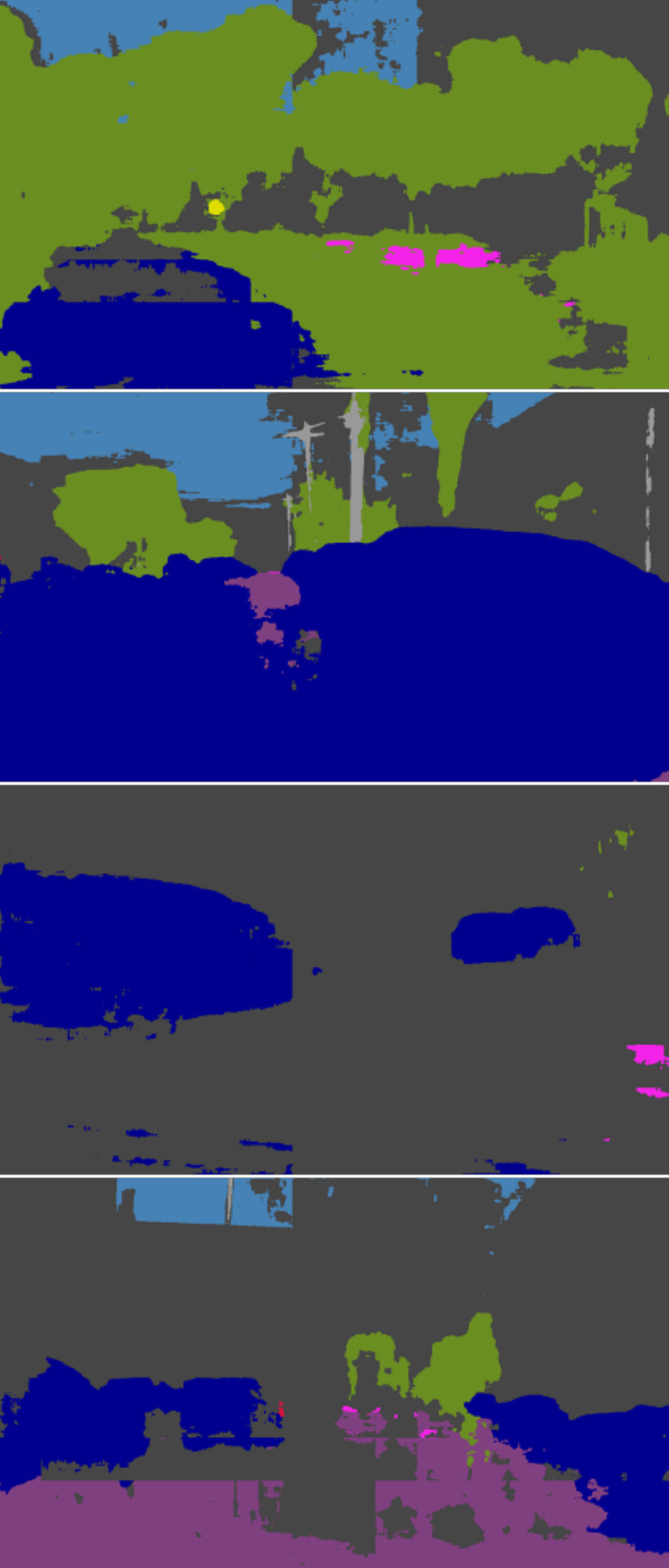}
      \caption{IBN-Net~\cite{pan2018ibnnet}}
    \end{subfigure}
    \hspace{-4.5pt}
    \begin{subfigure}{0.1385\linewidth}
      \includegraphics[width=1.0\linewidth]{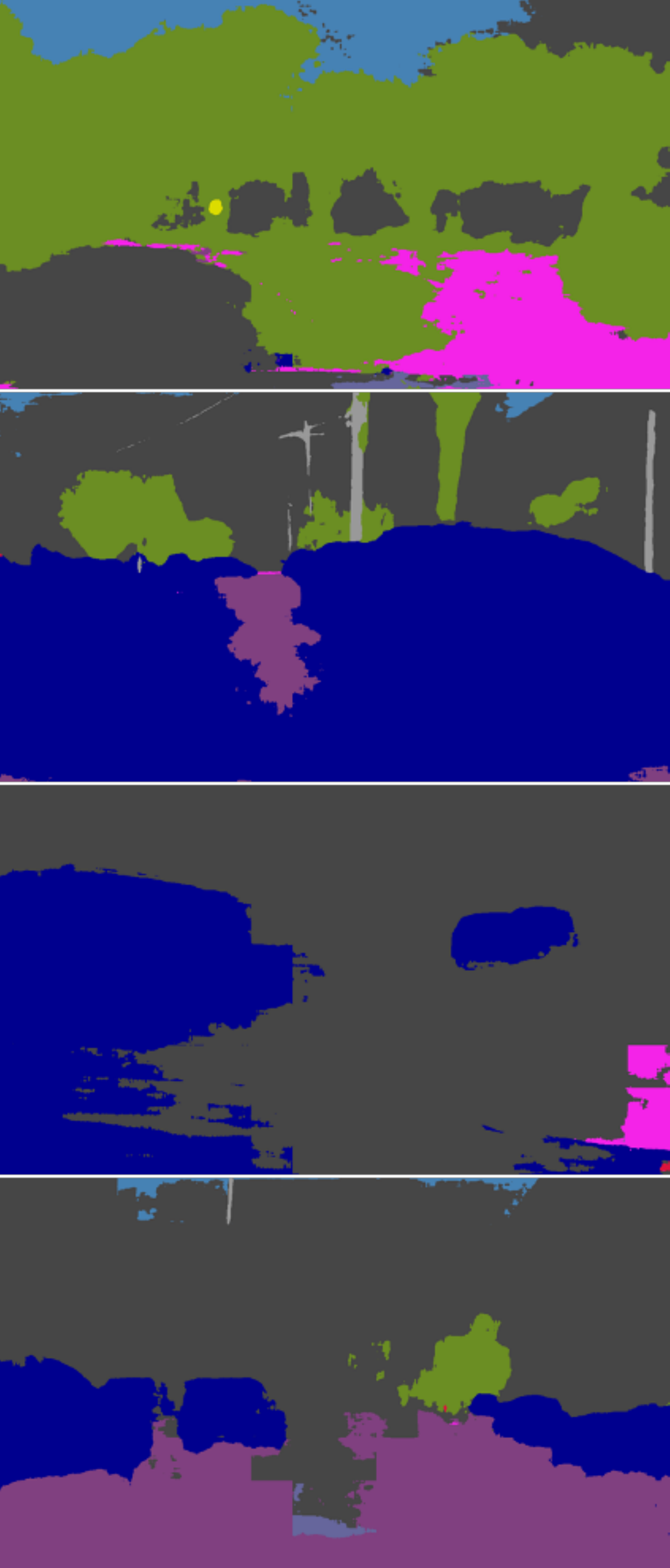}
      \caption{RobustNet~\cite{choi2021robustnet}}
    \end{subfigure}
    \hspace{-4.5pt}
    \begin{subfigure}{0.1385\linewidth}
      \includegraphics[width=1.0\linewidth]{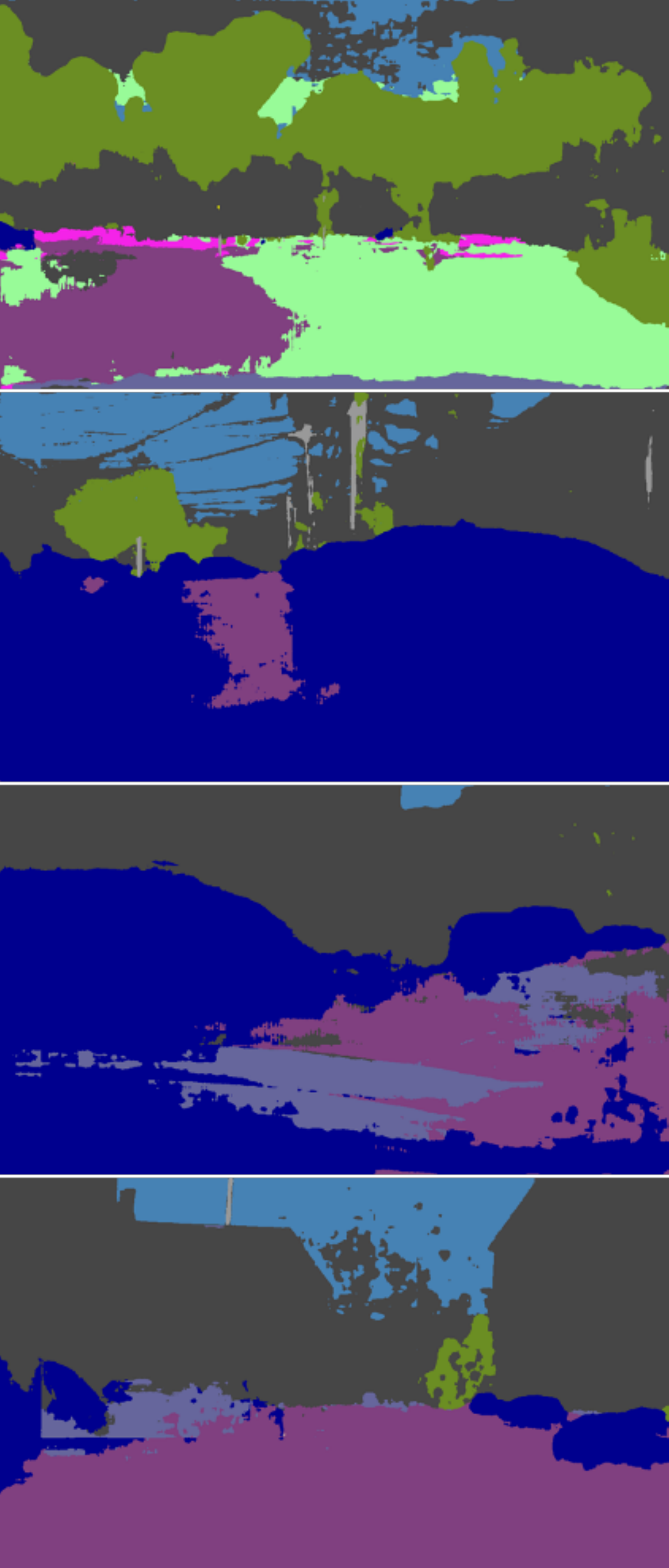}
      \caption{MLDG~\cite{mldg}}
    \end{subfigure}
    \hspace{-4.5pt}
    \begin{subfigure}{0.1385\linewidth}
      \includegraphics[width=1.0\linewidth]{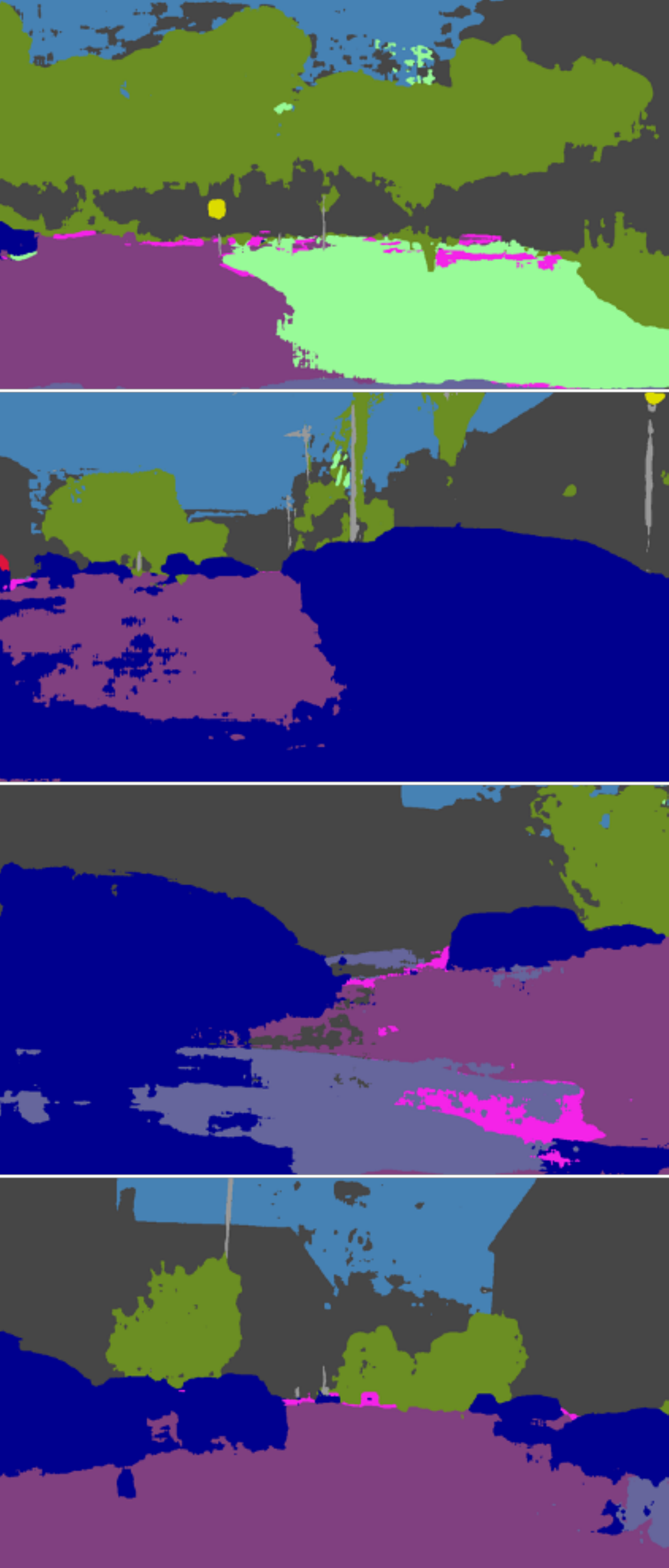}
      \caption{Ours}
    \end{subfigure}

    \vspace{-7pt}
 	\caption{\textbf{Source (G+S)$\rightarrow$Target (B):} [1/2] Qualitative comparison on the BDD100K dataset. All methods adopt DeepLabV3+ with ResNet50. (Best viewed in color.)
 	}\label{fig:qualbdd1}\vspace{-7pt}
\end{figure*}

\begin{figure*}[!ht]
	\centering
    \includegraphics[width=0.9\linewidth]{figure/classlabel.pdf}\hfill
    \\
    {\raisebox{80pt}{\rotatebox[origin=c]{90}{\centering\footnotesize Illumination and Reflection}}\hspace{1.6pt}}
    \includegraphics[width=0.1387\linewidth]{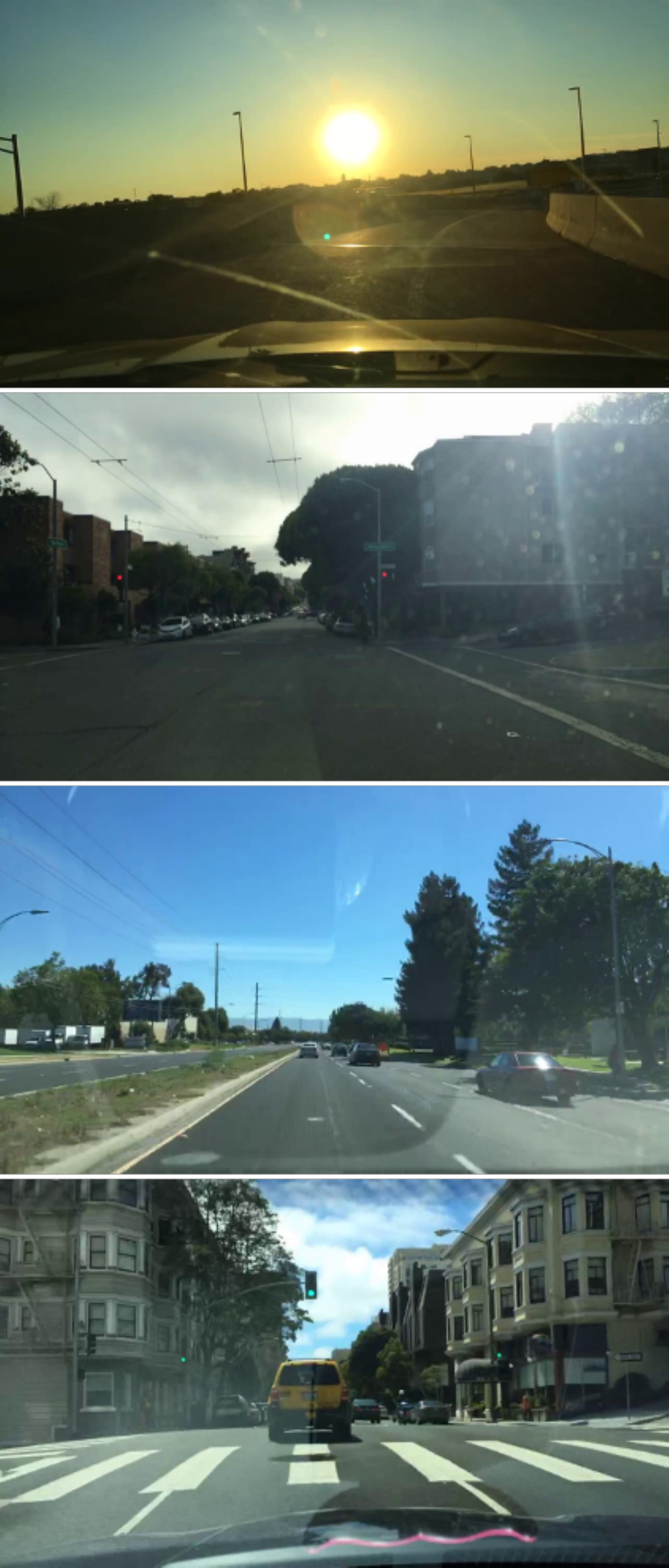}
    \hspace{-4.5pt}
    \includegraphics[width=0.1388\linewidth]{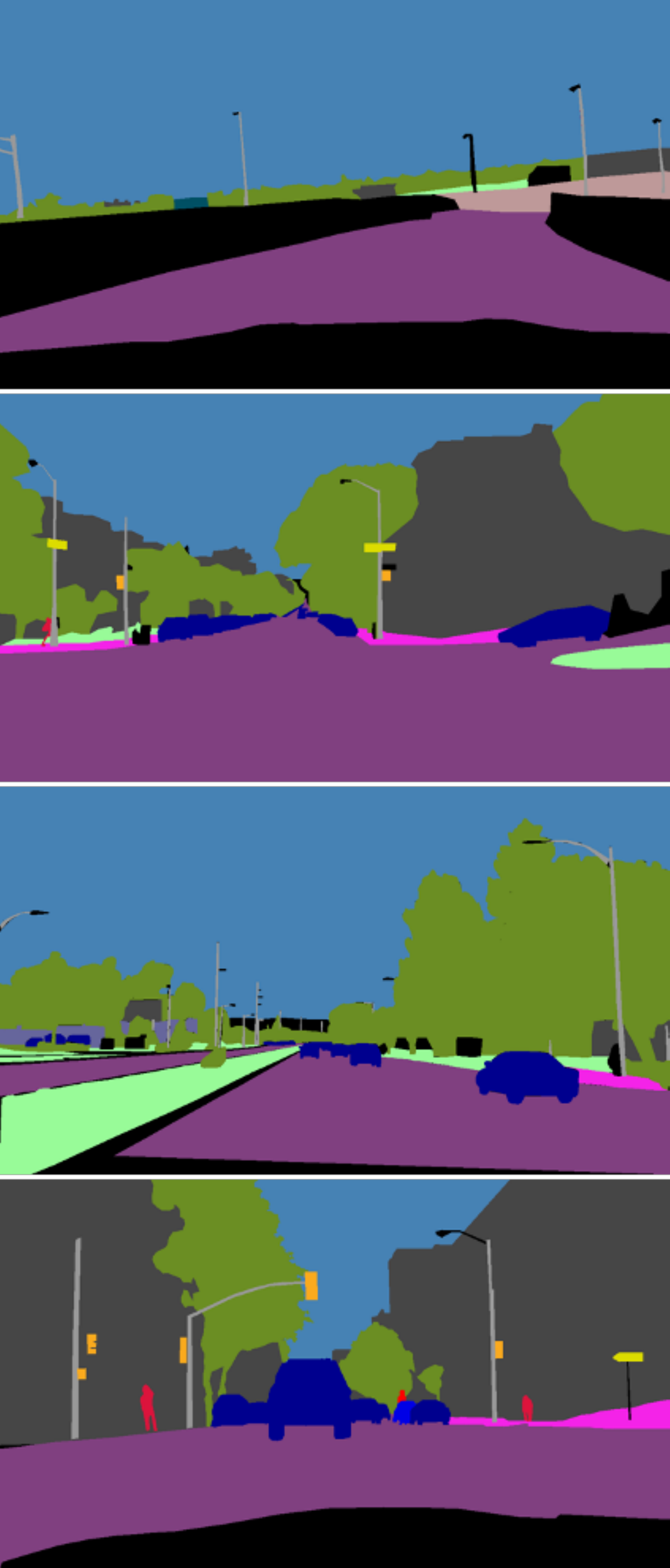}
    \hspace{-4.5pt}
    \includegraphics[width=0.1388\linewidth]{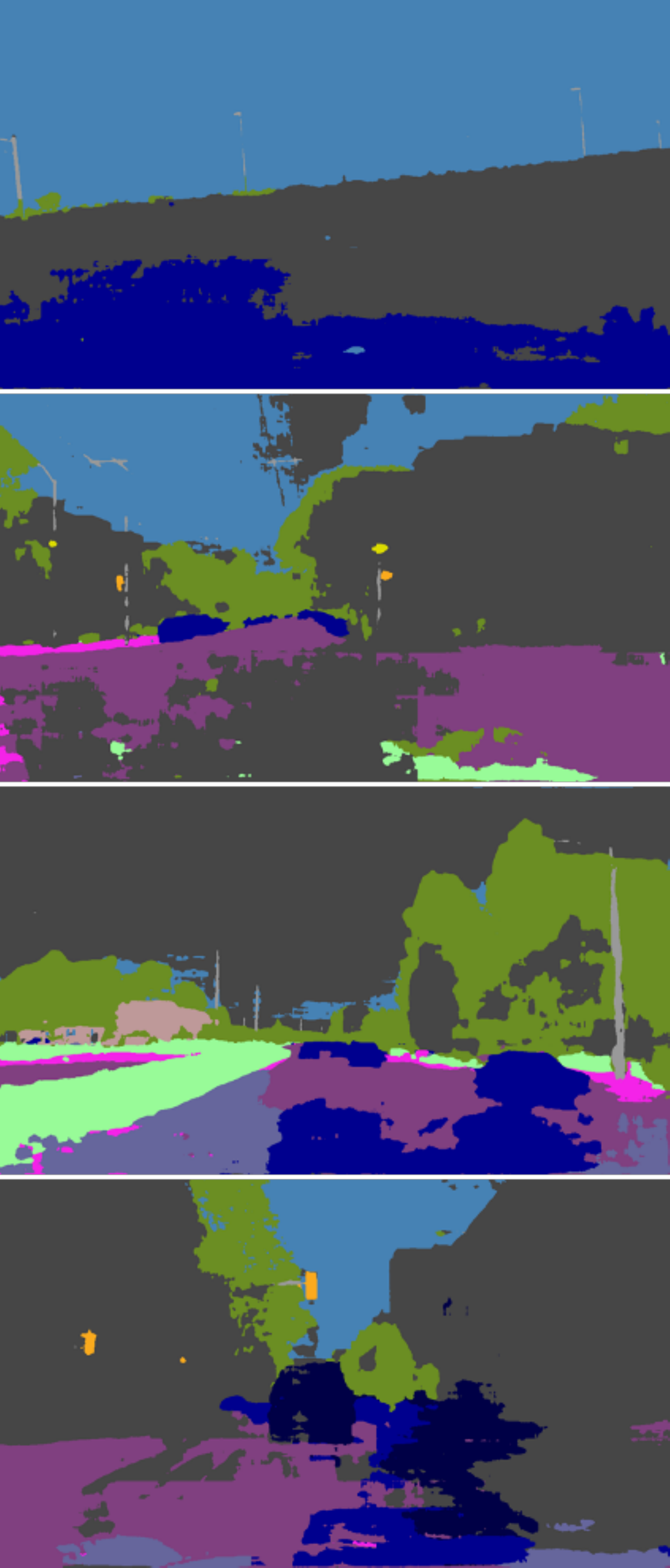}
    \hspace{-4.5pt}
    \includegraphics[width=0.1385\linewidth]{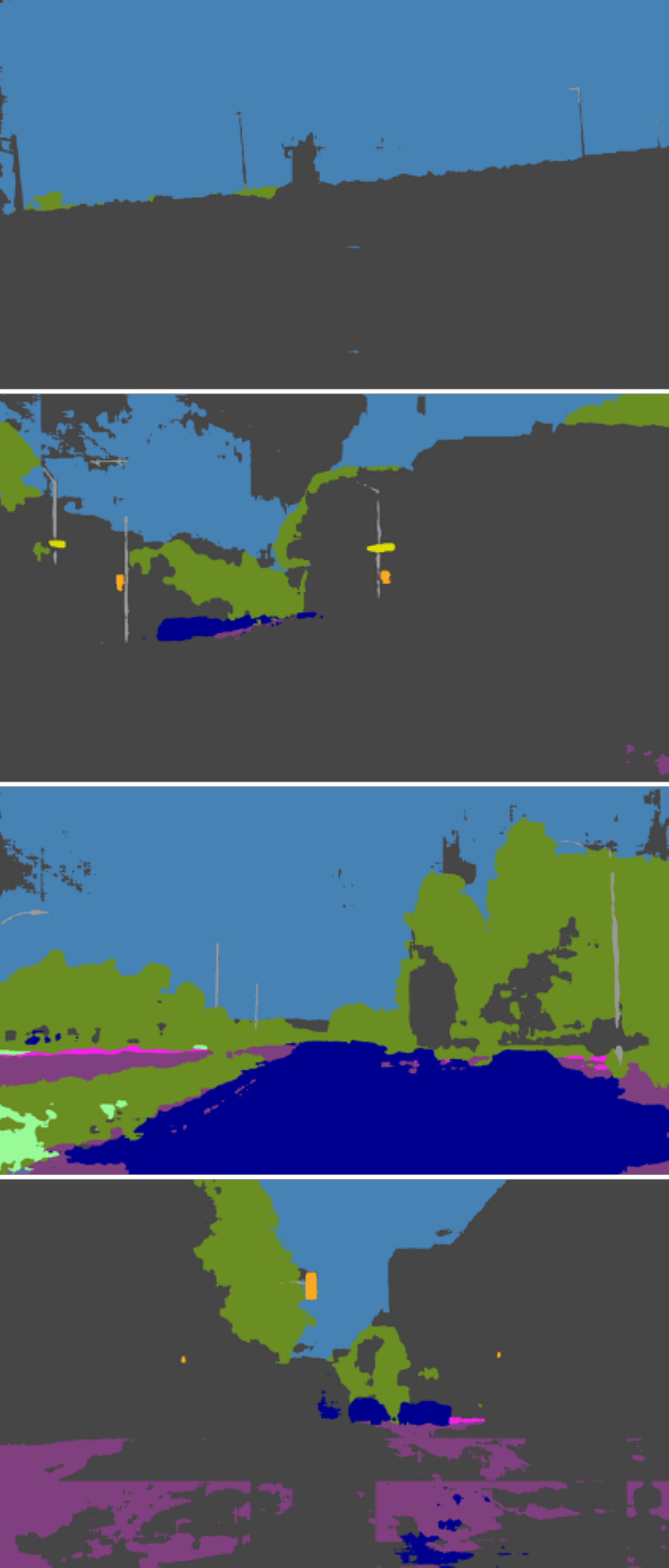}
    \hspace{-4.5pt}
    \includegraphics[width=0.1388\linewidth]{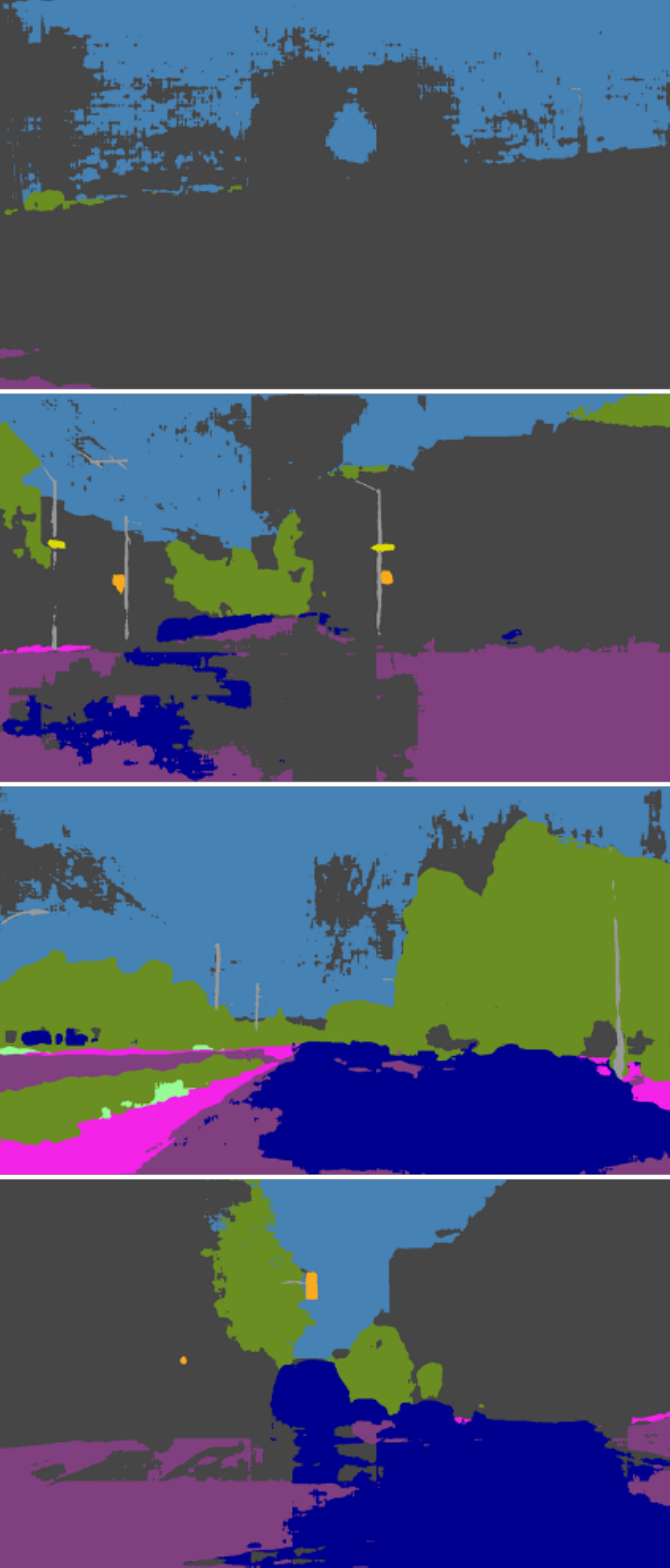}
    \hspace{-4.5pt}
    \includegraphics[width=0.1385\linewidth]{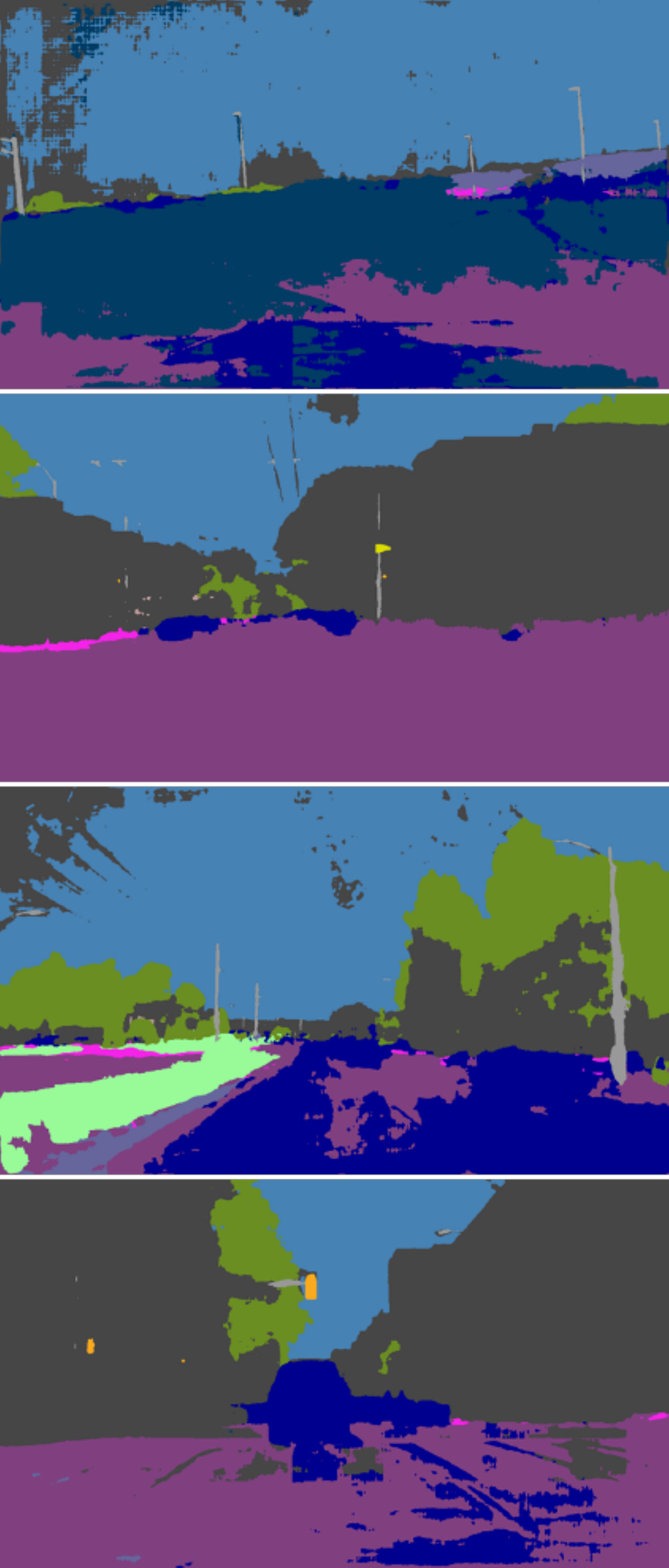}
    \hspace{-4.5pt}
    \includegraphics[width=0.1385\linewidth]{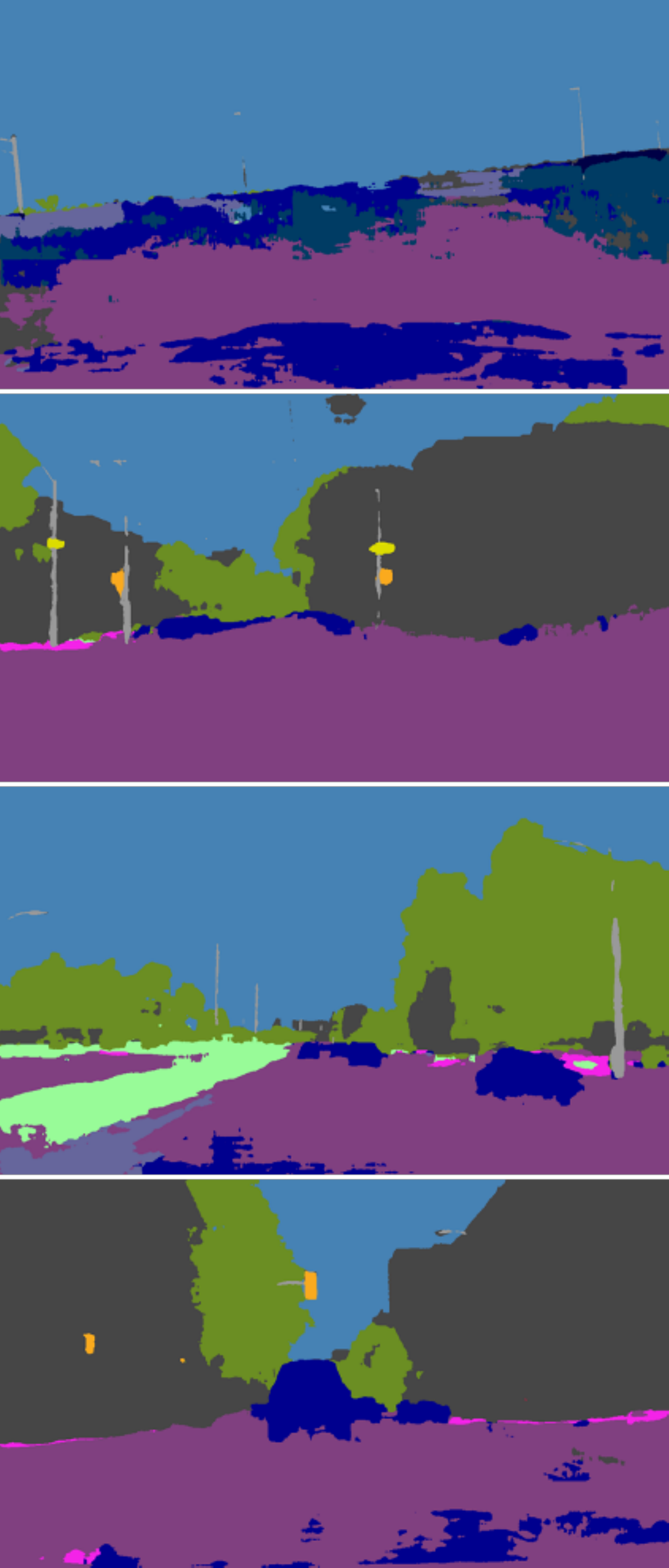}
    \\
    \vspace{4pt}
	{\raisebox{80pt}{\rotatebox[origin=c]{90}{\footnotesize Time Variation}\hspace{1.6pt}}}
    \includegraphics[width=0.1388\linewidth]{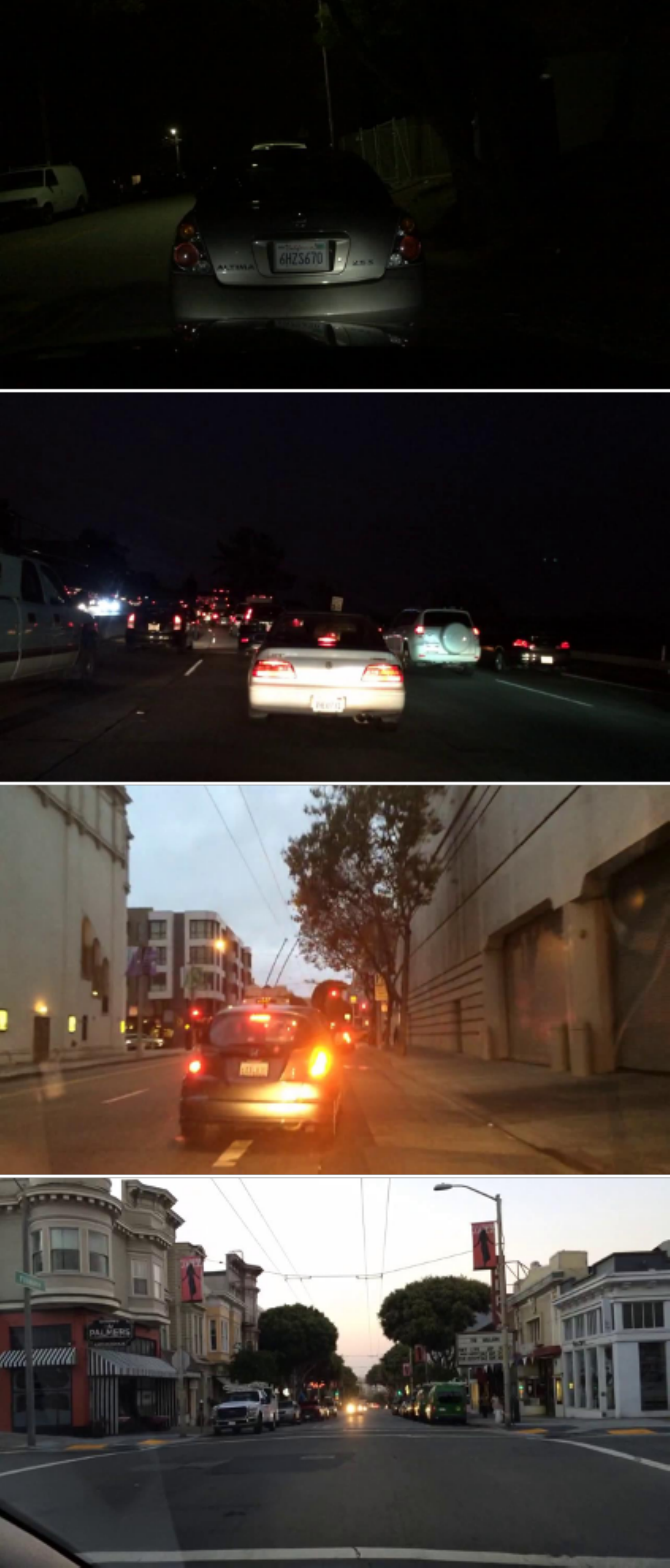}
    \hspace{-4.5pt}
    \includegraphics[width=0.1388\linewidth]{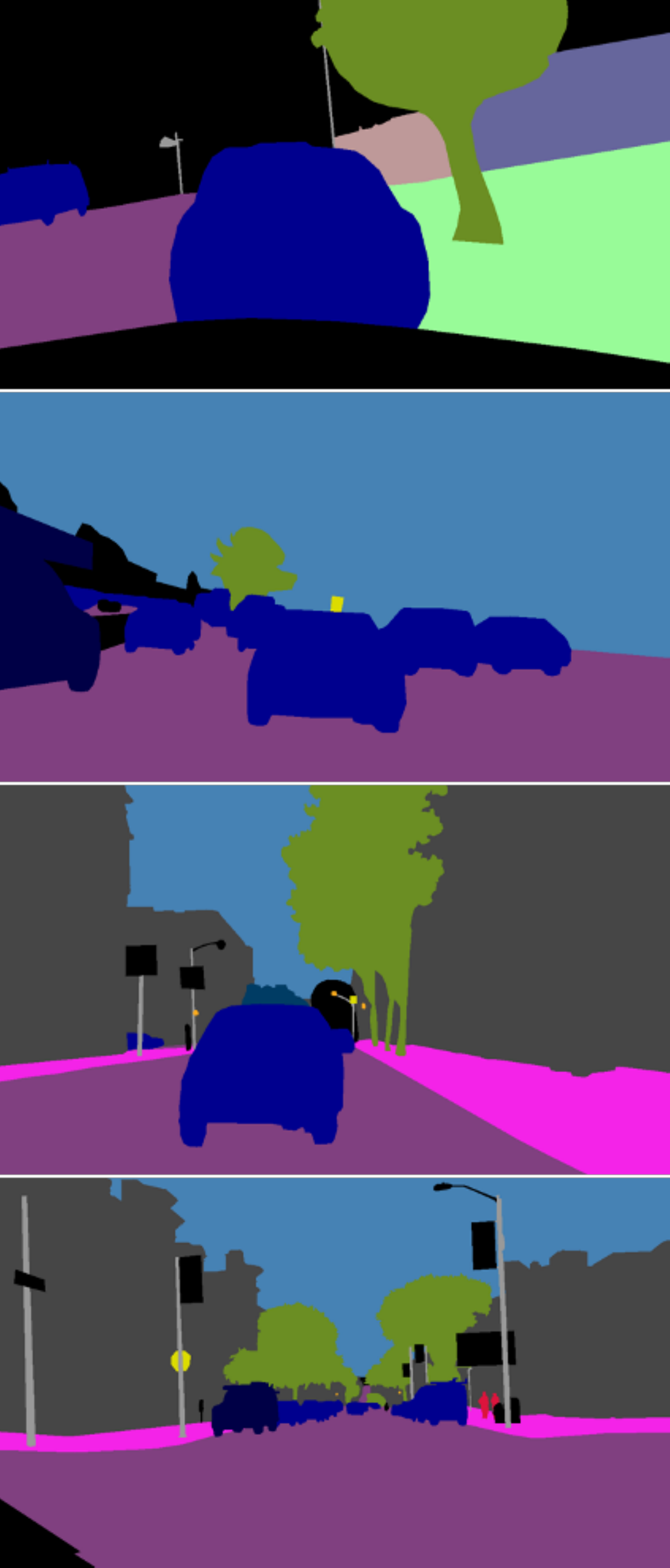}
    \hspace{-4.5pt}
    \includegraphics[width=0.1385\linewidth]{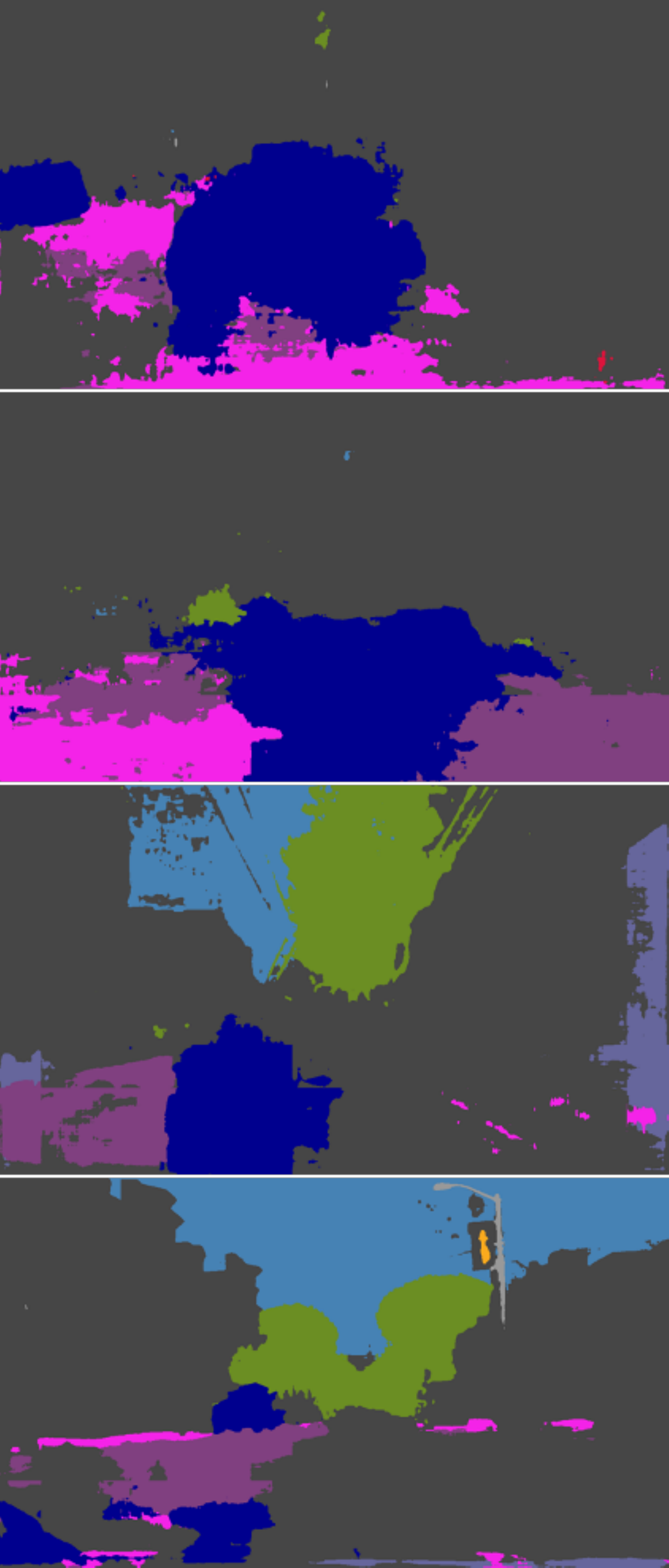}
    \hspace{-4.5pt}
    \includegraphics[width=0.1388\linewidth]{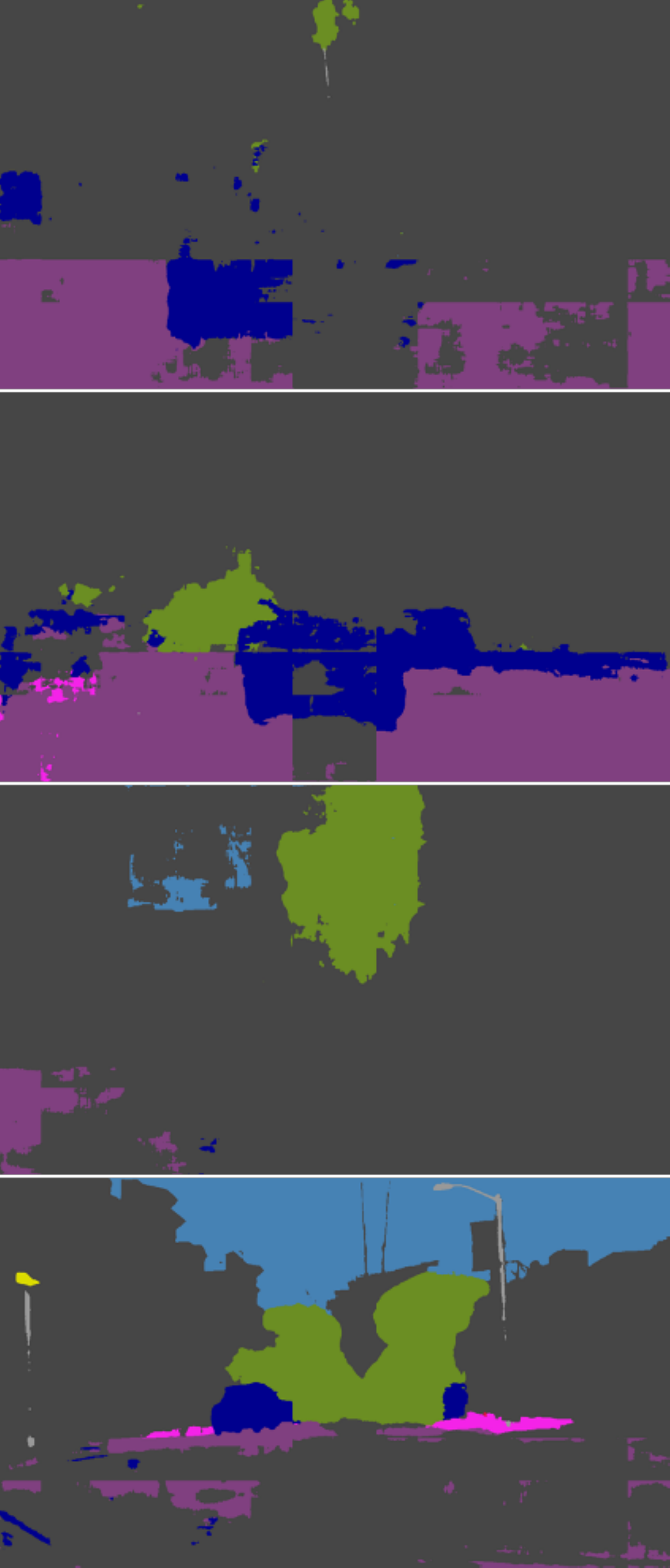}
    \hspace{-4.5pt}
    \includegraphics[width=0.1385\linewidth]{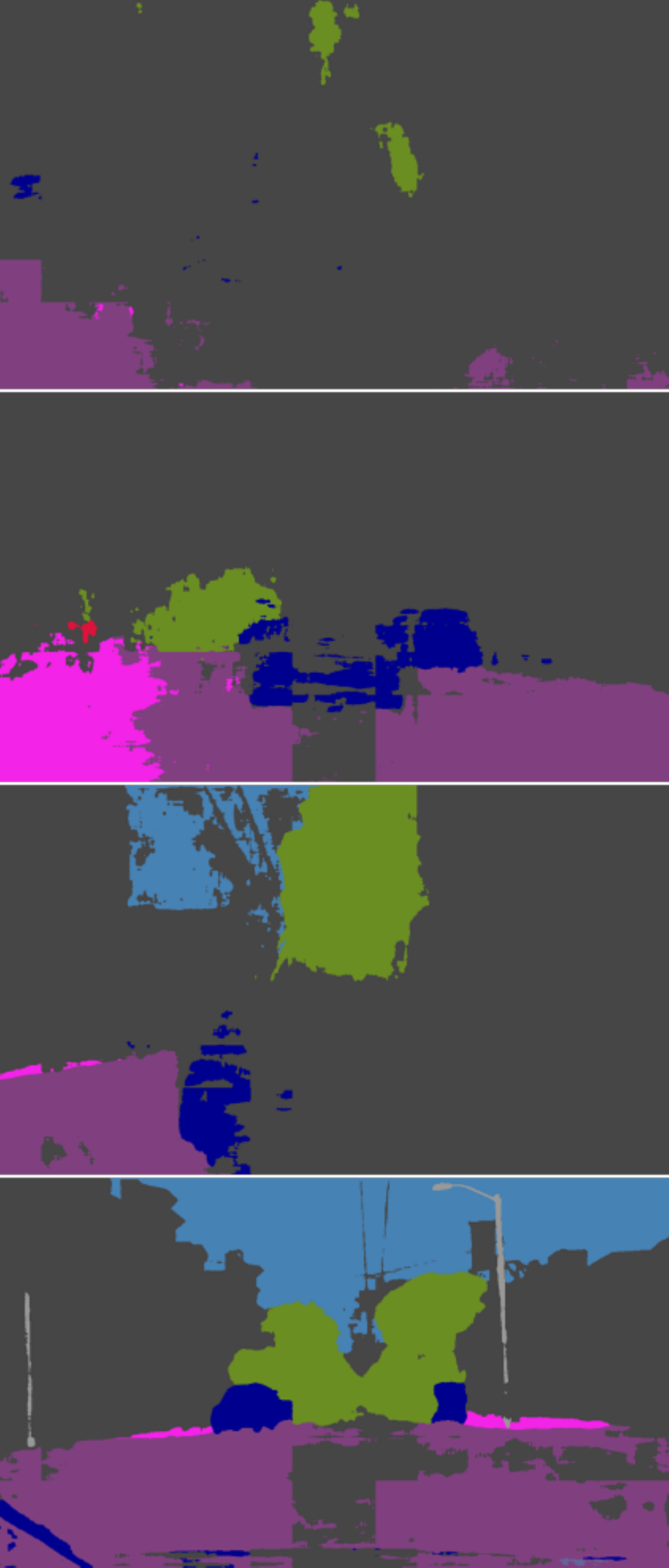}
    \hspace{-4.5pt}
    \includegraphics[width=0.1385\linewidth]{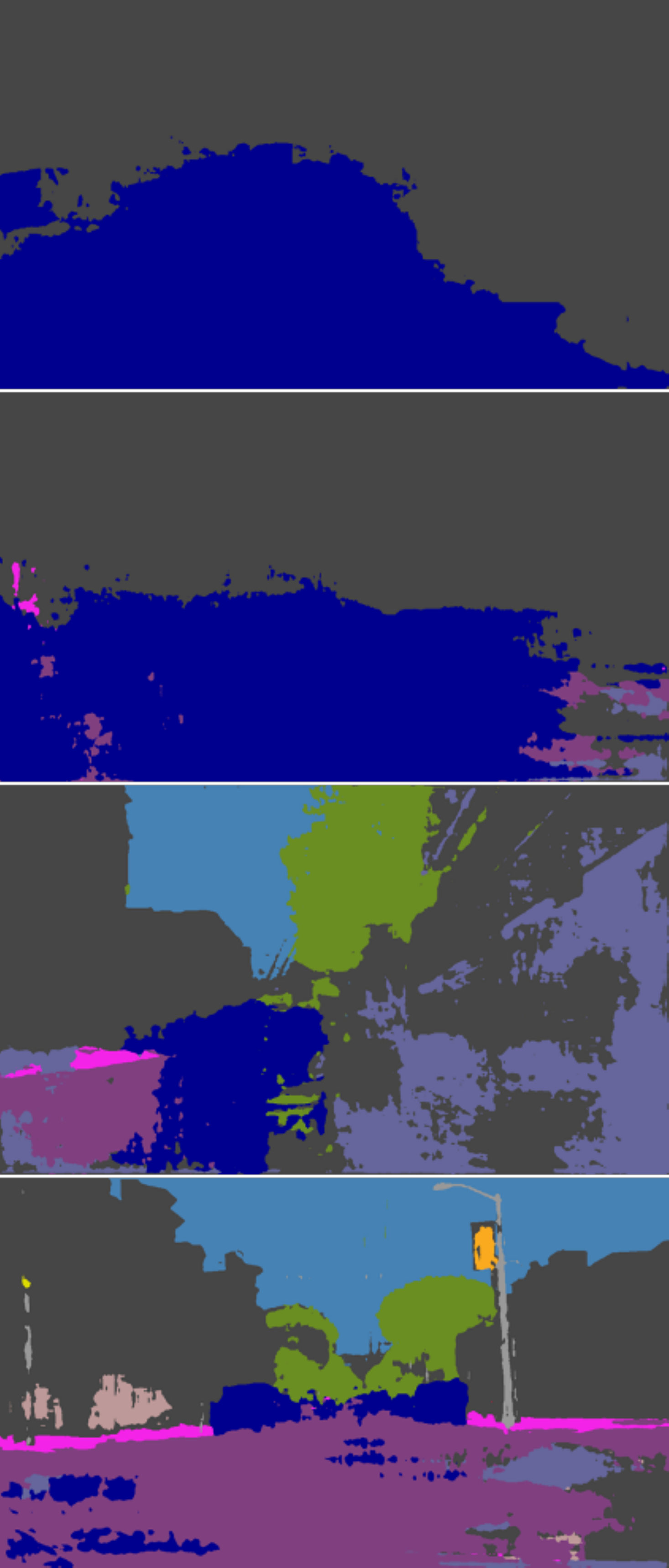}
    \hspace{-4.5pt}
    \includegraphics[width=0.1388\linewidth]{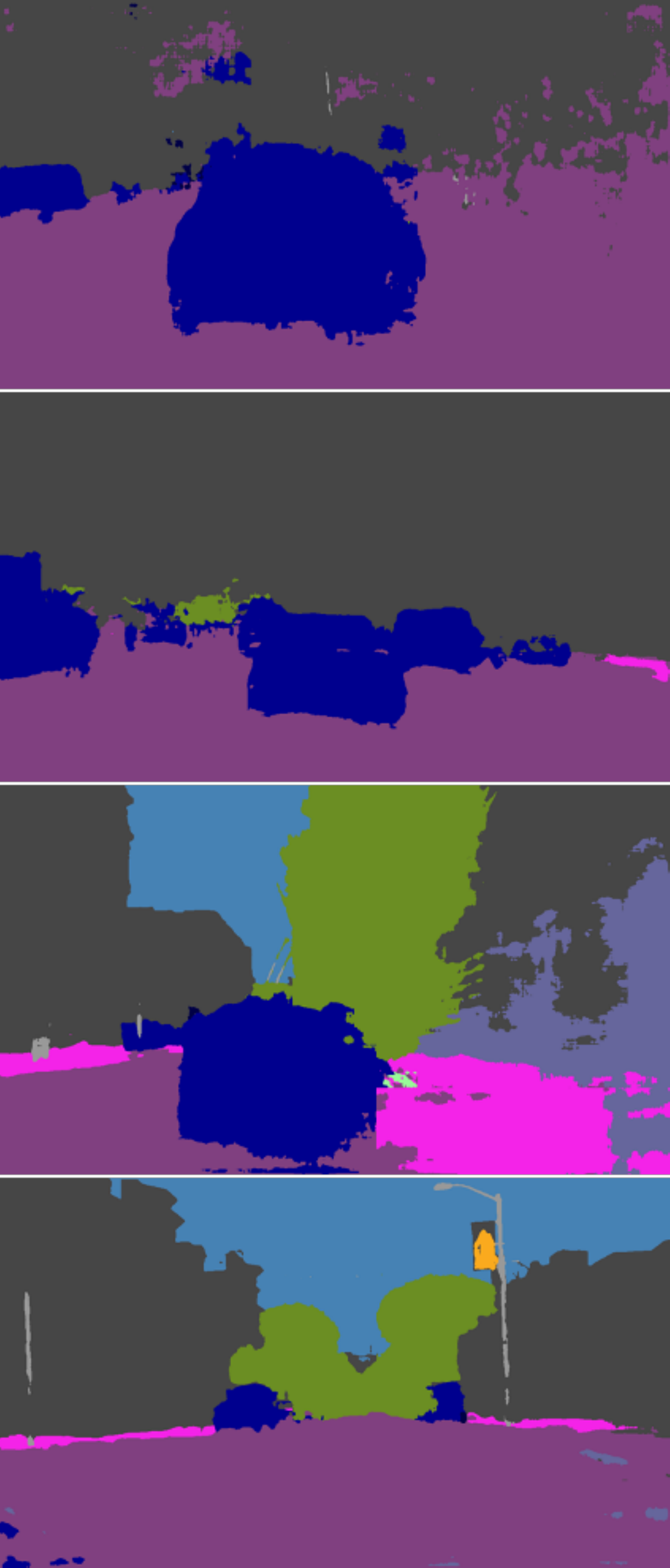}
    \\
    \vspace{4pt}
	{\raisebox{15pt}{\rotatebox[origin=c]{90}{\footnotesize Unseen Structures and Slope}\hspace{0.8pt}}}
	\begin{subfigure}{0.1387\linewidth}
      \includegraphics[width=1.0\linewidth]{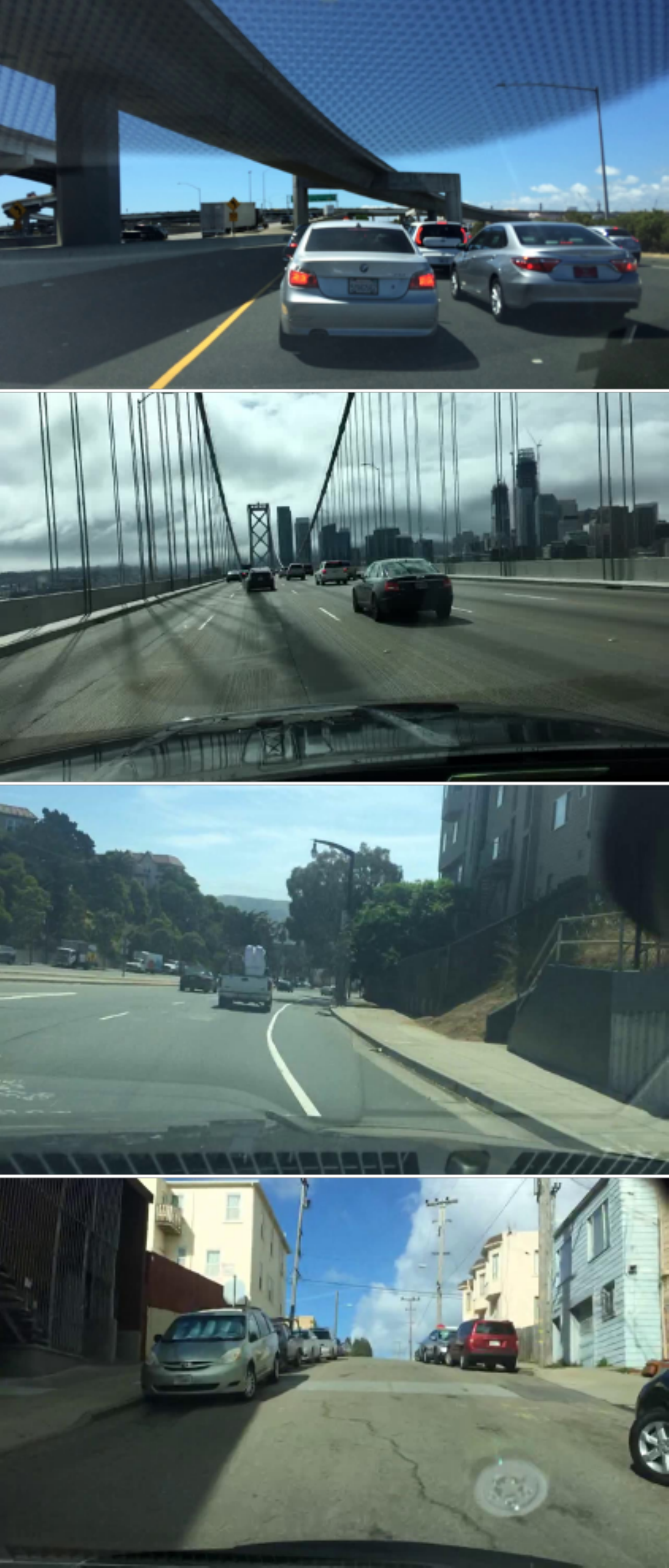}
      \caption{Images}
    \end{subfigure}
    \hspace{-4.5pt}
    \begin{subfigure}{0.1388\linewidth}
      \includegraphics[width=1.0\linewidth]{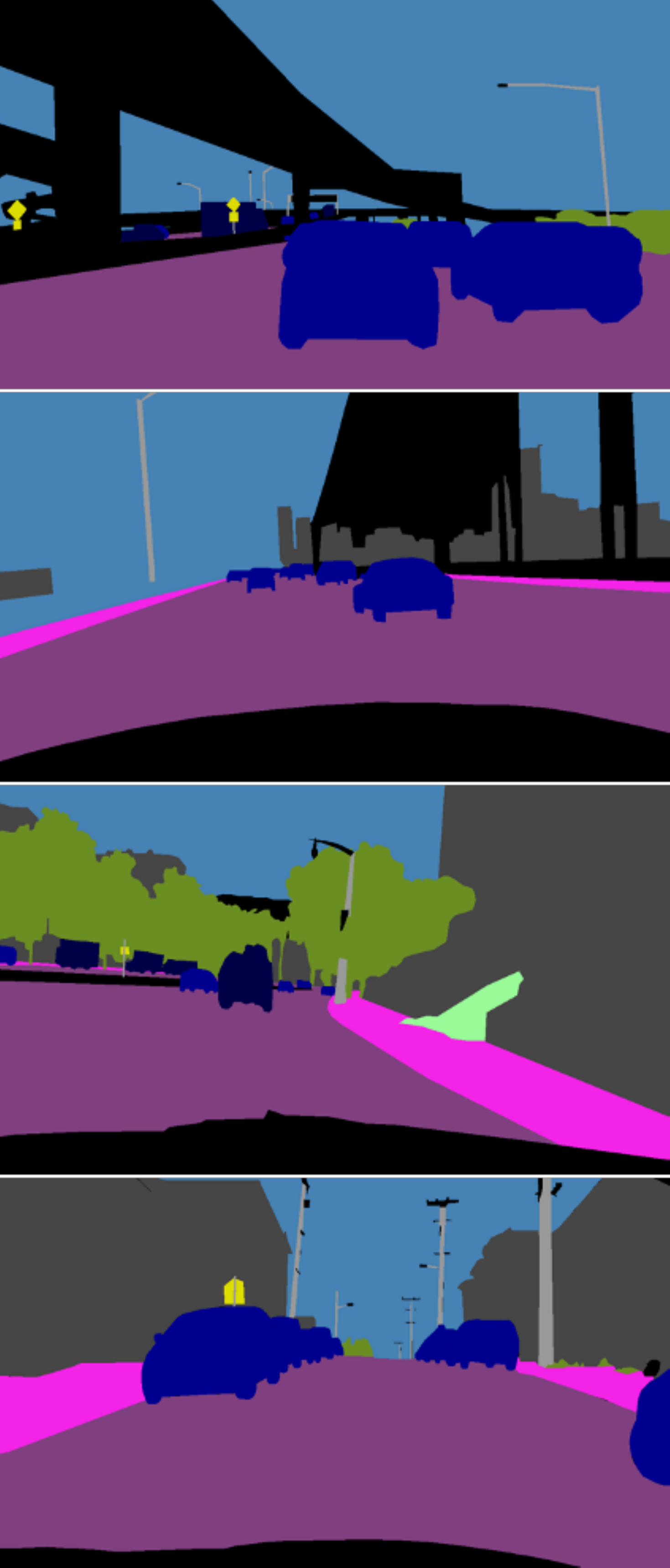}
      \caption{Ground Truth}
    \end{subfigure}
    \hspace{-4.5pt}
    \begin{subfigure}{0.1385\linewidth}
      \includegraphics[width=1.0\linewidth]{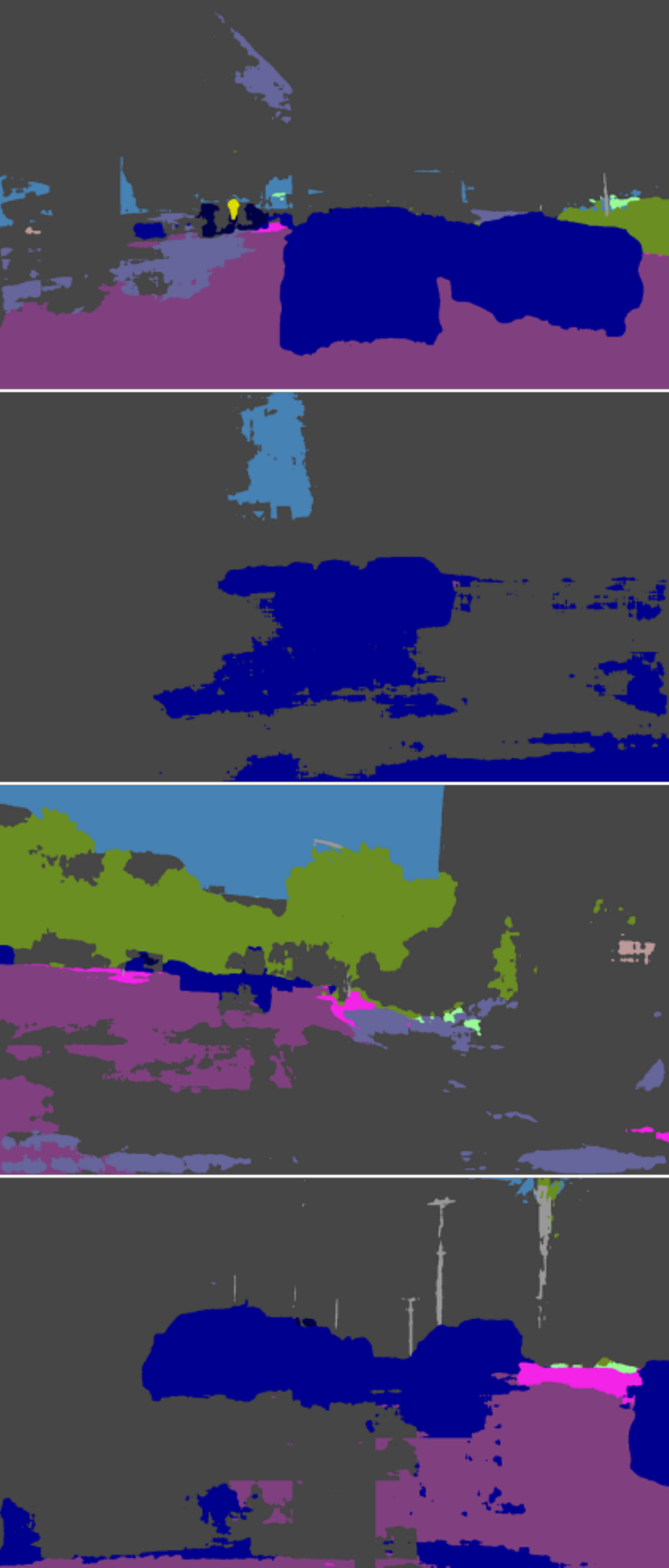}
      \caption{Baseline}
    \end{subfigure}
    \hspace{-4.5pt}
    \begin{subfigure}{0.1388\linewidth}
      \includegraphics[width=1.0\linewidth]{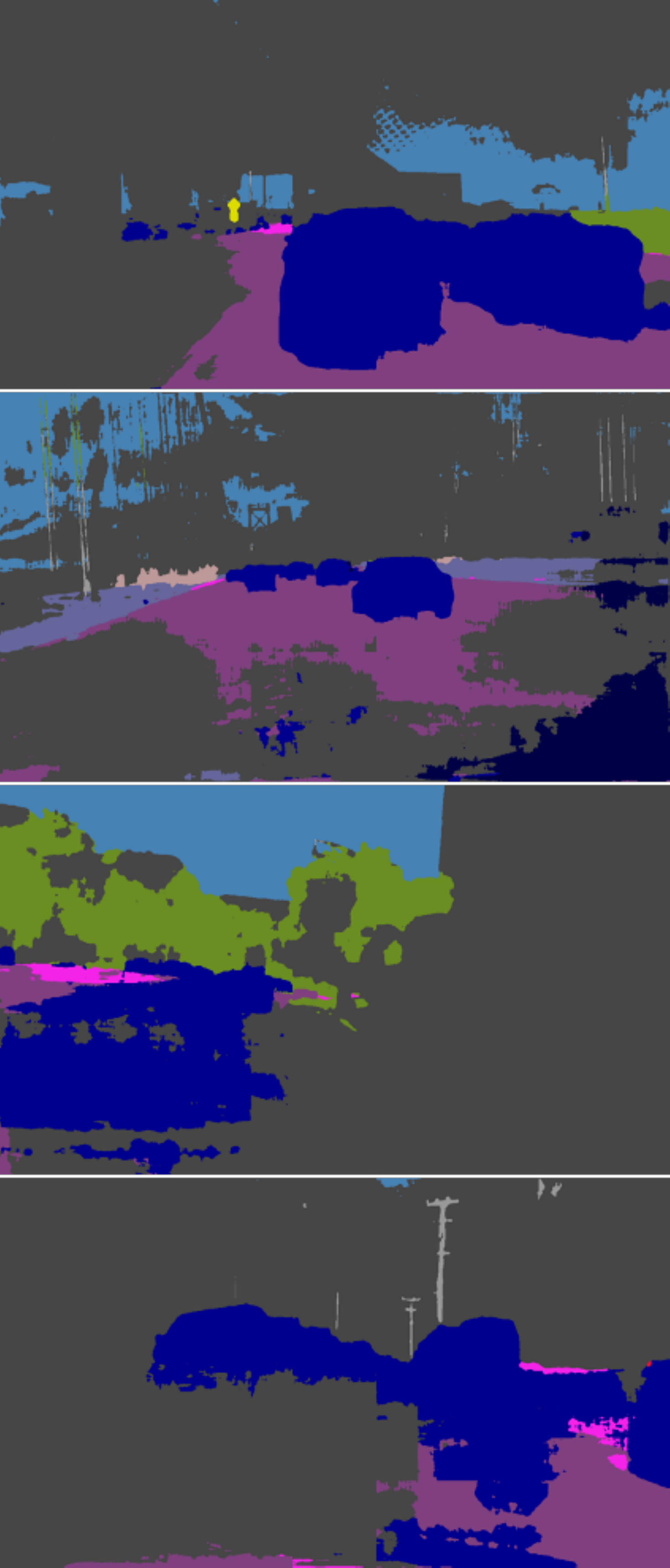}
      \caption{IBN-Net~\cite{pan2018ibnnet}}
    \end{subfigure}
    \hspace{-4.5pt}
    \begin{subfigure}{0.1388\linewidth}
      \includegraphics[width=1.0\linewidth]{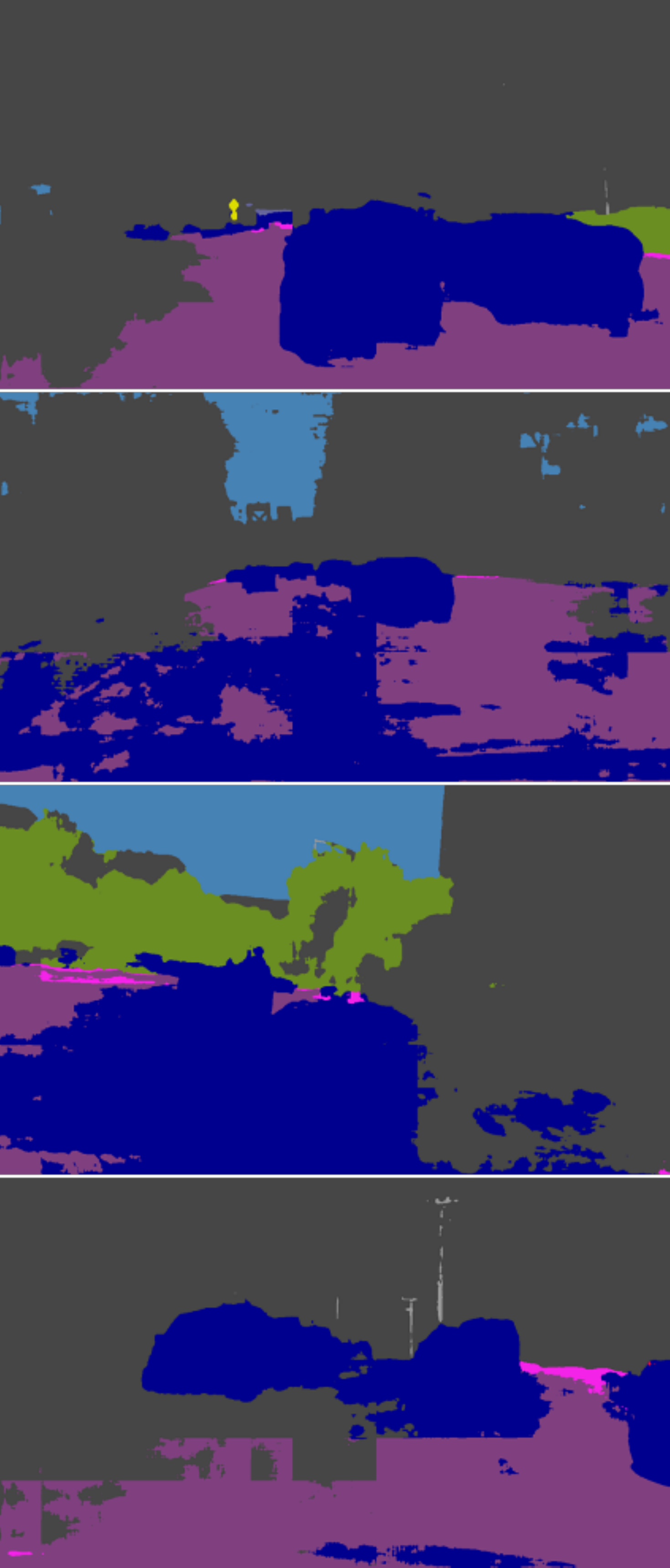}
      \caption{RobustNet~\cite{choi2021robustnet}}
    \end{subfigure}
    \hspace{-4.5pt}
    \begin{subfigure}{0.1385\linewidth}
      \includegraphics[width=1.0\linewidth]{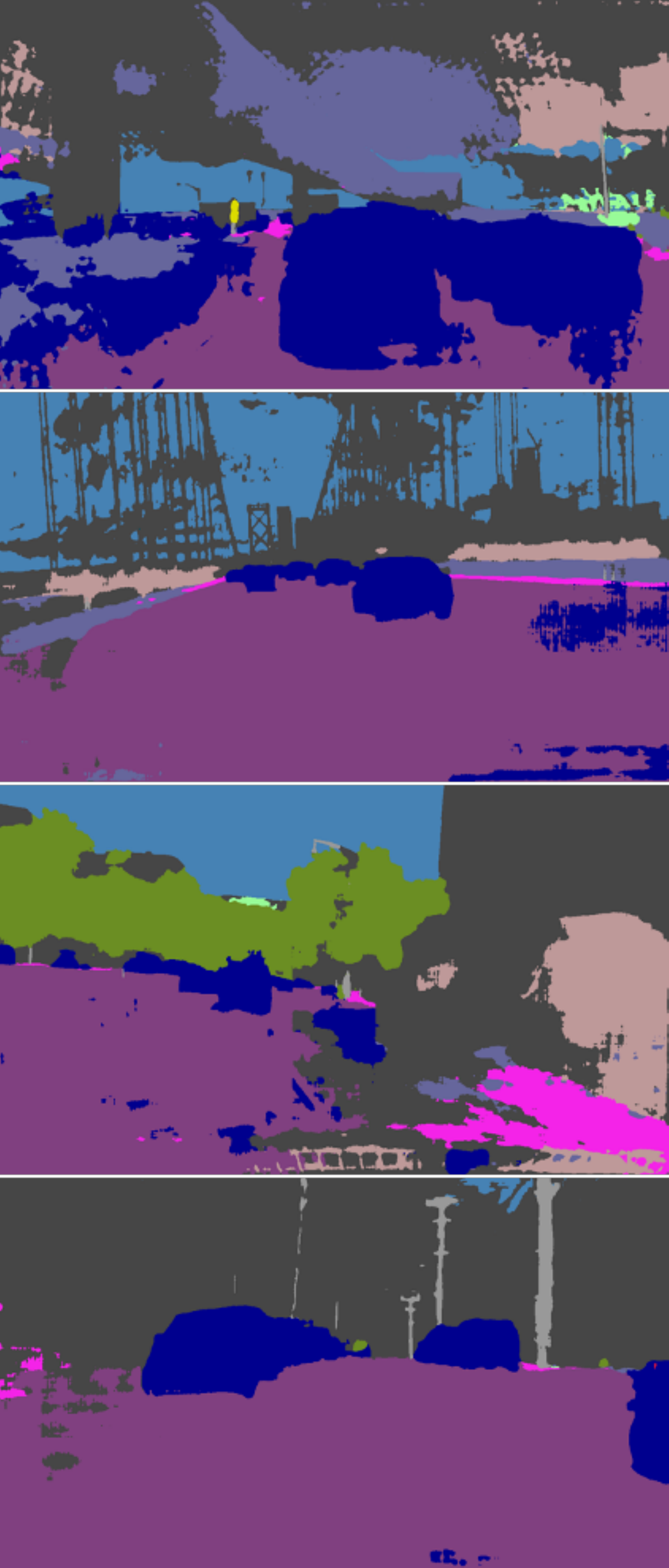}
      \caption{MLDG~\cite{mldg}}
    \end{subfigure}
    \hspace{-4.5pt}
    \begin{subfigure}{0.1388\linewidth}
      \includegraphics[width=1.0\linewidth]{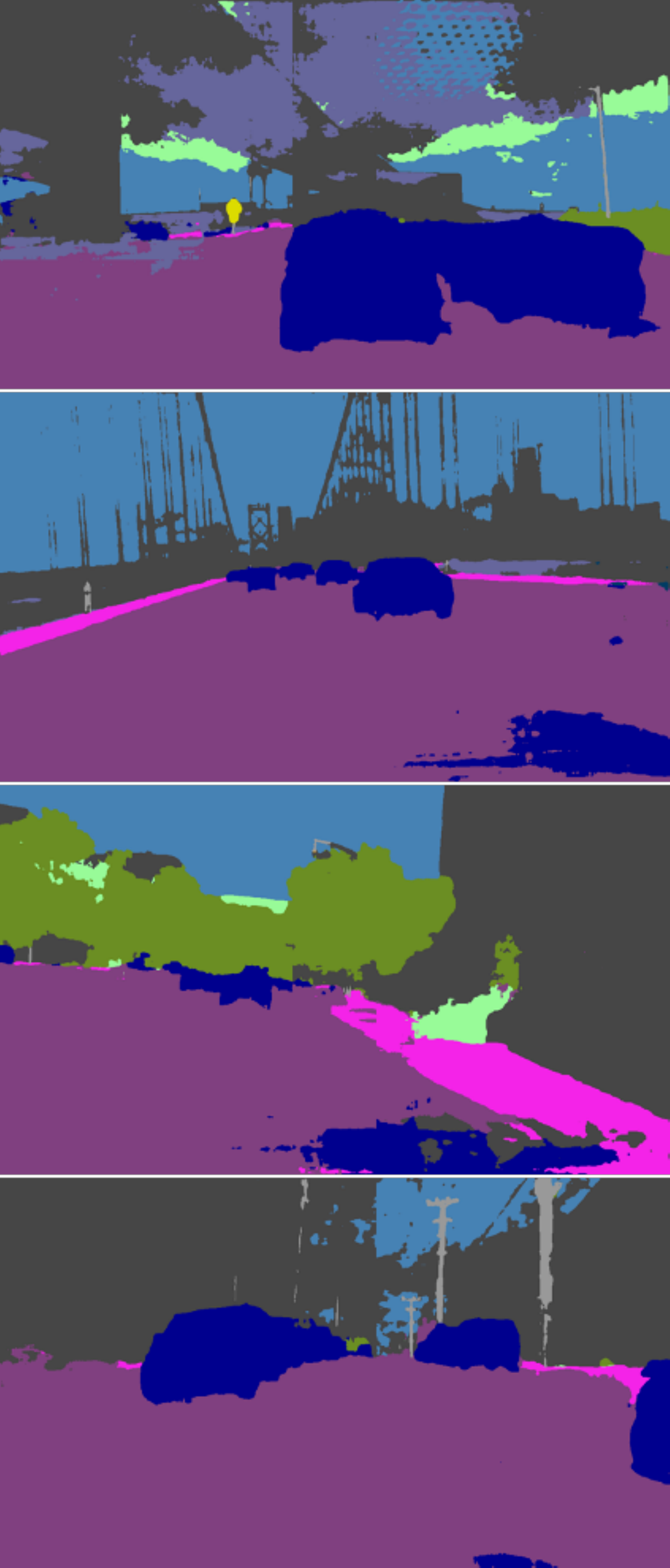}
      \caption{Ours}
    \end{subfigure}
	
    \vspace{-7pt}
 	\caption{
 	\textbf{Source (G+S)$\rightarrow$Target (B):} [2/2] Qualitative comparison on the BDD100K dataset. All methods adopt DeepLabV3+ with ResNet50. (Best viewed in color.)
 	}\label{fig:qualbdd2}\vspace{-7pt}
\end{figure*}

\figref{fig:qualbdd1} and \figref{fig:qualbdd2} show the predicted segmentation results on the BDD100K dataset.
Compared to the Cityscapes dataset that only contains images acquired primarily in daytime and relatively simple weather conditions (\ie overcast or sunny), the BDD100K includes images acquired in various weather conditions, time zones, and locations.
To compare the performance with regard to the variants of weather conditions, in \figref{fig:qualbdd1}, we selected the images taken in snowy or rainy weather conditions, and the baseline showed the vulnerable performance to this change.
The normalization-based and vanilla meta-learning-based methods were also sensitive to visual changes in the road or sky caused by snow and rain.
In contrast, our method predicted less damaged maps and showed reasonably estimation results on roads, sky, and vegetation regions.
\figref{fig:qualbdd2} shows the segmentation results under illumination and time changes.
In addition, \figref{fig:qualbdd2} shows the prediction maps under object visual changes due to the reflection of car glass, road slope, or unseen structures.
To sum up, our method showed more robust results with respect to various visual changes existing in the real world than other DG methods.

\begin{figure*}[!ht]
	\centering
	\includegraphics[width=0.9\linewidth]{figure/classlabel.pdf}
	\\
    \begin{subfigure}{0.1428\linewidth}
      \includegraphics[width=1.0\linewidth]{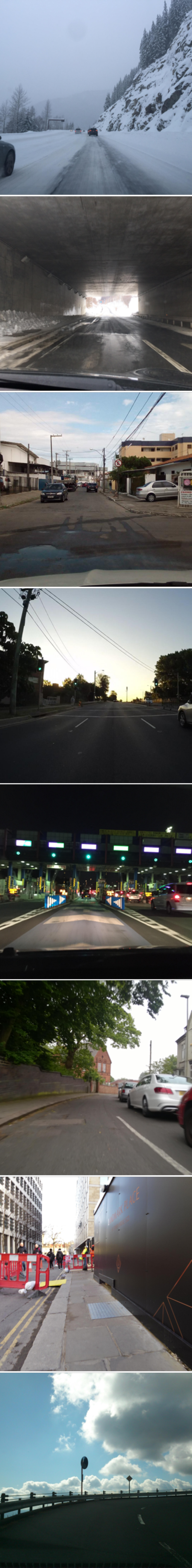}
      \caption{Images}
    \end{subfigure}
    \hspace{-4.5pt}
    \begin{subfigure}{0.1428\linewidth}
      \includegraphics[width=1.0\linewidth]{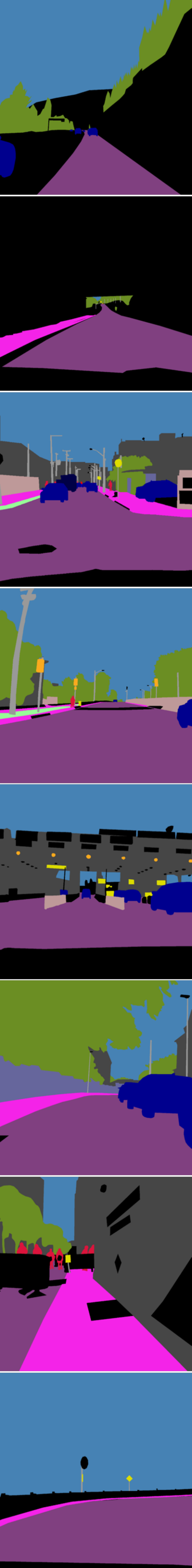}
      \caption{Ground Truth}
    \end{subfigure}
    \hspace{-4.5pt}
    \begin{subfigure}{0.1428\linewidth}
      \includegraphics[width=1.0\linewidth]{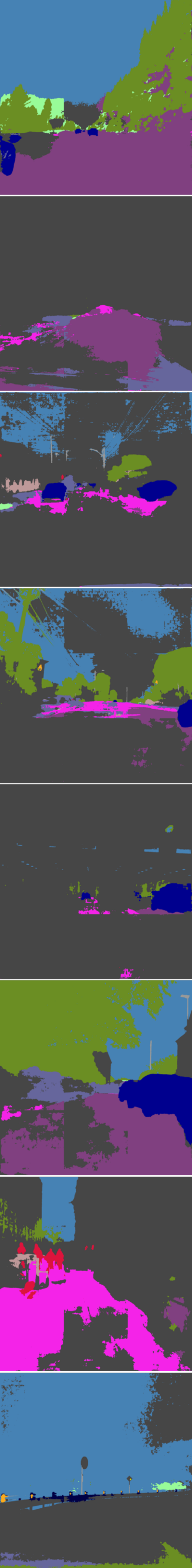}
      \caption{Baseline}
    \end{subfigure}
    \hspace{-4.5pt}
    \begin{subfigure}{0.1428\linewidth}
      \includegraphics[width=1.0\linewidth]{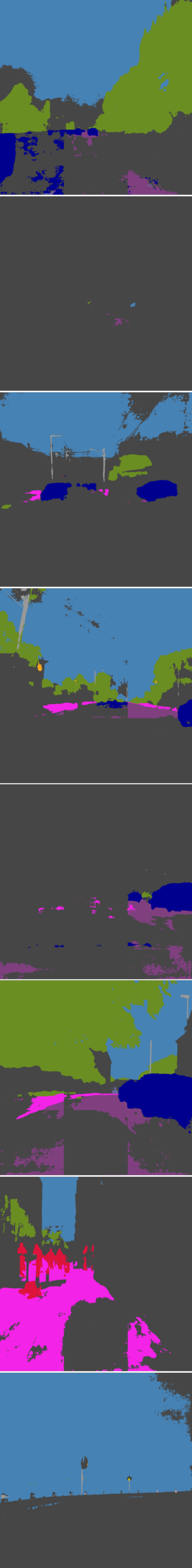}
      \caption{IBN-Net~\cite{pan2018ibnnet}}
    \end{subfigure}
    \hspace{-4.5pt}
    \begin{subfigure}{0.1424\linewidth}
      \includegraphics[width=1.0\linewidth]{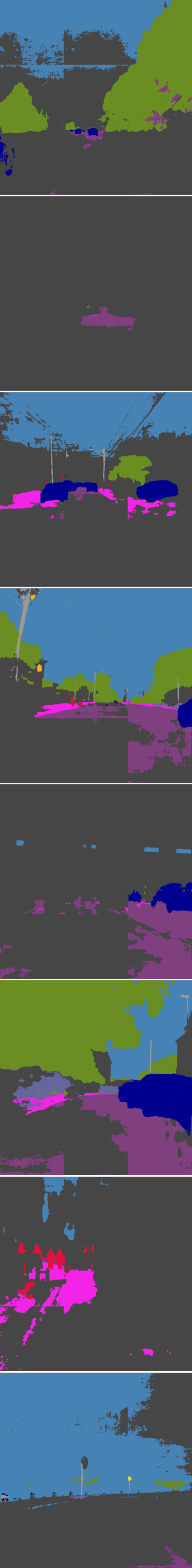}
      \caption{RobustNet~\cite{choi2021robustnet}}
    \end{subfigure}
    \hspace{-4.5pt}
    \begin{subfigure}{0.1428\linewidth}
      \includegraphics[width=1.0\linewidth]{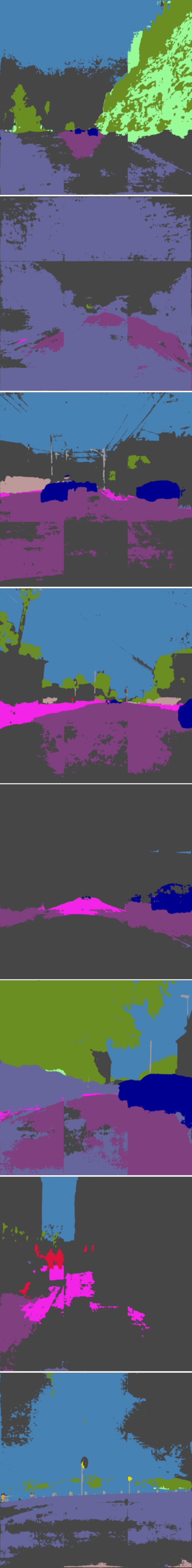}
      \caption{MLDG~\cite{mldg}}
    \end{subfigure}
    \hspace{-4.5pt}
    \begin{subfigure}{0.1425\linewidth}
      \includegraphics[width=1.0\linewidth]{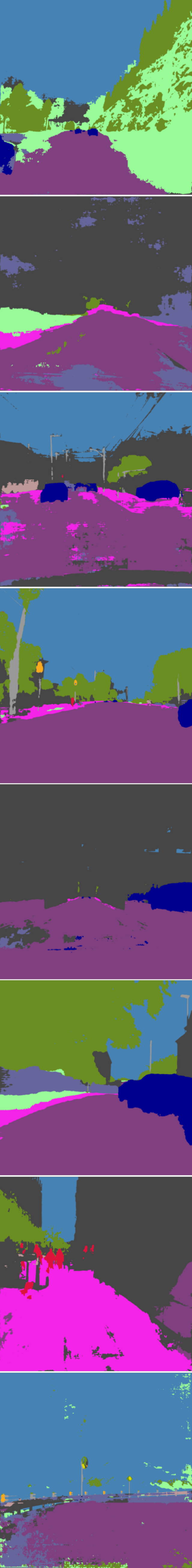}
      \caption{Ours}
    \end{subfigure}

    \vspace{-9pt}
 	\caption{
 	\textbf{Source (G+S)$\rightarrow$Target (M):} [1/2] Qualitative comparison on the Mapillary dataset. All methods adopt DeepLabV3+ with ResNet50. (Best viewed in color.)
 	}\label{fig:qualmap1}\vspace{-7pt}
\end{figure*}

\begin{figure*}[!ht]
	\centering
    \includegraphics[width=0.9\linewidth]{figure/classlabel.pdf}\hfill \\
    \begin{subfigure}{0.1424\linewidth}
      \includegraphics[width=1.0\linewidth]{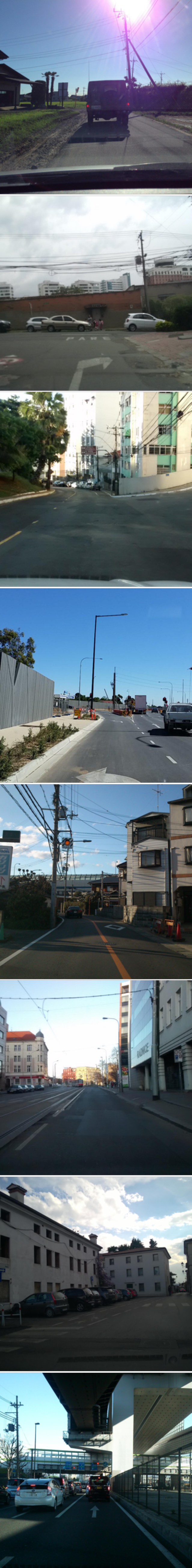}
      \caption{Images}
    \end{subfigure}
    \hspace{-4.6pt}
    \begin{subfigure}{0.1428\linewidth}
      \includegraphics[width=1.0\linewidth]{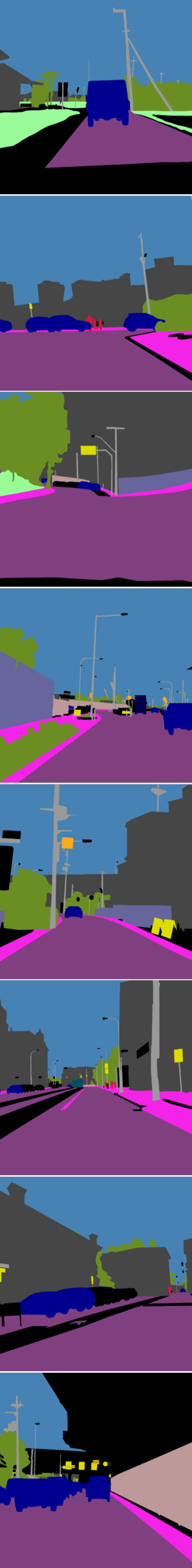}
      \caption{Ground Truth}
    \end{subfigure}
    \hspace{-4.6pt}
    \begin{subfigure}{0.1428\linewidth}
      \includegraphics[width=1.0\linewidth]{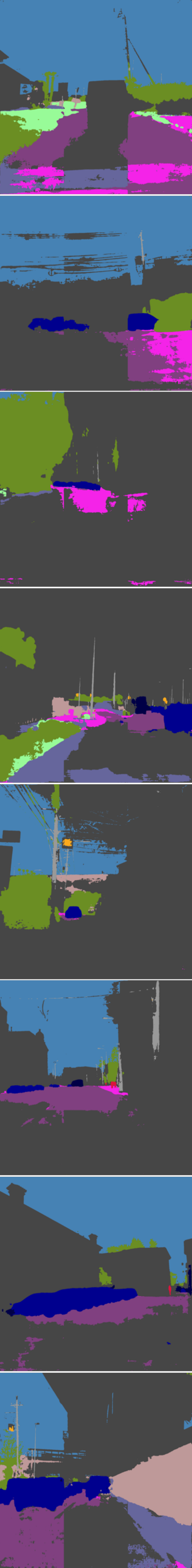}
      \caption{Baseline}
    \end{subfigure}
    \hspace{-4.6pt}
    \begin{subfigure}{0.1428\linewidth}
      \includegraphics[width=1.0\linewidth]{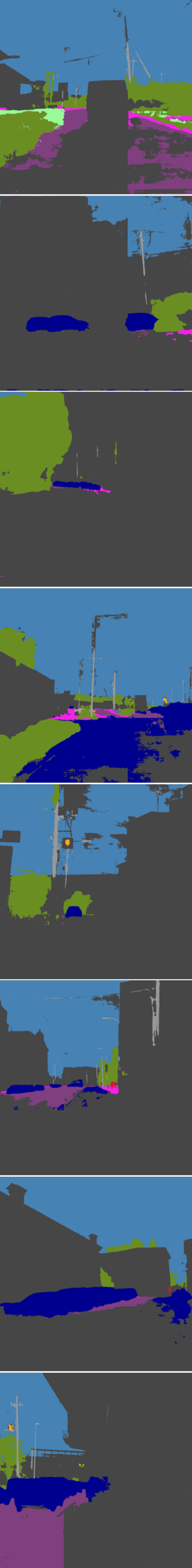}
      \caption{IBN-Net~\cite{pan2018ibnnet}}
    \end{subfigure}
    \hspace{-4.6pt}
    \begin{subfigure}{0.1428\linewidth}
      \includegraphics[width=1.0\linewidth]{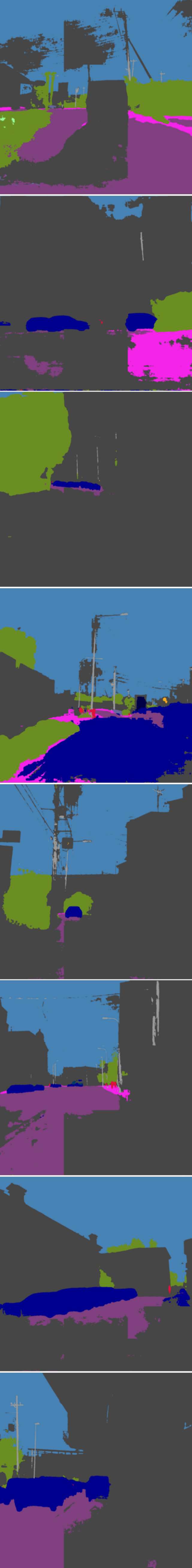}
      \caption{RobustNet~\cite{choi2021robustnet}}
    \end{subfigure}
    \hspace{-4.6pt}
    \begin{subfigure}{0.1428\linewidth}
      \includegraphics[width=1.0\linewidth]{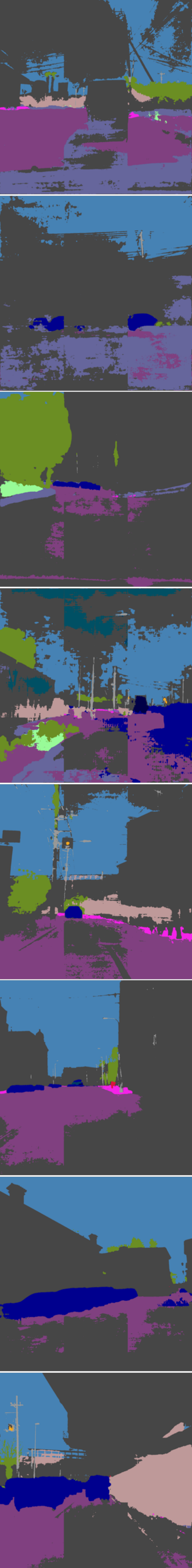}
      \caption{MLDG~\cite{mldg}}
    \end{subfigure}
    \hspace{-4.6pt}
    \begin{subfigure}{0.1428\linewidth}
      \includegraphics[width=1.0\linewidth]{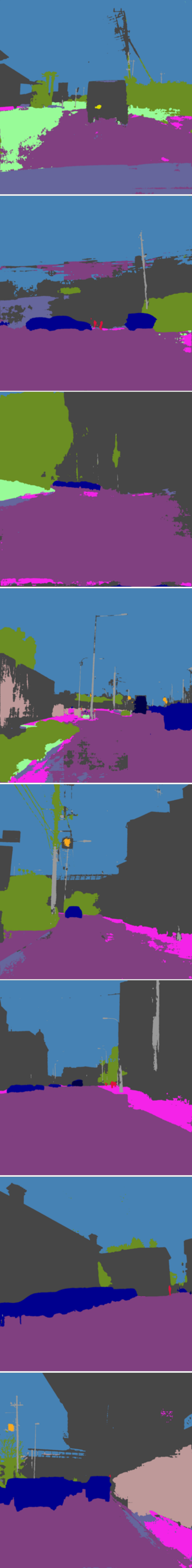}
      \caption{Ours}
    \end{subfigure}
    
    \vspace{-9pt}
 	\caption{
 	\textbf{Source (G+S)$\rightarrow$Target (M):} [2/2] Qualitative comparison on the Mapillary dataset. All methods adopt DeepLabV3+ with ResNet50. (Best viewed in color.)
 	}\label{fig:qualmap2}\vspace{-7pt}
\end{figure*}

Finally, \figref{fig:qualmap1} and \figref{fig:qualmap2} show the segmentation results on the Mapillary dataset.
The Mapillary dataset contains images acquired from various environments in Europe and Asia.
Our method showed more reasonable results than other methods even when the viewpoint or scene structure changes in places such as sidewalks, countryside, residential areas, and shoulder roads.
Moreover, our method successfully predicted a partially snowy or wet road and cloudy sky.
Therefore, we can describe that our memory-guided meta-learning method effectively enhances the encoder features on various distribution shifts.

\newpage 

\newpage

\newpage

\end{document}